\providecommand{\IncludeAppendix}{1}
\documentclass[letterpaper]{article}
\usepackage{aaai2027}
\usepackage[hyphens]{url}
\usepackage{graphicx}
\usepackage{natbib}[sort]
\usepackage{caption}
\usepackage{algorithm}
\usepackage{algorithmic}
\usepackage{booktabs}
\usepackage{amsmath}
\usepackage{amssymb}
\usepackage{amsthm}

\usepackage{colortbl}
\usepackage{tabularx}
\usepackage{multirow}
\usepackage{cuted}
\usepackage{tikz}
\usetikzlibrary{positioning,arrows.meta,fit,calc,backgrounds}

\definecolor{headerblue}{HTML}{D6E4F2}   
\definecolor{tableblue}{HTML}{EEF4FB}    
\definecolor{tablebluealt}{HTML}{E4EDF7} 
\definecolor{sectionblue}{HTML}{BDD3EA}  
\definecolor{bestblue}{HTML}{ABC5E1}     
\definecolor{vlmheader}{HTML}{ECE3F3}    
\definecolor{vitheader}{HTML}{DDECF7}    
\definecolor{hierheader}{HTML}{DDEFE8}   
\definecolor{cnnheader}{HTML}{F7E7DC}    
\definecolor{summaryheader}{HTML}{F5EDD8}
\definecolor{nearblue}{HTML}{E9F0F7}     
\definecolor{farblue}{HTML}{D8E8F5}      
\newcommand{\bestcell}[1]{\cellcolor{bestblue}\textbf{#1}}

\definecolor{obsaccent}{HTML}{2F6FCE}
\definecolor{obsfill}{HTML}{F4F8FC}
\newcommand{\obsbox}[2]{%
  \par\vspace{2pt}\noindent
  \begin{tikzpicture}
  \node[draw=obsaccent, line width=0.55pt, fill=obsfill, rounded corners=3pt,
        inner sep=3pt, align=left, text width=\dimexpr\linewidth-10pt\relax]
    {\footnotesize\textcolor{obsaccent}{\textbf{#1}}%
     \if\relax\detokenize{#2}\relax\else.\ #2\fi};
  \end{tikzpicture}%
  \par\vspace{2pt}
}

\definecolor{claimaccent}{HTML}{2F6FCE}
\newcommand{\claimbox}[2]{%
  \par\smallskip\noindent
  \begin{tikzpicture}
  \node[draw=claimaccent, line width=0.9pt, fill=tableblue, rounded corners=4pt,
        inner sep=6pt, align=left, text width=\dimexpr\linewidth-16pt\relax]
    {\small\textcolor{claimaccent}{\textbf{#1.}}\ #2};
  \end{tikzpicture}%
  \par\smallskip
}

\newcommand{\claimboxnotitle}[1]{%
  \par\smallskip\noindent
  \begin{tikzpicture}
  \node[draw=claimaccent, line width=0.9pt, fill=tableblue, rounded corners=4pt,
        inner sep=6pt, align=left, text width=\dimexpr\linewidth-16pt\relax]
    {\small{#1}};
  \end{tikzpicture}%
  \par\smallskip
}

\definecolor{insightaccent}{HTML}{1B7A47}
\definecolor{insightfill}{HTML}{EAF5EE}
\newcommand{\insightbox}[2]{%
  \par\smallskip\noindent
  \begin{tikzpicture}
  \node[draw=insightaccent, line width=0.9pt, fill=insightfill, rounded corners=4pt,
        inner sep=5pt, align=left, text width=\dimexpr\linewidth-14pt\relax]
    {\small\textcolor{insightaccent}{\textbf{#1.}}\ #2};
  \end{tikzpicture}%
  \par\smallskip
}

\definecolor{takeawayaccent}{HTML}{8A5A00}
\definecolor{takeawayfill}{HTML}{FFF8E8}
\newcommand{\takeawaybox}[2]{%
  \par\smallskip\noindent
  \begin{tikzpicture}
  \node[draw=takeawayaccent, line width=0.7pt, fill=takeawayfill, rounded corners=3pt,
        inner sep=4pt, align=left, text width=\dimexpr\linewidth-12pt\relax]
    {\footnotesize\textcolor{takeawayaccent}{\textbf{#1.}}\ #2};
  \end{tikzpicture}%
  \par\smallskip
}

\definecolor{methodaccent}{HTML}{4B5563}
\definecolor{methodfill}{HTML}{F7F8FA}

\title{Representation Trajectories Matters:\\ Complementary Evidence for OOD Detection and Image Classification}

\author{
    Ignacio M. De la Jara\textsuperscript{1, 3},
    Cristian Rodriguez-Opazo\textsuperscript{2},
    Hamed Damirchi\textsuperscript{1},
    Stephen Gould\textsuperscript{2},
    Damith Ranasinghe\textsuperscript{1,3}
}

\affiliations{
    \textsuperscript{1}University of Adelaide,
    \textsuperscript{2}Australian National University,
    \textsuperscript{3}Naval Group Pacific\\
    ignacio.mezadelajara@adelaide.edu.au
}

\begin{document}
\ifnum\IncludeAppendix=0
\makeatletter
\newcommand{\submissionlabel}[3]{%
  \expandafter\gdef\csname r@#1\endcsname{{#2}{#3}}}
\submissionlabel{app:model-inventory}{C}{11}
\submissionlabel{app:complementarity}{H}{14}
\submissionlabel{app:domain-shift}{K}{18}
\submissionlabel{app:classification-shift}{O}{26}
\submissionlabel{app:deployment}{G}{14}
\submissionlabel{tab:continuity-full}{11}{18}
\makeatother

\nocite{bao2022beit,dosovitskiy2021vit,fang2023eva02,he2016identity,
he2016resnet,he2022mae,huang2017densenet,liu2021swin,liu2022convnext,
liu2022swinv2,peng2022beitv2,radosavovic2020regnet,shapley1953value,
tan2021efficientnetv2,touvron2022deit3,woo2023convnextv2,zagoruyko2016wide}

\fi
\maketitle

\begin{abstract}
Vision models do not form a representation at once; each block revises it. We
ask whether the resulting computation path contains evidence that the final
representation discards, and whether that evidence improves OOD detection and
image classification on clean and shifted data. Unlike approaches that treat
intermediate layers as separate snapshots, we retain sample identity across
depth and study the transformations connecting successive states. We separate
class-coherent transport from input-specific innovation, and coordinate
movement from relational reorganization. Across supervised, self-supervised,
vision--language, hierarchical, and convolutional encoders, these paths show
strong sample-specific continuity and architecture-specific depth profiles
that recur across datasets. They are also practically useful. An ID-only
transition-surprise score complements strong final-state detectors, reducing
FPR95 in 131/152 non-saturated comparisons on a balanced OpenOOD grid; gains
are largest for visually disruptive and semantically far shifts, and remain
positive on near-OOD for most detectors. Frozen update probes improve 71/72
clean model--dataset cases, while shifted-data gains vary with architecture
and corruption type. Computation paths therefore provide a broadly useful
reliability signal whose value is determined jointly by model organization and
the shift encountered.
\end{abstract}

\section{Introduction}
\label{sec:intro}

A vision encoder does not simply output an image representation; it computes one.
Each block updates the current state, progressively changing its geometry and class evidence.
Yet most analyses and downstream systems retain only the endpoint of this computation.
This endpoint is natural for recognition, but it discards the route by which the model arrived there.

This raises a fundamental question:
\emph{does the path through a vision encoder contain evidence that is not captured by its endpoint?}
We study this question at three levels:
whether an image's state at one layer predicts its later states,
whether classes follow coherent routes,
and whether architectures organise those routes differently.
We then ask whether such path evidence improves reliability under distribution shift.

Prior work has visualised intermediate features
\citep{zeiler2014visualizing},
compared representations across layers and models
\citep{kornblith2019similarity},
and scored individual layers for OOD detection
\citep{lee2018mahalanobis}.
Closer to the computation itself are between-layer transformation
smoothness in text encoders \citep{jelenic2024blood} and fusion of
intermediate CLIP evidence \citep{meza2025mysteries}.
All score layers or adjacent pairs, within a single modality or encoder
family.
We instead ask whether the \emph{same input's path} through the network
has coherent structure across depth, and whether it provides evidence
complementary to the final representation.

\begin{figure}[!t]
\centering
\includegraphics[width=\columnwidth]{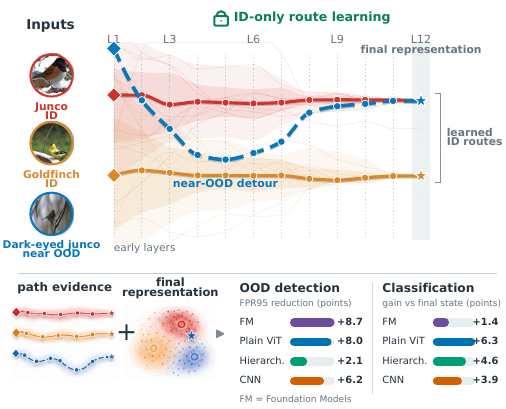}
\caption{\textbf{The trajectories matters.}
Images with similar final representations can follow different computational routes.
Typical ID samples remain close to learned class trajectories, whereas OOD samples deviate despite similar endpoints.
Trajectory evidence complements the final representation for both OOD detection and image classification; bars report average gains across architecture families.}
\label{fig:teaser}
\end{figure}

Our approach links the layer states of each input into a trajectory rather than treating them as independent snapshots.
For each class, we estimate a typical ID route and express individual samples relative to it.
This separates class-shared motion from input-specific deviations.

We first verify that these trajectories are not arbitrary collections of layer states.
Paired and shuffled controls show that the same image retains predictable cross-layer structure, while direction controls prevent us from interpreting depth as a physical-time process.

Structured trajectories need not look the same across architectures.
Different encoders may distribute class evidence, geometric movement, and relational change across different layers.
For this reason, we study trajectories in each model's native block sequence rather than imposing a common middle layer.
Our evaluation spans supervised, self-supervised, vision--language, hierarchical, and convolutional encoders.

We evaluate whether trajectory structure matters beyond analysis using OOD detection and image classification.
OOD detection asks whether atypical routes reveal distribution shift, while recognition asks whether intermediate updates retain class evidence that the final state does not fully expose.

For OOD detection, an ID-only transition-surprise score complements a final-state density score.
Across the non-saturated balanced OpenOOD evaluation, this lowers FPR95 in 131 of 152 backbone--benchmark comparisons, including every CIFAR-10 and CIFAR-100 case.
Gains are largest for visually disruptive and semantically far shifts, and hold on near-OOD for most detectors; the residual limitation is confined to the hardest ImageNet near splits.

For recognition, frozen probes on native block updates are selected and fused with a final-state probe using ID/source validation only.
This improves 71 of 72 clean model--dataset cases.
Under label-preserving shift, however, the benefit depends on architecture and corruption type.
The trajectory therefore complements rather than replaces the final representation, providing a reliability signal whose value depends jointly on the model and the shift encountered.

\insightbox{In short}{The final representation says what the model represents,
confidence says how certain the prediction is, and the trajectory says whether
the computation followed a familiar class route.
We use deviations from that route as complementary evidence for OOD detection and recognition.}

\paragraph{Contributions.}
\begin{enumerate}\setlength{\itemsep}{1pt}
  \item We formulate representation trajectories as a reliability object and
  decompose native updates into class-coherent transport and sample-specific
  innovation.
  \item We establish sample-specific continuity through paired, shuffled,
  history, gap, and direction controls, while documenting the absence of a
  privileged forward arrow.
  \item We show that architecture families allocate class evidence and
  representational change differently across native blocks, motivating
  architecture-balanced trajectory analysis.
  \item We introduce an ID-only trajectory-surprise complement to final-state
  OOD scores and map when intermediate computation helps recognition, including
  broad gains and explicit neutral or negative results.
\end{enumerate}

\section{Related Work}

\begin{figure*}[t]
\centering
\includegraphics[width=0.94\textwidth]{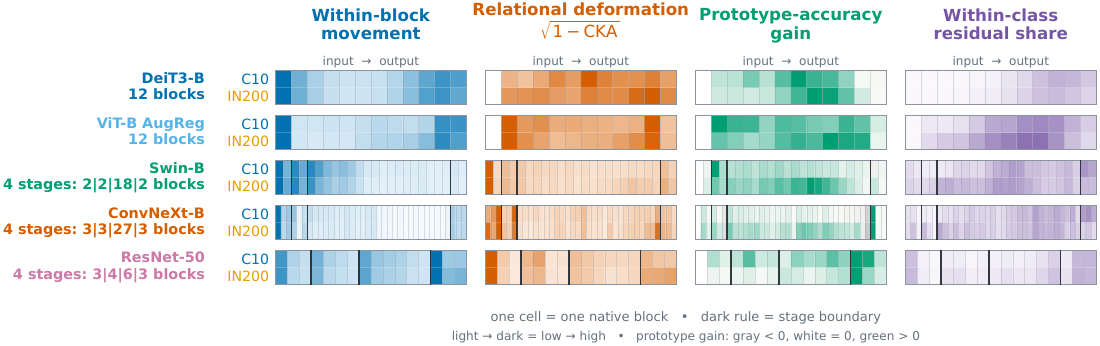}
\caption{\textbf{Architectures allocate work differently.} Each strip follows
one checkpoint--dataset pair across its native blocks. Movement, relational
deformation $\sqrt{1-\mathrm{CKA}}$, class-evidence gain, and residual share
peak at different depths; colour is normalised within each row except residual
share.}
\vspace{-1em}
\label{fig:block-work}
\end{figure*}

\paragraph{Final-representation reliability.}
Post-hoc OOD detectors typically score the final prediction or representation:
MSP~\citep{hendrycks2017baseline}, ODIN~\citep{liang2018odin},
energy~\citep{liu2020energy}, MaxLogit~\citep{hendrycks2022scaling},
ReAct~\citep{sun2021react}, and ASH~\citep{djurisic2023ash} operate on logits or
final-state activations. Feature-space methods include Mahalanobis
distance~\citep{lee2018mahalanobis}, deep $k$-NN~\citep{sun2022knn},
ViM~\citep{wang2022vim}, Relative Mahalanobis~\citep{ren2021relativemaha}, and
Mahalanobis++~\citep{mueller2025mahapp}. Vision--language detectors similarly
score the final image--text representation, as in
MCM~\citep{ming2022mcm}. These methods estimate final-state typicality or
confidence. Our trajectory score instead asks whether the computation leading
to that state provides complementary ID-only evidence.

\paragraph{Intermediate representations.}
Early work visualised intermediate features, analysed their evolution across
depth, and linked units to semantic
concepts~\citep{zeiler2014visualizing,yosinski2014transferable,bau2017network}.
Subsequent studies used linear probes and representation similarity to
characterise the emergence of class information and feature
geometry~\citep{alain2017probes,kornblith2019similarity,raghu2021vision},
including for foundation encoders~\citep{bolya2025pe}. OOD methods further
exploit intermediate representations by combining per-layer Mahalanobis
scores~\citep{lee2018mahalanobis}, learning feature
mixtures~\citep{xmahalanobis2025}, fusing intermediate
evidence~\citep{meza2025mysteries}, using intermediate-layer
classifiers~\citep{uselis2025ilc}, or measuring between-layer
transformations~\citep{jelenic2024blood}. These approaches treat intermediate
representations as observations to combine rather than modelling the
computation connecting them. Related work also shows that representations
learned under the same objective can remain complementary, motivating adaptive
combinations across backbones~\citep{10350366,rodriguezopazo2025synergydiversityclipenhancing}.


\paragraph{Representation trajectories.}
Representation trajectories have recently emerged as a distinct line of work.
Population-level studies characterise neural networks as paths through
representation space, describing how feature geometry evolves across
layers~\citep{lange2022paths}. More recently, trajectories in large language
models have been used to analyse reasoning dynamics and predict reasoning
correctness and truthfulness~\citep{damirchi2026truth}. We instead preserve the
identity of each sample across depth and study the computation trajectory of a
frozen vision model, asking whether it provides complementary evidence beyond
the final representation for OOD detection and image classification.

\obsbox{The missing comparison}{
Previous work studies final representations, individual intermediate states,
learned combinations of representations, or reasoning trajectories. We instead
ask whether the recorded computation trajectory of each sample provides
complementary evidence beyond the final representation.
}

\section{Representation Trajectories}
\label{sec:theory}

Vision models compute representations through a sequence of intermediate transformations, yet most methods retain only the endpoint of this process. We represent the forward pass as a \emph{representation trajectory},
\begin{equation}
    \tau(x)=(z_1(x),\ldots,z_L(x)),
\end{equation}
where $z_L$ is the final representation used by downstream tasks. The corresponding update between consecutive representations is $u_l(x)=z_{l+1}(x)-z_l(x)$, which captures the local movement of the trajectory between adjacent layers.
For architectures with varying dimensionality, updates are computed within each stage, while cross-stage comparisons use a fixed Gaussian random projection \citep{johnson1984extensions,dasgupta2003elementary}.

\claimboxnotitle{
The trajectory is the complete dynamical evolution of a sample through the network, characterized by its states and the transitions between them.
}

\subsection{Class routes and innovations}

A trajectory describes how one image evolves through the network, but not what is shared across a class. We therefore estimate the mean trajectory of each ID class,
\begin{equation}
\mu_l^c=\mathbb E[z_l(X)\mid Y=c],
\end{equation}
which we call the \emph{class route}. Every sample is then expressed relative to its class route,
\begin{equation}
r_l(x,c)=z_l(x)-\mu_l^c.
\end{equation}
The residual trajectory therefore isolates the computation that remains specific to each sample. Each update naturally decomposes into:
\begin{equation}
\underbrace{z_{l+1}-z_l}_{\text{total update}}
=
\underbrace{\mu_{l+1}^c-\mu_l^c}_{\text{class-coherent transport}}
+
\underbrace{r_{l+1}-r_l}_{\text{sample-specific innovation}}.
\label{eq:decomposition}
\end{equation}

Equation~\ref{eq:decomposition} separates computation shared across a class from computation that is specific to an individual image. At test time, when the class label is unknown, the class route is estimated from the model's final representation.

\subsection{Locating the quantity that must be measured}
The remaining question is whether a trajectory contains information
beyond the final representation. Two inputs may arrive at similar final
representations while following different computational routes. If so,
the trajectory preserves information that cannot be recovered from the
endpoint alone. The following decomposition identifies where such
information would have to reside.

Let $R=(Z_1,\ldots,Z_{L-1})$ denote the trajectory and $E=Z_L$ the
final representation, and let $P$ and $Q$ denote the in-distribution
and shifted joint distributions over $(R,E)$. Whenever the
corresponding KL divergences are well defined, the chain rule gives
\begin{equation}
\begin{aligned}
D_{\mathrm{KL}}(P_{R,E}\|Q_{R,E})
&=
D_{\mathrm{KL}}(P_E\|Q_E)\\
&\quad+
\mathbb E_{e\sim P_E}
D_{\mathrm{KL}}
\!\left(
P_{R\mid E=e}
\|
Q_{R\mid E=e}
\right).
\end{aligned}
\label{eq:chain}
\end{equation}
Equation~\ref{eq:chain} separates the divergence between two
distributions into a component visible from the final representation
and a residual component that survives after conditioning on that
endpoint. This is an identity: the residual term is non-negative by
construction, and vanishes whenever the blocks are invertible. It
does not does not assert that a computation path carries additional
evidence; it specifies where such evidence would have to reside, and
which quantity an experiment would have to estimate.

\claimbox{The endpoint need not determine the route}{
Conditioning on the final representation leaves a residual divergence
that may or may not be exploitable. Equation~\ref{eq:chain} says where
to look; Section~4 builds a score that looks there, and
Section~5.2 reports what it finds.
}

\subsection{Four complementary measurements}




Figure~\ref{fig:block-work} uses four measurements to show that a trajectory is
not captured by a single layer statistic. Movement measures the size of the
local update. Relational deformation measures how pairwise geometry changes
between adjacent states. Prototype-accuracy gain measures where class evidence
increases. Within-class residual share measures how much of the update remains
sample-specific after subtracting the class route.
The four measurements peak at different layers. A block can move features
substantially without rewriting pairwise relations, or add class evidence with
only modest movement. \textbf{Thus trajectory analysis should distinguish movement,
geometry, class evidence, and sample-specific innovation rather than reduce the
path to one score.}

The profiles also differ across architectures. Two 12-block plain ViTs exhibit
different schedules despite identical depth, while Swin and ConvNeXt concentrate
some changes around stage boundaries and continue accumulating class evidence
inside their longest stages. Across CIFAR-10 and ImageNet-200, movement and
relational deformation are more stable than the timing of prototype-accuracy
gain. These patterns motivate studying each model's native trajectory rather
than relying on a hand-picked middle layer.

\section{Using the Recorded Path}
\label{sec:method}


We use the trajectory representation in two complementary tasks. OOD detection uses 
an ID-only transition-surprise score derived from residual trajectories, while image 
classification combines probes on native updates with the final representation.
Both keep the backbone frozen and learn only from ID.

\subsection{ID-only transition surprise}
\label{sec:method-ood}

For each training sample, we assign the class route and form residual states
$r_{1:L}$. A single MLP $f_\theta$, shared across transitions and conditioned
by a learned depth embedding $e_l$, predicts
$\widehat r_{l+1}=f_\theta(r_l+e_l)$. On ID fitting data, we estimate the
component-wise prediction-error scale
\[
\widehat\sigma^2_{l,d}
=N^{-1}\sum_i
\left(r_{i,l+1,d}-[f_\theta(r_{i,l}+e_l)]_d\right)^2.
\]
The trajectory surprise is
\begin{equation}
 D(x)=\sum_{l=1}^{L-1}\sum_d
 \frac{(r_{l+1,d}(x)-\widehat r_{l+1,d}(x))^2}
      {\widehat\sigma^2_{l,d}}.
 \label{eq:dynamics-score}
\end{equation}
Component-wise scaling emphasises reliably predictable ID transitions, while
summation allows small departures to accumulate across depth.

\paragraph{Novel-evidence fusion.}
Let $S_0$ be a final-state OOD score such as normalised Mahalanobis++. We
standardise $S_0$ and $D$ on held-out ID data, denoting the resulting scores
by $\widetilde{S}_0$ and $\widetilde{D}$, regress
$\widetilde{D}$ on $\widetilde{S}_0$, and retain the residual path evidence:
\begin{equation}
S(x)=\widetilde{S}_0(x)
+0.3\left[\widetilde{D}(x)-a\,\widetilde{S}_0(x)-b\right].
\label{eq:combiner}
\end{equation}
The coefficient $0.3$ is fixed \emph{a priori} and never adjusted per
checkpoint, benchmark, or split. It is deliberately conservative, keeping
the final-state score dominant so that the trajectory acts as a correction
rather than a replacement. As shown in
Fig~\ref{fig:weight-sweep}, larger weights provide only marginal additional
average improvement while increasing both benchmark-to-benchmark variability
and the worst-case degradation. The choice of $0.3$ therefore favours robust
complementary evidence over aggressive score replacement. The offsets
$(a,b)$ and all standardisation statistics are estimated from ID data only.
This tests whether the trajectory provides evidence beyond the final
representation. We repeat the same sequential test after absorbing Relative
Mahalanobis and compare against random-noise and redundant-score partners.

\subsection{Update probes for recognition}
\label{sec:method-recognition}

For every compatible native update $u_l$, we train a linear probe
$q_l(y\mid u_l)$~\citep{alain2017probes}. An inner source-validation split selects probe
regularisation. Using a separate fusion-validation split, we rank probes,
learn non-negative ensemble weights and temperatures, and
fit a final-state margin gate:
\begin{align}
 q_{\mathrm{path}}(y\mid x)&=\sum_l w_lq_l(y\mid u_l(x)),\\
 q_{\mathrm{fuse}}&=(1-\alpha)q_L+\alpha q_{\mathrm{path}}.
\end{align}
The gate can restrict path corrections to final-state-ambiguous samples. After
selection, every setting is frozen for the clean test set, corruptions, or an
unseen target domain. A separately seeded final-state ensemble is the
matched control for generic ensembling.

\subsection{Validating trajectory continuity}

\begin{figure}[t]
\centering
\includegraphics[width=\columnwidth]
{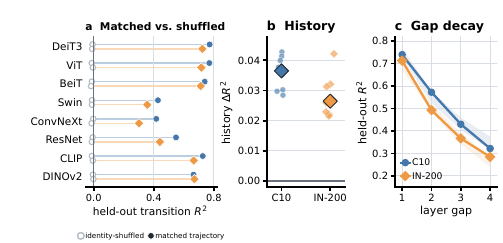}
\caption{\textbf{Matched trajectories preserve sample-specific continuity.}
(a) Matched transitions yield higher held-out prediction $R^2$ than
identity-shuffled pairs in all 16 cases.
(b) History adds predictive information.
(c) Predictability decays with layer gap.
Full values are in Appendix Table~\ref{tab:continuity-full}.}
\label{fig:path-coherence}
\end{figure}

Before using paths downstream, we test whether consecutive residuals preserve
sample-specific information. In Figure~\ref{fig:path-coherence}, the
\emph{matched-trajectory} condition predicts $r_{l+1}(x)$ from $r_l(x)$ for the
same input. The \emph{identity-shuffled} control instead pairs $r_{l+1}(x)$
with $r_l(x')$ from another image, preserving both layer marginals while
breaking their identity link. Their held-out $R^2$ difference therefore
isolates cross-depth continuity. Matched trajectories win in all 16
checkpoint--dataset comparisons (a); preceding-state history adds information
(b), and predictability decays as the target moves farther ahead (c).


\begin{figure*}[!t]
\centering
\includegraphics[width=0.94\textwidth]{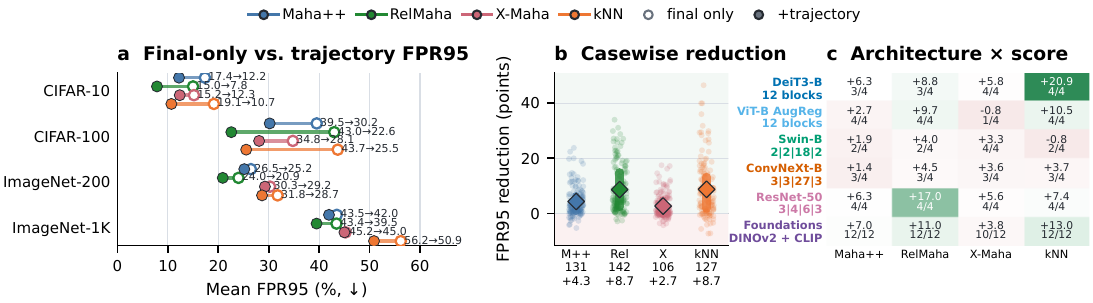}
\caption{\textbf{Trajectory gains depend on score and architecture.}
\emph{(a)} Final-layer-only (open) and trajectory-augmented (filled) FPR95;
\emph{(b)} 152 paired reductions with improved-case counts, interquartile ranges, and mean $\Delta$;
\emph{(c)} mean reduction and improved/total cases by architecture and score.}
\label{fig:results-at-glance}
\vspace{-1.5em}
\end{figure*}

Separate MLP, GRU, LSTM, causal-attention, and temporal-convolution predictors
all find reverse prediction easier (Appendix~\ref{app:direction-capacity}). We
therefore use the one-step MLP because it gives the best median forward
prediction, while interpreting the result as evidence of cross-depth continuity
rather than a temporal arrow. Consistent performance under fixed layer
permutations further argues against privileged ordering.

\section{Experiments}
\label{sec:experiments}

\subsection{Evaluation protocol}

To evaluate the trajectory signals introduced in the previous sections, we
study their predictive utility on complementary OOD detection and recognition
tasks using fixed backbones, ID-only training, and standard benchmarks
throughout.

\paragraph{OOD detection.}
We evaluate the fixed path-surprise method on OpenOOD v1.5
\citep{yang2022openood,zhang2023openood}, covering MNIST, CIFAR-10,
CIFAR-100, ImageNet-200, and ImageNet-1K with their prescribed OOD splits.
The balanced evaluation comprises 190 backbone--benchmark pairs, using the
same 38 supervised, self-supervised, vision--language, hierarchical, and CNN
checkpoints on each benchmark; four additional ImageNet-only checkpoints are
reported in Appendix~\ref{app:model-inventory} but excluded from the balanced
comparison. The method is fitted using labelled ID data only with three random
seeds and no OOD exposure. We report AUROC and FPR95, using Mahalanobis++ as
the primary final-state baseline together with Relative Mahalanobis,
X-Mahalanobis, and final-state kNN~\citep{sun2022knn}.

\paragraph{Recognition and shift.}
Recognition uses the same six representative architectures (CLIP ViT-B/16,
DeiT3-B, supervised ViT-B, Swin-B/SwinV2-B, ConvNeXt-B, and ResNet-50) across
twelve datasets spanning generic objects, textures, scenes, remote sensing,
and fine-grained recognition, reporting mean top-1 accuracy over three probe
seeds. Shift robustness is evaluated on CIFAR-100-C, PACS, and Office-Home:
for CIFAR-100-C, probes and fusion are selected on clean CIFAR-100 before
evaluation on all corruption types and severities, while PACS and Office-Home
follow leave-one-domain-out evaluation, where the target domain is never used
to tune probes, temperatures, ensemble weights, or gates.

\begin{table*}[!t]
\centering
\caption{\textbf{Clean classification.} Final state linear-probe $\rightarrow$ residual-path fusion accuracy (top-1 \%, three seeds). Green $\Delta$ is the absolute gain; the right margin summarizes relative dataset gains.}
\label{tab:clean-classification}
\scriptsize
\setlength{\tabcolsep}{2.65pt}
\renewcommand{\arraystretch}{1.08}
\begin{tabularx}{\textwidth}{>{\raggedright\arraybackslash}p{23mm}*{7}{>{\centering\arraybackslash}X}}
\specialrule{0.1em}{0pt}{0pt}
\rowcolor{headerblue}
\textbf{Dataset} & \multicolumn{1}{c}{\cellcolor{vlmheader}\textbf{VLM}} & \multicolumn{2}{c}{\cellcolor{vitheader}\textbf{Plain ViTs}} & \multicolumn{1}{c}{\cellcolor{hierheader}\textbf{Hierarchical}} & \multicolumn{2}{c}{\cellcolor{cnnheader}\textbf{CNNs}} & \multicolumn{1}{c}{\cellcolor{summaryheader}\textbf{Row}} \\
\rowcolor{headerblue}
 & \cellcolor{vlmheader}\textbf{CLIP-B} & \cellcolor{vitheader}\textbf{DeiT3-B} & \cellcolor{vitheader}\textbf{ViT-B} & \cellcolor{hierheader}\textbf{Swin} & \cellcolor{cnnheader}\textbf{ConvNeXt-B} & \cellcolor{cnnheader}\textbf{ResNet-50} & \cellcolor{summaryheader}\textbf{Rel. gain} \\
\midrule
\rowcolor{tablebluealt}
CIFAR-100 & \shortstack{$80.80\!\rightarrow\!\mathbf{83.53}$\\[-1pt]{\tiny\textcolor{insightaccent}{$\Delta=+2.72$}}} & \shortstack{$83.22\!\rightarrow\!\mathbf{86.65}$\\[-1pt]{\tiny\textcolor{insightaccent}{$\Delta=+3.43$}}} & \shortstack{$77.07\!\rightarrow\!\mathbf{82.12}$\\[-1pt]{\tiny\textcolor{insightaccent}{$\Delta=+5.05$}}} & \shortstack{$80.96\!\rightarrow\!\mathbf{83.89}$\\[-1pt]{\tiny\textcolor{insightaccent}{$\Delta=+2.93$}}} & \shortstack{$78.95\!\rightarrow\!\mathbf{81.51}$\\[-1pt]{\tiny\textcolor{insightaccent}{$\Delta=+2.57$}}} & \shortstack{$72.75\!\rightarrow\!\mathbf{77.31}$\\[-1pt]{\tiny\textcolor{insightaccent}{$\Delta=+4.55$}}} & \textcolor{insightaccent}{4.5\%} \\
\rowcolor{tableblue}
DTD & \shortstack{$75.41\!\rightarrow\!\mathbf{77.16}$\\[-1pt]{\tiny\textcolor{insightaccent}{$\Delta=+1.76$}}} & \shortstack{$67.43\!\rightarrow\!\mathbf{74.70}$\\[-1pt]{\tiny\textcolor{insightaccent}{$\Delta=+7.27$}}} & \shortstack{$63.79\!\rightarrow\!\mathbf{70.99}$\\[-1pt]{\tiny\textcolor{insightaccent}{$\Delta=+7.20$}}} & \shortstack{$69.43\!\rightarrow\!\mathbf{75.48}$\\[-1pt]{\tiny\textcolor{insightaccent}{$\Delta=+6.05$}}} & \shortstack{$69.66\!\rightarrow\!\mathbf{75.94}$\\[-1pt]{\tiny\textcolor{insightaccent}{$\Delta=+6.28$}}} & \shortstack{$67.52\!\rightarrow\!\mathbf{73.48}$\\[-1pt]{\tiny\textcolor{insightaccent}{$\Delta=+5.96$}}} & \textcolor{insightaccent}{8.4\%} \\
\rowcolor{tablebluealt}
EuroSAT & \shortstack{$95.86\!\rightarrow\!\mathbf{97.49}$\\[-1pt]{\tiny\textcolor{insightaccent}{$\Delta=+1.64$}}} & \shortstack{$95.70\!\rightarrow\!\mathbf{98.05}$\\[-1pt]{\tiny\textcolor{insightaccent}{$\Delta=+2.35$}}} & \shortstack{$95.72\!\rightarrow\!\mathbf{97.93}$\\[-1pt]{\tiny\textcolor{insightaccent}{$\Delta=+2.21$}}} & \shortstack{$96.38\!\rightarrow\!\mathbf{97.78}$\\[-1pt]{\tiny\textcolor{insightaccent}{$\Delta=+1.40$}}} & \shortstack{$94.69\!\rightarrow\!\mathbf{97.48}$\\[-1pt]{\tiny\textcolor{insightaccent}{$\Delta=+2.79$}}} & \shortstack{$95.70\!\rightarrow\!\mathbf{97.59}$\\[-1pt]{\tiny\textcolor{insightaccent}{$\Delta=+1.89$}}} & \textcolor{insightaccent}{2.1\%} \\
\rowcolor{tableblue}
Caltech-101 & \shortstack{$96.91\!\rightarrow\!\mathbf{97.27}$\\[-1pt]{\tiny\textcolor{insightaccent}{$\Delta=+0.36$}}} & \shortstack{$94.68\!\rightarrow\!\mathbf{97.27}$\\[-1pt]{\tiny\textcolor{insightaccent}{$\Delta=+2.59$}}} & \shortstack{$91.86\!\rightarrow\!\mathbf{95.24}$\\[-1pt]{\tiny\textcolor{insightaccent}{$\Delta=+3.38$}}} & \shortstack{$94.82\!\rightarrow\!\mathbf{95.78}$\\[-1pt]{\tiny\textcolor{insightaccent}{$\Delta=+0.96$}}} & \shortstack{$92.82\!\rightarrow\!\mathbf{94.51}$\\[-1pt]{\tiny\textcolor{insightaccent}{$\Delta=+1.69$}}} & \shortstack{$93.55\!\rightarrow\!\mathbf{95.39}$\\[-1pt]{\tiny\textcolor{insightaccent}{$\Delta=+1.84$}}} & \textcolor{insightaccent}{1.9\%} \\
\rowcolor{tablebluealt}
CIFAR-10 & \shortstack{$95.24\!\rightarrow\!\mathbf{95.46}$\\[-1pt]{\tiny\textcolor{insightaccent}{$\Delta=+0.22$}}} & \shortstack{$96.77\!\rightarrow\!\mathbf{97.42}$\\[-1pt]{\tiny\textcolor{insightaccent}{$\Delta=+0.65$}}} & \shortstack{$93.88\!\rightarrow\!\mathbf{95.08}$\\[-1pt]{\tiny\textcolor{insightaccent}{$\Delta=+1.20$}}} & \shortstack{$95.51\!\rightarrow\!\mathbf{96.00}$\\[-1pt]{\tiny\textcolor{insightaccent}{$\Delta=+0.49$}}} & \shortstack{$94.75\!\rightarrow\!\mathbf{95.20}$\\[-1pt]{\tiny\textcolor{insightaccent}{$\Delta=+0.44$}}} & \shortstack{$91.48\!\rightarrow\!\mathbf{92.99}$\\[-1pt]{\tiny\textcolor{insightaccent}{$\Delta=+1.51$}}} & \textcolor{insightaccent}{0.8\%} \\
\rowcolor{tableblue}
CUB-200 & \shortstack{$79.28\!\rightarrow\!\mathbf{80.54}$\\[-1pt]{\tiny\textcolor{insightaccent}{$\Delta=+1.26$}}} & \shortstack{$73.18\!\rightarrow\!\mathbf{81.53}$\\[-1pt]{\tiny\textcolor{insightaccent}{$\Delta=+8.34$}}} & \shortstack{$63.14\!\rightarrow\!\mathbf{70.03}$\\[-1pt]{\tiny\textcolor{insightaccent}{$\Delta=+6.89$}}} & \shortstack{$70.64\!\rightarrow\!\mathbf{78.10}$\\[-1pt]{\tiny\textcolor{insightaccent}{$\Delta=+7.46$}}} & \shortstack{$66.21\!\rightarrow\!\mathbf{66.75}$\\[-1pt]{\tiny\textcolor{insightaccent}{$\Delta=+0.54$}}} & \shortstack{$64.75\!\rightarrow\!\mathbf{65.08}$\\[-1pt]{\tiny\textcolor{insightaccent}{$\Delta=+0.33$}}} & \textcolor{insightaccent}{6.0\%} \\
\rowcolor{tablebluealt}
Stanford Cars & \shortstack{$84.81\!\rightarrow\!\mathbf{86.16}$\\[-1pt]{\tiny\textcolor{insightaccent}{$\Delta=+1.35$}}} & \shortstack{$59.12\!\rightarrow\!\mathbf{75.14}$\\[-1pt]{\tiny\textcolor{insightaccent}{$\Delta=+16.02$}}} & \shortstack{$39.54\!\rightarrow\!\mathbf{55.02}$\\[-1pt]{\tiny\textcolor{insightaccent}{$\Delta=+15.48$}}} & \shortstack{$50.38\!\rightarrow\!\mathbf{62.16}$\\[-1pt]{\tiny\textcolor{insightaccent}{$\Delta=+11.78$}}} & \shortstack{$50.61\!\rightarrow\!\mathbf{53.30}$\\[-1pt]{\tiny\textcolor{insightaccent}{$\Delta=+2.69$}}} & \shortstack{$48.28\!\rightarrow\!\mathbf{59.81}$\\[-1pt]{\tiny\textcolor{insightaccent}{$\Delta=+11.54$}}} & \textcolor{insightaccent}{17.7\%} \\
\rowcolor{tableblue}
Oxford-IIIT Pets & \shortstack{$92.54\!\rightarrow\!\mathbf{92.75}$\\[-1pt]{\tiny\textcolor{insightaccent}{$\Delta=+0.21$}}} & \shortstack{$92.79\!\rightarrow\!\mathbf{93.78}$\\[-1pt]{\tiny\textcolor{insightaccent}{$\Delta=+0.99$}}} & \shortstack{$91.00\!\rightarrow\!\mathbf{91.74}$\\[-1pt]{\tiny\textcolor{insightaccent}{$\Delta=+0.74$}}} & \shortstack{$93.06\!\rightarrow\!\mathbf{94.13}$\\[-1pt]{\tiny\textcolor{insightaccent}{$\Delta=+1.07$}}} & \shortstack{$92.15\!\rightarrow\!\mathbf{92.43}$\\[-1pt]{\tiny\textcolor{insightaccent}{$\Delta=+0.28$}}} & \shortstack{$92.68\!\rightarrow\!\mathbf{92.60}$\\[-1pt]{\tiny\textcolor{red!70!black}{$\Delta=-0.07$}}} & \textcolor{insightaccent}{0.6\%} \\
\rowcolor{tablebluealt}
Food-101 & \shortstack{$91.99\!\rightarrow\!\mathbf{92.39}$\\[-1pt]{\tiny\textcolor{insightaccent}{$\Delta=+0.40$}}} & \shortstack{$78.07\!\rightarrow\!\mathbf{83.23}$\\[-1pt]{\tiny\textcolor{insightaccent}{$\Delta=+5.16$}}} & \shortstack{$72.11\!\rightarrow\!\mathbf{78.40}$\\[-1pt]{\tiny\textcolor{insightaccent}{$\Delta=+6.29$}}} & \shortstack{$79.44\!\rightarrow\!\mathbf{82.47}$\\[-1pt]{\tiny\textcolor{insightaccent}{$\Delta=+3.03$}}} & \shortstack{$76.38\!\rightarrow\!\mathbf{79.01}$\\[-1pt]{\tiny\textcolor{insightaccent}{$\Delta=+2.63$}}} & \shortstack{$68.08\!\rightarrow\!\mathbf{75.98}$\\[-1pt]{\tiny\textcolor{insightaccent}{$\Delta=+7.89$}}} & \textcolor{insightaccent}{5.5\%} \\
\rowcolor{tableblue}
Flowers-102 & \shortstack{$95.63\!\rightarrow\!\mathbf{96.58}$\\[-1pt]{\tiny\textcolor{insightaccent}{$\Delta=+0.95$}}} & \shortstack{$82.37\!\rightarrow\!\mathbf{94.31}$\\[-1pt]{\tiny\textcolor{insightaccent}{$\Delta=+11.94$}}} & \shortstack{$76.72\!\rightarrow\!\mathbf{88.47}$\\[-1pt]{\tiny\textcolor{insightaccent}{$\Delta=+11.75$}}} & \shortstack{$81.13\!\rightarrow\!\mathbf{90.61}$\\[-1pt]{\tiny\textcolor{insightaccent}{$\Delta=+9.48$}}} & \shortstack{$71.39\!\rightarrow\!\mathbf{82.97}$\\[-1pt]{\tiny\textcolor{insightaccent}{$\Delta=+11.58$}}} & \shortstack{$77.91\!\rightarrow\!\mathbf{86.47}$\\[-1pt]{\tiny\textcolor{insightaccent}{$\Delta=+8.57$}}} & \textcolor{insightaccent}{11.2\%} \\
\rowcolor{tablebluealt}
FGVC-Aircraft & \shortstack{$50.17\!\rightarrow\!\mathbf{55.50}$\\[-1pt]{\tiny\textcolor{insightaccent}{$\Delta=+5.33$}}} & \shortstack{$45.94\!\rightarrow\!\mathbf{59.27}$\\[-1pt]{\tiny\textcolor{insightaccent}{$\Delta=+13.32$}}} & \shortstack{$36.96\!\rightarrow\!\mathbf{46.47}$\\[-1pt]{\tiny\textcolor{insightaccent}{$\Delta=+9.51$}}} & \shortstack{$41.56\!\rightarrow\!\mathbf{48.44}$\\[-1pt]{\tiny\textcolor{insightaccent}{$\Delta=+6.88$}}} & \shortstack{$46.16\!\rightarrow\!\mathbf{46.75}$\\[-1pt]{\tiny\textcolor{insightaccent}{$\Delta=+0.59$}}} & \shortstack{$38.63\!\rightarrow\!\mathbf{46.65}$\\[-1pt]{\tiny\textcolor{insightaccent}{$\Delta=+8.02$}}} & \textcolor{insightaccent}{16.8\%} \\
\rowcolor{tableblue}
SUN397 & \shortstack{$77.10\!\rightarrow\!\mathbf{78.11}$\\[-1pt]{\tiny\textcolor{insightaccent}{$\Delta=+1.00$}}} & \shortstack{$64.04\!\rightarrow\!\mathbf{68.92}$\\[-1pt]{\tiny\textcolor{insightaccent}{$\Delta=+4.88$}}} & \shortstack{$58.21\!\rightarrow\!\mathbf{63.86}$\\[-1pt]{\tiny\textcolor{insightaccent}{$\Delta=+5.65$}}} & \shortstack{$63.43\!\rightarrow\!\mathbf{67.25}$\\[-1pt]{\tiny\textcolor{insightaccent}{$\Delta=+3.82$}}} & \shortstack{$60.42\!\rightarrow\!\mathbf{64.73}$\\[-1pt]{\tiny\textcolor{insightaccent}{$\Delta=+4.31$}}} & \shortstack{$59.35\!\rightarrow\!\mathbf{63.54}$\\[-1pt]{\tiny\textcolor{insightaccent}{$\Delta=+4.20$}}} & \textcolor{insightaccent}{6.2\%} \\
\specialrule{0.1em}{0pt}{0pt}
\end{tabularx}
\end{table*}

\subsection{OOD: Trajectory Information for Detection}

We begin by evaluating whether trajectory information consistently complements
final-state OOD detectors across architectures, benchmarks, and detector
families. Figure~\ref{fig:results-at-glance} summarises the answer. Across the
152 non-saturated cases, mean FPR95 falls by $4.33$ points with
Mahalanobis++, $8.67$ with Relative Mahalanobis, $2.69$ with
X-Mahalanobis, and $8.75$ with kNN, corresponding to 131, 142, 106, and
127 improved cases, respectively. Mahalanobis++ improves every CIFAR
comparison, while 28/38 ImageNet-200 and 27/38 ImageNet-1K
architecture--benchmark pairs improve. AUROC and FPR95 remain strongly aligned
(Spearman $\rho=0.87$), indicating that the gains are not specific to a single
detector or evaluation metric.

Beyond these overall improvements,
Figure~\ref{fig:results-at-glance}(c) reveals that the dominant pattern is an
architecture--score interaction rather than a simple ordering by family or
depth. DeiT3-B and ViT-B AugReg both contain 12 transformer blocks, yet their
kNN reductions are $20.95$ and $10.49$, while their X-Mahalanobis reductions
are $+5.78$ and $-0.80$. Swin-B is similarly detector dependent: kNN worsens by
$0.84$ points on average, whereas X-Mahalanobis improves all four benchmarks by
$3.30$. Among CNNs, ResNet-50 improves every benchmark for every detector and
is strongest with Relative Mahalanobis ($+17.03$), whereas ConvNeXt-B gains
only $+4.55$ and improves 3/4 cases. These contrasts suggest that trajectory
information depends on how intermediate computation is organised rather than on
architectural family or depth alone.

The foundation models reinforce this picture. DINOv2 together with the two
CLIP checkpoints improve all 12 benchmark combinations with both kNN
($+13.01$) and Relative Mahalanobis ($+11.03$). Strong final-state
representations learned through large-scale self-supervised and vision-language
pretraining therefore retain complementary trajectory information.

The detector-specific behaviour helps explain these differences.
X-Mahalanobis already pools multiple intermediate layers, so its smaller
marginal gain is consistent with partial overlap with trajectory information.
Relative Mahalanobis and kNN instead correct global or local final-state
density, leaving more complementary trajectory information available.
Baseline FPR95 and reduction correlate only moderately
($\rho=0.30$--$0.49$ across detectors), showing that the improvements are not
simply a consequence of larger baseline headroom.
Figure~\ref{fig:results-at-glance} therefore focuses on detectors for which an
identical final-state/trajectory pairing is well defined; MSP is reported in
the appendix as a conventional confidence reference.

\begin{table}[tb]
\centering
\caption{\textbf{Near versus Far OOD.} Balanced mean FPR95 over the same
152 cases as Figure~\ref{fig:results-at-glance}. Cells report final state
$\rightarrow$ final state + trajectory, mean $\Delta$, and improved cases;
bold marks the larger reduction within each score.}
\label{tab:ood-nearfar}
\scriptsize
\renewcommand{\arraystretch}{1.08}
\begin{tabularx}{\linewidth}{X*{2}{>{\centering\arraybackslash}X}}
\specialrule{0.1em}{0pt}{0pt}
\rowcolor{headerblue}
\textbf{Score} & \cellcolor{nearblue}\textbf{Near OOD} &
\cellcolor{farblue}\textbf{Far OOD} \\
\midrule
\rowcolor{tablebluealt}
Maha++ & \shortstack{$42.9\!\rightarrow\!40.8$\\[-1pt]{\tiny$\Delta=\textcolor{insightaccent}{+2.09}$; 101/152}} & \shortstack{$24.8\!\rightarrow\!19.2$\\[-1pt]{\tiny$\Delta=\textcolor{insightaccent}{\mathbf{+5.60}}$; 141/152}} \\
\rowcolor{tableblue}
RelMaha & \shortstack{$43.6\!\rightarrow\!38.0$\\[-1pt]{\tiny$\Delta=\textcolor{insightaccent}{+5.66}$; 119/152}} & \shortstack{$24.1\!\rightarrow\!13.7$\\[-1pt]{\tiny$\Delta=\textcolor{insightaccent}{\mathbf{+10.43}}$; 151/152}} \\
\rowcolor{tablebluealt}
X-Maha & \shortstack{$46.2\!\rightarrow\!45.5$\\[-1pt]{\tiny$\Delta=\textcolor{insightaccent}{+0.64}$; 74/152}} & \shortstack{$22.4\!\rightarrow\!18.6$\\[-1pt]{\tiny$\Delta=\textcolor{insightaccent}{\mathbf{+3.79}}$; 121/152}} \\
\rowcolor{tableblue}
kNN & \shortstack{$51.9\!\rightarrow\!46.6$\\[-1pt]{\tiny$\Delta=\textcolor{insightaccent}{+5.31}$; 117/152}} & \shortstack{$29.2\!\rightarrow\!18.5$\\[-1pt]{\tiny$\Delta=\textcolor{insightaccent}{\mathbf{+10.63}}$; 134/152}} \\
\specialrule{0.1em}{0pt}{0pt}
\end{tabularx}
\vspace{-1em}
\end{table}


To understand where trajectory information contributes most,
Table~\ref{tab:ood-nearfar} separates Near and Far OOD. Every detector
improves more on Far OOD, with Far/Near reduction ratios ranging from
$1.84\times$ for Relative Mahalanobis to $5.97\times$ for X-Mahalanobis. This
cannot be explained by additional baseline headroom, since Far-OOD FPR95 is
already substantially lower than Near-OOD FPR95 for every detector. Instead,
far shifts alter both low-level evidence and semantic destination, producing
trajectories that depart more visibly from labelled ID trajectories. Near-OOD 
examples are often confidently assigned to a plausible
in-distribution class, so they follow that class's characteristic trajectory
despite being semantically novel. The additional signal is correspondingly
smaller, but it does not disappear: Mahalanobis++, Relative Mahalanobis, and
kNN all still improve most Near cases, and only X-Mahalanobis is neutral
there ($+0.64$, 74/152), consistent with its multi-layer coverage already
capturing much of the weaker Near-OOD signal. The shortfalls instead
concentrate on the hardest ImageNet near splits.

\begin{figure}[h]
\centering
\includegraphics[width=\columnwidth]{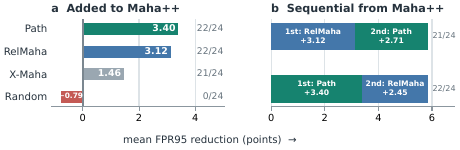}
\caption{\textbf{Path evidence is complementary.}
It gives the largest gain over Maha++ (a), and remains useful with RelMaha in either order (b). Counts report improvements out of 24 cases.}
\label{fig:ood-complementarity}
\vspace{-1em}
\end{figure}

To isolate the contribution of trajectory information,
Figure~\ref{fig:ood-complementarity} compares it against alternative
complements to Mahalanobis++. Added individually, trajectory information
provides the largest average reduction ($3.40$ FPR95 points), exceeding
Relative Mahalanobis ($3.12$), X-Mahalanobis ($1.46$), while random scores
provide no benefit ($-0.79$). The sequential analysis then asks whether this
gain disappears once Relative Mahalanobis is already present. It does not.
Adding trajectory information after Relative Mahalanobis still reduces FPR95
by a further $2.71$ points (21/24 cases), while adding Relative Mahalanobis
after trajectory information contributes an additional $2.45$ points (22/24
cases). The two signals therefore remain complementary regardless of order,
showing that trajectory information captures reliability cues that are not
absorbed by strong final-state detectors. Across the full benchmark suite,
trajectory information lowers FPR95 in 131/152 non-saturated cases and every
CIFAR comparison.

\subsection{Recognition: Trajectory Information for Classification}

Image classification provides a complementary test of whether trajectory
information extends beyond the final representation.
Table~\ref{tab:clean-classification} reports the complete evaluation across
architectures and datasets. Fusing trajectory information improves 71/72
model--dataset cases by $+4.41$ accuracy points on average. The largest dataset
means occur on Cars ($+9.81$), Flowers ($+9.04$), and Aircraft ($+7.28$),
where subordinate classes share broad object structure. This suggests that
final representations preserve category semantics while compressing part- and
texture-level cues that remain linearly accessible along the trajectory.
Baseline error and gain have Spearman $\rho=0.65$ over the 72 cells ($0.73$
between dataset means), indicating that harder tasks benefit more, although
difficulty alone does not determine the improvement.

The architecture analysis strengthens the result: all six backbones improve,
with mean gains ranging from $+1.43$ to $+6.41$ points. The two plain ViTs lead
with $+6.41$ and $+6.28$, followed by ResNet-50 ($+4.68$), Swin ($+4.61$),
ConvNeXt ($+3.03$), and CLIP ($+1.43$). The low correlation between baseline
headroom and gain ($\rho=0.37$ across model means) shows that these improvements
reflect architecture-specific use of information across depth, rather than
only easier recovery on weaker baselines. Even CLIP improves despite its high
mean final-state accuracy of $84.65$. A matched state-probe ablation provides a complementary mechanistic view
without changing this task-level result. Residual coordinates contribute a
further $+0.63$ to $+1.39$ points for both plain ViTs, CLIP, and ResNet-50,
while Swin and ConvNeXt remain within $0.25$ points of their matched
state-based variants. Thus every evaluated architecture benefits from
information distributed across depth, and four of six obtain an additional
advantage from representing the layer-to-layer updates explicitly.

Taken together with the OOD analysis, these results show that no single
architecture extracts trajectory information most effectively for every
objective. Plain ViTs obtain the largest clean-recognition gains, whereas
ResNet-50 remains consistently trajectory-sensitive across both tasks.
Conversely, CLIP has relatively little clean-recognition headroom yet belongs
to the foundation-model group with the strongest OOD improvements. Trajectory
information therefore complements different aspects of the final representation
depending on the downstream objective.

\subsection{Classification under shift}

\begin{figure}[t]
\centering
\includegraphics[width=\columnwidth]{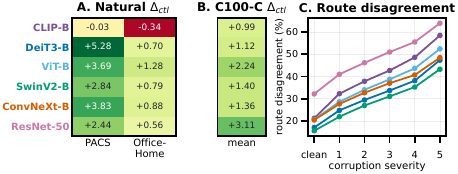}
\caption{\textbf{Trajectory information under label-preserving shift remains architecture dependent.}
For the same six checkpoints, (a--b) report gain beyond a matched final-state
ensemble and (c) reports disagreement by severity. Full natural and synthetic
shift tables are in Appendices~\ref{app:domain-shift} and
\ref{app:classification-shift}.}
\label{fig:shift-generalization}
\vspace{-1em}
\end{figure}

Distribution shift provides a complementary test of whether trajectory
information generalises beyond clean recognition.
Figure~\ref{fig:shift-generalization} shows that it largely does. On
CIFAR-100-C, fusing trajectory information improves 169/180 conditions, and
every checkpoint has a positive control-adjusted mean. ResNet-50 is strongest
($+3.11$), while ViT-B still gains $+2.24$, showing that the effect is not
CNN-specific. PACS and Office-Home retain gains of $+3.01$ and $+0.65$ beyond
the matched control across 39/48 positive folds. CLIP marks the boundary on
natural shift, whereas synthetic failures are concentrated on additive noise
and severe contrast. Trajectory information therefore generalises beyond clean
recognition, although the magnitude of the gain depends on architecture, shift
type, and final-state headroom. Complete per-model and per-domain results
appear in Appendices~\ref{app:domain-shift} and
\ref{app:classification-shift}.


\section{Interpretation and Limitations}
\label{sec:discussion}

\paragraph{Three reliability objects.}
The experiments are most coherent when final representations, confidence
estimates, and trajectories are not forced into one role. Final representations
concentrate class information and remain the primary recogniser. Confidence
scores are better suited to identifying the model's own errors. Trajectories
instead measure computational typicality, namely whether an input evolves like
labelled ID examples assigned to the same class. This explains why trajectory
information is particularly effective for OOD detection while contributing
different information for recognition.

\paragraph{Checkpoint evidence and practical limitations.}
Cross-family comparisons should be interpreted as replicated checkpoint
evidence rather than causal architecture interventions because checkpoints
differ in pretraining, scale, and optimisation. The OOD branch combines
trajectory and final-state scores using a fixed weight without OOD validation,
whereas the classification branch relies on labelled source data and is
therefore not zero-shot. Recording intermediate representations introduces a
modest computational cost of only $0.29$--$0.36$ ms per image and
$35.6$--$82.3$ MiB at batch size one across six backbones
(Appendix~\ref{app:deployment}).


\paragraph{What the path does not create.}
Because a frozen network is deterministic, its trajectory cannot manufacture
distributional information absent from earlier states. It can only retain and
organise evidence that later blocks compress. The practical result is therefore
not that paths always contain more usable information, but that a final state can
discard evidence useful for reliability.


\section{Conclusion}

In this work, we propose a new way to think about representations in vision
models: vision models expose final representations but compute trajectories
that provide complementary evidence. Controls reveal coherent,
architecture-dependent routes rather than a temporal process. Trajectory
information improves OOD detection and classification on clean and shifted
images, although its benefit depends on the architecture and the shift
encountered. Final representations carry content, confidence conveys
uncertainty, and trajectories indicate whether computation follows a familiar
route. Thus, reliable vision should ask both \emph{what representation was
produced?} and \emph{how did the model arrive there?}

\FloatBarrier
\begingroup
\small
\setlength{\bibsep}{0pt plus 0.2ex}
\bibliography{references}
\endgroup

\ifnum\IncludeAppendix=1
\clearpage

\appendix

\counterwithin{figure}{section}
\counterwithin{table}{section}
\renewcommand\thefigure{\thesection\arabic{figure}}
\renewcommand\thetable{\thesection\arabic{table}}

\raggedbottom
\renewcommand{\dbltopfraction}{0.95}
\renewcommand{\dblfloatpagefraction}{0.75}
\setcounter{dbltopnumber}{3}
\section{Appendix Roadmap}
\label{app:roadmap}

The appendix follows the paper's evidence chain and remains in the AAAI
two-column format throughout.  Section links below separate empirical audits
from reproduction details while keeping the order of the main claims.

{\small
\paragraph{Scope and extensions.}
Appendix~\ref{app:limitations-future} defines the validated global-readout
scope and develops local patch and spatial trajectories as a complementary
future extension.

\paragraph{Models and reproduction.}
Appendix~\ref{app:model-inventory} lists every frozen checkpoint by exact
\texttt{timm}/Hugging Face identifier. Appendix~\ref{app:datasets} maps every
recognition, distribution-shift, and OOD dataset to the visual family and
evaluation role it represents. Appendix~\ref{app:optimization} gives learning
rates, optimizers, splits, seeds, stopping, calibration, and ID-only selection.
Appendix~\ref{app:deployment} gives executable pseudocode and a fresh
per-architecture latency/memory audit. Appendix~\ref{app:details} fixes metrics,
uncertainty, recognition controls, qualitative layer examples,
architecture-wise layer attributions, the dataset-conditioned depth audit,
hardware, and result provenance.

\paragraph{Formal scope.}
Appendix~\ref{app:path-theory} collects the minimal information, geometry, and
class-scatter identities needed by the paper.  Each consequence is paired with
an existing empirical control, and the section explicitly separates what the
identities permit from what the experiments establish.

\paragraph{Mechanism and ablations.}
Appendix~\ref{app:continuity-controls} reports every paired, shuffled-identity,
history, and layer-gap control behind Fig.~\ref{fig:path-coherence}.
Appendix~\ref{app:ablations} compares raw, globally centred, final-state-only, and
class-route residual coordinates, then varies transition context, CNN/Swin
stage alignment, predictor family, and prediction direction.
Appendix~\ref{app:geometry-audit} separates native-block movement from
relational rewrite, and Appendix~\ref{app:architecture-breadth} tests whether
depth profiles recur across model scale and evaluation dataset.

\paragraph{Reliability evidence.}
Appendix~\ref{app:complementarity} holds the Mahalanobis++ base and the fusion
rule fixed and varies only the added score, comparing path surprise against
Relative Mahalanobis, X-Mahalanobis, and random noise, then adding Relative
Mahalanobis and path surprise in both orders. Appendices
\ref{app:domain-shift} and \ref{app:classification-shift} contain every
held-out natural-domain and synthetic-corruption result. Finally,
Appendix~\ref{app:ood-full} moves from a layer-wise view of ID/near/far
neighbourhoods to the exhaustive gain distribution and complete
checkpoint-level AUROC/FPR95 tables.}

\insightbox{Suggested reading paths}{For mechanism verification: formal scope
$\rightarrow$ continuity
$\rightarrow$ complementarity $\rightarrow$ architecture breadth
$\rightarrow$ shifted classification $\rightarrow$ complete OOD results. For
reproduction: checkpoint inventory $\rightarrow$ dataset family map
$\rightarrow$ optimization setup $\rightarrow$ deployment cost
$\rightarrow$ protocol and provenance.}

\begin{table}[t]
\centering
\noindent\textbf{Section index}\par\smallskip
{\footnotesize
\renewcommand{\arraystretch}{1.10}
\begin{tabularx}{\dimexpr\linewidth-2pt\relax}{p{7.5mm}X}
\toprule
\rowcolor{headerblue}\textbf{App.} & \textbf{Question answered} \\
\midrule
\rowcolor{tableblue}\ref{app:roadmap} & How is the evidence organised? \\
\rowcolor{tablebluealt}\ref{app:limitations-future} & What does the global study establish, and how can local paths extend it? \\
\rowcolor{tableblue}\ref{app:model-inventory} & Which 42 checkpoints are evaluated? \\
\rowcolor{tablebluealt}\ref{app:datasets} & Which datasets and visual or shift families are evaluated? \\
\rowcolor{tableblue}\ref{app:optimization} & How are fitting, selection, and calibration separated? \\
\rowcolor{tablebluealt}\ref{app:path-theory} & Why can a path matter, and what does the formalism not prove? \\
\rowcolor{tableblue}\ref{app:deployment} & What are the exact test-time rules and deployment costs? \\
\rowcolor{tablebluealt}\ref{app:complementarity} & Does the gain come from trajectory information or from adding any second score? \\
\rowcolor{tableblue}\ref{app:details} & Which metrics, controls, and layer-wise recognition audits are used? \\
\rowcolor{tablebluealt}\ref{app:continuity-controls} & Is continuity sample-specific across identity, history, and gap? \\
\rowcolor{tableblue}\ref{app:domain-shift} & Does classification transfer across natural domains? \\
\rowcolor{tablebluealt}\ref{app:ablations} & Which representation, context, map, and predictor choices matter? \\
\rowcolor{tableblue}\ref{app:geometry-audit} & Is coordinate movement distinct from relational rewrite? \\
\rowcolor{tablebluealt}\ref{app:architecture-breadth} & Do schedules recur across scale, datasets, and interventions? \\
\rowcolor{tableblue}\ref{app:classification-shift} & What happens under every CIFAR-100-C corruption? \\
\rowcolor{tablebluealt}\ref{app:ood-full} & Where are the complete checkpoint-level MSP, kNN, and Mahalanobis-family results? \\
\bottomrule
\end{tabularx}}
\end{table}
\section{Scope, Limitations, and Future Directions}
\label{app:limitations-future}

\paragraph{A deliberate global scope.}
The experiments retain one image-level vector at each selected depth.  For
token-based encoders this is the checkpoint's native global readout---usually
the \textsc{cls} token, and a mean- or attention-pooled readout when the model
does not use one.  Hierarchical Transformers and CNNs contribute globally
pooled stage features.  This is a useful common abstraction for the two tasks
studied here: both OOD detection and classification require an image-level
decision, and the same readout can be evaluated across diverse architectures.
It also enables the low-overhead implementation in
Appendix~\ref{app:deployment}.  The broad gains in the paper therefore
establish that \emph{global} computation paths already contain useful
complementary evidence.

\paragraph{Local features offer complementary resolution.}
Patch tokens and spatial feature maps retain where evidence occurs, so their
trajectories could complement the global route when a texture, object part,
background region, or anomaly is spatially concentrated.  A local path could
pool per-location surprise for image-level OOD detection while also producing
a spatial anomaly map.  The gains on visually disruptive shifts motivate this
extension and the focused question of whether its largest additional benefit
occurs for near OOD, localized corruptions, or dense anomaly localization.

\paragraph{An architecture-aware extension.}
Local trajectories pose an architecture-aware correspondence problem: patch
mixing, shifted windows, downsampling, and resolution changes prevent naive
index matching across depth, favoring token matching, learned transport, or
coordinate-free neighbourhood changes over raw subtraction.  A matched
extension should compare global-only, local-only, and joint scores against an
equal-capacity local final-representation control, and report localization,
image-level accuracy, latency, and memory.  Local states arise in the same
backbone pass; retaining and scoring them, rather than immediately pooling, is
the additional cost to measure.

\insightbox{Validated scope with a clear extension}{Global image-level paths
already improve both tasks across architectures, so the conclusions do not
depend on local features.  Local trajectories are the complementary next step:
they add spatial resolution while testing the same matched-control question---
whether the route contributes beyond the final representation.}

\begin{table}[H]
\centering
\caption{\textbf{Validated global scope and complementary local extension.}
The same matched-control principle carries from image-level to spatial paths.}
\label{tab:global-local-scope}
\scriptsize
\setlength{\tabcolsep}{2.6pt}
\renewcommand{\arraystretch}{1.05}
\begin{tabularx}{\linewidth}{p{17mm}XX}
\toprule
\rowcolor{headerblue}
\textbf{Axis} & \textbf{Global path (this work)} & \textbf{Local extension} \\
\midrule
\rowcolor{tableblue}
Readout & One global vector per depth & Patch tokens or spatial feature maps \\
\rowcolor{tablebluealt}
Evidence & Image-level route typicality & Spatially concentrated route evidence \\
\rowcolor{tableblue}
Output & Image score or class prediction & Image score plus anomaly map \\
\rowcolor{tablebluealt}
Alignment & Native vector or stage projection & Token matching across mixing and resolution changes \\
\rowcolor{tableblue}
Key control & Matched global final representation & Matched local final representation \\
\rowcolor{tablebluealt}
Cost & Measured low overhead after pooling & One-pass maps plus local scoring to be benchmarked \\
\bottomrule
\end{tabularx}
\end{table}

\section{Frozen Model Inventory}
\label{app:model-inventory}

The empirical grid is intentionally broader than the representative models in
the main figures. Table~\ref{tab:model-inventory} gives the exact identifier
needed to resolve every checkpoint and marks the four ImageNet-only additions.

\paragraph{Coverage logic.}
The 38 unstarred checkpoints form the protocol-matched grid used for balanced
counts. Four starred foundation encoders add current self-supervised and
vision--language coverage on ImageNet without silently changing the common
denominator. The inventory spans 14 plain or masked-image ViTs, four
hierarchical Transformers, two self-supervised encoders, 17 CNNs, and five
vision--language encoders. Each row is one public checkpoint, not a training
seed or an architecture-level replicate.

\paragraph{Native readouts.}
Plain, self-supervised, and vision--language Transformers expose every native
vision block. Swin and CNN checkpoints expose stage outputs after global
average pooling. A seed-0 Gaussian projection maps each stage to the largest
stage width before per-stage $\ell_2$ normalisation; this makes the predictor
dimensionally compatible but does not assert that native stage coordinates are
geometrically identical.

\paragraph{Identifier policy.}
Names are recorded exactly as resolved by \texttt{timm} or Hugging Face so the
evaluation set can be reconstructed without guessing a recipe, pretraining
corpus, image resolution, or fine-tuning variant. Cross-family conclusions are
therefore checkpoint evidence; controlled architecture claims are reserved for
the matched breadth audits in Appendix~\ref{app:architecture-breadth}.

\begin{strip}
\centering
\captionof{table}{\textbf{Frozen vision models (42 checkpoints).} Exact \texttt{timm}/
Hugging Face identifiers; unstarred models form the balanced 38-model grid and
$^*$ marks four ImageNet-only additions.}
\label{tab:model-inventory}
\scriptsize
\renewcommand{\arraystretch}{1.10}
\begin{minipage}[t]{0.485\textwidth}
\begin{tabularx}{\linewidth}{X}
\toprule
\rowcolor{headerblue}\textbf{Exact \texttt{timm} / Hugging Face identifier} \\
\midrule
\rowcolor{vitheader}\textbf{Plain and masked-image ViTs (14)} \\
\url{beit_base_patch16_224.in22k_ft_in22k_in1k} \\
\url{beit_large_patch16_224.in22k_ft_in22k_in1k} \\
\url{beitv2_base_patch16_224.in1k_ft_in22k_in1k} \\
\url{deit3_base_patch16_224.fb_in1k} \\
\url{deit3_base_patch16_224.fb_in22k_ft_in1k} \\
\url{deit3_large_patch16_224.fb_in1k} \\
\url{deit3_small_patch16_224.fb_in1k} \\
\url{eva02_base_patch14_224.mim_in22k} \\
\url{vit_base_patch16_224.mae} \\
\url{vit_base_patch16_224.augreg_in21k_ft_in1k} \\
\url{vit_base_patch16_224.augreg_in1k} \\
\url{vit_base_patch32_224.augreg_in21k_ft_in1k} \\
\url{vit_large_patch16_224.augreg_in21k_ft_in1k} \\
\url{vit_small_patch16_224.augreg_in21k_ft_in1k} \\
\addlinespace[2pt]
\rowcolor{hierheader}\textbf{Hierarchical Transformers (4)} \\
\url{swin_base_patch4_window7_224.ms_in22k_ft_in1k} \\
\url{swinv2_base_window16_256.ms_in1k} \\
\url{swinv2_base_window12to16_192to256.ms_in22k_ft_in1k} \\
\url{swinv2_tiny_window16_256.ms_in1k} \\
\addlinespace[2pt]
\rowcolor{vlmheader}\textbf{Self-supervised foundation encoders (2)} \\
\url{vit_base_patch14_dinov2.lvd142m} \\
\url{facebook/dinov3-vitb16-pretrain-lvd1689m}$^*$ \\
\bottomrule
\end{tabularx}
\end{minipage}\hfill
\begin{minipage}[t]{0.485\textwidth}
\begin{tabularx}{\linewidth}{X}
\toprule
\rowcolor{headerblue}\textbf{Exact \texttt{timm} / Hugging Face identifier} \\
\midrule
\rowcolor{cnnheader}\textbf{Convolutional encoders (17)} \\
\url{convnext_base.fb_in1k} \\
\url{convnext_base.fb_in22k_ft_in1k} \\
\url{convnext_large.fb_in1k} \\
\url{convnext_tiny.fb_in1k} \\
\url{convnextv2_base.fcmae_ft_in1k} \\
\url{convnextv2_base.fcmae_ft_in22k_in1k} \\
\url{convnextv2_tiny.fcmae_ft_in1k} \\
\url{densenet121.ra_in1k} \\
\url{tf_efficientnetv2_l.in21k_ft_in1k} \\
\url{tf_efficientnetv2_m.in21k_ft_in1k} \\
\url{tf_efficientnetv2_s.in21k_ft_in1k} \\
\url{regnety_040.pycls_in1k} \\
\url{resnet101.a1h_in1k} \\
\url{resnet152.a1h_in1k} \\
\url{resnet50.a1_in1k} \\
\url{resnetv2_50.a1h_in1k} \\
\url{wide_resnet50_2.racm_in1k} \\
\addlinespace[2pt]
\rowcolor{vlmheader}\textbf{Vision--language encoders (5)} \\
\url{vit_large_patch14_clip_224.openai} \\
\url{vit_base_patch16_clip_224.openai} \\
\url{openai/clip-vit-base-patch16}$^*$ \\
\url{PE-Core-B16-224}$^*$ \\
\url{google/siglip2-base-patch16-512}$^*$ \\
\bottomrule
\end{tabularx}
\end{minipage}

{\scriptsize
\textbf{Architecture sources.}
Plain and masked-image Vision Transformers include
ViT~\citep{dosovitskiy2021vit} (including the AugReg training
recipe~\citep{steiner2022augreg}),
MAE~\citep{he2022mae},
BEiT/BEiT-v2~\citep{bao2022beit,peng2022beitv2},
DeiT~III~\citep{touvron2022deit3},
and EVA-02~\citep{fang2023eva02}.
Hierarchical Transformers comprise
Swin/Swin~V2~\citep{liu2021swin,liu2022swinv2}.
Convolutional architectures include
ResNet, ResNet~V2, and WideResNet
~\citep{he2016resnet,he2016identity,zagoruyko2016wide},
DenseNet~\citep{huang2017densenet},
EfficientNet~V2~\citep{tan2021efficientnetv2},
RegNet~\citep{radosavovic2020regnet},
and ConvNeXt/ConvNeXt~V2
~\citep{liu2022convnext,woo2023convnextv2}.
Foundation encoders include
DINOv2/DINOv3~\citep{oquab2024dinov2,simeoni2025dinov3},
CLIP~\citep{radford2021clip},
Perception Encoder (PE-Core)~\citep{bolya2025pe},
and SigLIP~2~\citep{tschannen2025siglip2}.}

\refstepcounter{section}
\label{app:datasets}
{\large\bfseries \thesection\quad Datasets and Evaluation Families\par}
\vspace{1.5mm}
\parbox{0.96\textwidth}{\small
The suite is heterogeneous by design. Table~\ref{tab:dataset-families} names
the visual or shift family contributed by each dataset, rather than treating
all accuracy or OOD benchmarks as interchangeable. A dataset may legitimately
serve more than one role: DTD tests clean texture recognition and also supplies
an ImageNet far-OOD texture shift, while CIFAR-10 and CIFAR-100 exchange ID and
near-OOD roles. Near/far assignments follow the OpenOOD v1.5 protocol
\citep{yang2022openood,zhang2023openood}; ``family'' describes the intended
evaluation axis, not a claim that it is the only factor present in the images.}
\vspace{2mm}
\captionof{table}{\textbf{Datasets and the evaluation families they represent.}
The recognition rows cover all 12 clean datasets; the remaining rows make the
natural-shift, synthetic-shift, and OpenOOD roles explicit.}
\label{tab:dataset-families}
\scriptsize
\setlength{\tabcolsep}{4pt}
\renewcommand{\arraystretch}{1.08}
\begin{tabularx}{0.98\textwidth}{p{29mm}p{63mm}X}
\toprule
\rowcolor{headerblue}
\textbf{Evaluation role} & \textbf{Datasets / prescribed splits} & \textbf{Family represented and diagnostic purpose} \\
\midrule
\rowcolor{vitheader}\multicolumn{3}{l}{\textbf{Clean recognition}} \\
\rowcolor{tablebluealt}
Broad object categories & CIFAR-10, CIFAR-100, Caltech-101 & Coarse-to-medium-grained natural objects; tests general class transfer across label-set size. \\
\rowcolor{tableblue}
Fine-grained objects & CUB-200, Stanford Cars, Oxford-IIIT Pets, Flowers-102, FGVC-Aircraft & Subordinate categories with shared global shape; stresses transient part, texture, and attribute cues. \\
\rowcolor{tablebluealt}
Appearance-centric labels & DTD, Food-101 & Texture/material attributes and visually defined food categories; tests local appearance evidence. \\
\rowcolor{tableblue}
Scenes and environment & SUN397, EuroSAT & Scene layout and remote-sensing land use; extends beyond centred object recognition. \\
\addlinespace[1pt]
\rowcolor{hierheader}\multicolumn{3}{l}{\textbf{Label-preserving distribution shift}} \\
\rowcolor{tablebluealt}
Natural domain shift & PACS; Office-Home & Photo/art/cartoon/sketch and art/clipart/product/real-domain changes with the class vocabulary fixed. \\
\rowcolor{tableblue}
Synthetic corruption & CIFAR-100-C & Six matched corruption types across noise, blur, weather, and digital families at five severities; labels remain fixed. \\
\addlinespace[1pt]
\rowcolor{vlmheader}\multicolumn{3}{l}{\textbf{OpenOOD detection}} \\
\rowcolor{tablebluealt}
ID benchmarks & MNIST; CIFAR-10; CIFAR-100; ImageNet-200; ImageNet-1K & Digits, coarse objects, and medium/full-scale object taxonomies; tests scale and semantic granularity. \\
\rowcolor{tableblue}
CIFAR near OOD & Other CIFAR dataset; Tiny ImageNet & Semantically adjacent natural objects and a broader small-image object vocabulary. \\
\rowcolor{tablebluealt}
CIFAR far OOD & MNIST, SVHN, Textures, Places365 & Digits, street numbers, texture, and scenes; changes visual statistics and semantic destination. \\
\rowcolor{tableblue}
MNIST near / far OOD & notMNIST, FashionMNIST / Textures, CIFAR-10, Tiny ImageNet, Places365 & Glyph/apparel neighbours versus natural-image, texture, object, and scene shifts. \\
\rowcolor{tablebluealt}
ImageNet near OOD & SSB-hard, NINCO & Semantically difficult or natural-object novelty close to the ImageNet support. \\
\rowcolor{tableblue}
ImageNet far OOD & iNaturalist, DTD, OpenImage-O & Fine-grained species, texture/style, and novel-object families; the same five splits are used for ImageNet-200 and ImageNet-1K. \\
\bottomrule
\end{tabularx}

\vspace{3mm}
\captionof{table}{\textbf{Complete learned-component setup.} ``Selection'' always
means held-out ID or source validation; no OOD, corruption, or target-domain
sample chooses a hyperparameter.}
\label{tab:optimization}
\setlength{\tabcolsep}{4pt}
\renewcommand{\arraystretch}{1.05}
\begin{tabularx}{0.98\textwidth}{p{30mm}Xp{36mm}}
\toprule
\rowcolor{headerblue}
\textbf{Component} & \textbf{Training configuration} & \textbf{Selection} \\
\midrule
\rowcolor{tableblue}
Path-surprise MLP & AdamW; lr $3\!\times\!10^{-4}$; wd $10^{-5}$; 80 epochs; batch 512; hidden 1024 & fixed recipe; seeds 0, 1, 2 \\
\rowcolor{tablebluealt}
Recognition probes & AdamW; lr $3\!\times\!10^{-3}$; wd grid $10^{-4}$, $10^{-3}$, $10^{-2}$, $10^{-1}$; at most 90 epochs; batch 2048; patience 10 & inner source split; seeds 0, 1, 2 \\
\rowcolor{tableblue}
Direction screen & AdamW; lr $3\!\times\!10^{-4}$; wd $10^{-4}$; at most 22 epochs; batch 512; patience 5 & ID validation only \\
\rowcolor{tablebluealt}
Continuity controls & Closed-form ridge regression, $\lambda=1$; one shared 64-D projection & held-out ID rows \\
\rowcolor{tableblue}
Temperatures / fusion & Bounded scalar and L-BFGS-B solves; weights $[0,3]$; gate quantiles .10, .20, .35, .50, .75, 1 & disjoint fusion validation \\
\bottomrule
\end{tabularx}
\end{strip}
\section{Optimization and Calibration Setup}
\label{app:optimization}

\paragraph{No encoder is trained.}
Every checkpoint in Table~\ref{tab:model-inventory} remains frozen. Learning is
restricted to small transition predictors and linear recognition probes on
cached features. Final state densities, class means, covariance estimates,
standardisation, and score residualisation are fitted in closed form. Thus a
``training epoch'' below never updates a public vision backbone.

\paragraph{Path-surprise predictor.}
For each transition, the predictor is a two-hidden-layer MLP with GELU
activations and a learned depth embedding. It predicts the next class-residual
state by mean-squared error. For component $d$ at transition $l$, the reported
$\sigma^2_{l,d}$ is exactly the mean squared prediction residual on ID fitting
data plus $10^{-6}$. The score divides by this quantity, sums across dimensions
and transitions, and is added at the globally fixed weight $0.3$ after ID-only
linear residualisation against the baseline detector. Score evaluation uses
batches of 256.

\paragraph{Support and calibration sizes.}
The manifests use separate fit/calibration examples per class. MNIST uses
$1000/300$, CIFAR-100 uses $200/30$, and the ImageNet benchmarks use $50/30$.
CIFAR-10 uses $1000/300$ except for the four IN21k-fine-tuned rows (DeiT3-B,
ConvNeXt-B, ConvNeXtV2-B, and SwinV2-B), whose original matched manifests use
$2000/300$. Every comparison holds this support fixed between the final
representation and +path arms. Three independently sampled seeds are reported.
The recognition branch standardises each probe input using probe-training rows,
selects weight decay and stopping epoch on an inner split, refits on the full
probe pool, and reserves a disjoint split for temperatures, non-negative
weights, candidate depth sets, and the final-state margin gate. Five-fold
out-of-fold selection is used when choosing residual-probe ensembles.

With the training and selection protocol fixed,
Appendix~\ref{app:path-theory} states the minimal formal conditions under which
the path can matter and the controls required to support each interpretation.

\section{Why Intermediate Paths Can Matter}
\label{app:path-theory}

\paragraph{Purpose and status.}
This section records only the formal consequences needed to interpret the
paper's empirical claims.  They are standard information and second-moment
identities, not a claim of a new learning theory.  They establish when a
recorded path \emph{can} complement a final representation and which stronger
interpretations require additional controls; they do not guarantee that the
finite predictor in Section~\ref{sec:method-ood} will recover the available
signal.

\subsection{Information can be retained, not created}

Let $R=(Z_1,\ldots,Z_{L-1})$ denote the recorded intermediate states and
$E=Z_L$ the final state.  Equation~\ref{eq:chain} decomposes the joint ID--OOD
divergence into final-state divergence and a non-negative conditional-route
term.  The same observation has an operational form.  Because $E$ is a
projection of $(R,E)$,
\begin{equation}
 \operatorname{TV}(P_{R,E},Q_{R,E})
 \geq \operatorname{TV}(P_E,Q_E),
 \qquad
 \mathcal R^*_{R,E}\leq \mathcal R^*_E,
 \label{eq:app-bayes-ordering}
\end{equation}
where $\mathcal R^*$ is the equal-prior Bayes OOD error.  An optimal rule with
the path available cannot be less informative than one restricted to the
final state.  A learned high-dimensional rule can still be worse because of
finite support, misspecification, or optimization; this is why all reported
comparisons use held-out data and a matched final-state control.

The path does not manufacture divergence.  A frozen feed-forward network is
deterministic, so data processing gives
\begin{equation}
 D_{\rm KL}(P_{Z_{l+1}}\|Q_{Z_{l+1}})
 \leq D_{\rm KL}(P_{Z_l}\|Q_{Z_l}).
 \label{eq:app-data-processing}
\end{equation}
Moreover, if recording begins at $Z_k$, the deterministic path
$Z_{k:L}$ contains $Z_k$ itself and therefore preserves its divergence:
$D_{\rm KL}(P_{Z_{k:L}}\|Q_{Z_{k:L}})
=D_{\rm KL}(P_{Z_k}\|Q_{Z_k})$ under the usual absolute-continuity and
measurability conditions.  Intermediate states may retain evidence that later
blocks compress; they cannot create evidence absent from the earlier state.

\subsection{State access is not order information}

Let $T=(Z_1,\ldots,Z_L)$ be the ordered path and let $B=b(T)$ be a
permutation-invariant bag containing the same states.  For a class or domain
variable $A$, because $B$ is a deterministic function of $T$,
\begin{equation}
 I(A;T)=I(A;B)+I(A;T\mid B).
 \label{eq:app-order-information}
\end{equation}
The last term is the information attributable specifically to order.  A fixed
global layer permutation is invertible and hence does not remove it.  Thus the
paper's fixed-permutation result rules out sensitivity to the literal forward
index, but it cannot establish that an ordered predictor beats the same states
given to a capacity-matched set encoder.  Identity shuffling answers a
different question: it destroys same-image continuity while preserving layer
marginals.  These distinctions justify the deliberately narrower
``cross-depth continuity'' claim in Section~\ref{sec:method-recognition}.

\subsection{Movement, innovation, and class separation}

For a common-coordinate block update $U_l=Z_{l+1}-Z_l$, define
$M_l(Y)=\mathbb E[U_l\mid Y]$ and $V_l=U_l-M_l(Y)$.  Conditional expectation is
an orthogonal projection in $L^2$, so
\begin{equation}
 \mathbb E\lVert U_l\rVert^2
 =\mathbb E\lVert M_l(Y)\rVert^2
 +\mathbb E\lVert V_l\rVert^2.
 \label{eq:app-update-energy}
\end{equation}
This is the population squared-energy version of the class-transport and
innovation decomposition in Equation~\ref{eq:decomposition}.  It is exact for
RMS energy, not for a mean unsquared norm, and it requires aligned coordinates;
the paper switches to relational measurements at width-changing transitions.

Movement still need not reorganize the sample geometry.  If the rows of
$H\in\mathbb R^{n\times d}$ are centered sample states, $U$ is the centered
update, and $K=HH^\top$, then
\begin{equation}
 K^+-K=HU^\top+UH^\top+UU^\top.
 \label{eq:app-gram-update}
\end{equation}
The cross terms depend on update orientation, not only on $\lVert U\rVert_F$.
For example, $H^+=HR$ with $R^\top R=I$ can move every coordinate while
leaving $K$ and linear CKA unchanged.  This is the exact reason movement and
CKA deformation are reported separately in
Appendix~\ref{app:geometry-audit}.

Nor does a large update necessarily improve classification.  Let $S_B$ and
$S_W$ be between- and within-class scatter, $a_c=\mu_c-\mu$,
$b_c=\mathbb E[U\mid c]-\mathbb E U$, $X=Z-\mu_Y$, and
$V=U-\mathbb E[U\mid Y]$.  The trace changes are
\begin{align}
 \Delta_B
 &=2\sum_c\pi_c\langle a_c,b_c\rangle
   +\sum_c\pi_c\lVert b_c\rVert^2,\nonumber\\
 \Delta_W
 &=2\,\mathbb E\langle X,V\rangle+\mathbb E\lVert V\rVert^2.
 \label{eq:app-scatter-update}
\end{align}
Consequently, the Fisher trace ratio
$J=\operatorname{tr}(S_B)/\operatorname{tr}(S_W)$ increases exactly when
$\Delta_B/\operatorname{tr}(S_B)>
\Delta_W/\operatorname{tr}(S_W)$.  Class-differential transport must build
between-class structure faster than sample-specific innovation builds
within-class variation.  This explains why movement peaks, CKA changes, and
linear-probe gains need not coincide.

\subsection{Reverse predictability is not a time arrow}

Predictive asymmetry depends on the representation and predictor family.  Let
$X\sim\mathcal N(0,1)$,
$Z_l=(X,0)$, and $Z_{l+1}=(X,M(X^2-1))$.  The computation is deterministically
forward, yet for affine predictors
\begin{equation}
 R^2_{l\rightarrow l+1}=\frac{1}{1+2M^2},
 \qquad R^2_{l+1\rightarrow l}=1.
 \label{eq:app-reverse-example}
\end{equation}
The restricted forward predictor cannot represent the quadratic component,
whereas the later state contains $X$ exactly.  Reverse-easier prediction can
therefore reflect projection or model restriction rather than causality,
irreversibility, or a learned arrow of time.  The multi-family direction audit
in Appendix~\ref{app:direction-capacity} is interpreted within this boundary.

\begin{table}[H]
\centering
\caption{\textbf{Formal consequence to empirical test.} Each identity narrows
the claim supported by an existing control; none predicts universal gain.}
\label{tab:theory-test-bridge}
\scriptsize
\setlength{\tabcolsep}{2.8pt}
\renewcommand{\arraystretch}{1.05}
\begin{tabularx}{\linewidth}{p{18mm}XX}
\toprule
\rowcolor{headerblue}
\textbf{Consequence} & \textbf{What it permits} & \textbf{Required evidence} \\
\midrule
Conditional KL & Route evidence beyond a matched final state & Final-state-only versus final-state + path \\
Order identity & Information may reside in ordering & Capacity-matched unordered-state control \\
Gram update & Movement and relational rewrite can separate & CKA plus invariant rotation control \\
Scatter update & Updates help only if relative separation grows & Layer probes and class-wise depth profiles \\
Restricted $R^2$ & Reverse prediction need not imply reverse computation & Predictor-family and direction audit \\
\bottomrule
\end{tabularx}
\end{table}

The section therefore supports a limited but sufficient conclusion: recording
intermediate states can expose evidence unavailable to a final-state rule,
while the usefulness, ordering, and recoverability of that evidence remain
empirical properties of each architecture, task, and shift.

\section{Deployment Cost and Algorithm Summary}
\label{app:deployment}

\paragraph{Measurement boundary.}
Table~\ref{tab:deployment-cost} is a fresh audit of the method in
Equations~\ref{eq:dynamics-score}--\ref{eq:combiner}, not the older recurrent
prototype used during development. Both arms execute the checkpoint's ordinary
head in one frozen-backbone pass from an already resident CUDA image tensor.
The path arm immediately pools each selected block/stage output, so token and
spatial maps are never retained. It then performs the seed-0 common-width stage
map, 1,000-class route assignment, and the current two-hidden-layer, width-1024
MLP score in FP32; backbone inference uses FP16 autocast. Image decoding and
CPU preprocessing are excluded.

\begin{table}[H]
\centering
\caption{\textbf{Current-method deployment cost.} Batch-1 baseline
$\rightarrow$ baseline + path on one RTX~4090. Bold latency increments are
absolute ms/image; memory increments are MiB.}
\label{tab:deployment-cost}
\scriptsize
\setlength{\tabcolsep}{2.2pt}
\renewcommand{\arraystretch}{1.06}
\begin{tabularx}{\linewidth}{@{}Xcrr@{}}
\specialrule{0.1em}{0pt}{0pt}
\rowcolor{headerblue}
\textbf{Backbone} & \textbf{States} & \textbf{Latency (ms/img)} & \textbf{Peak (MiB)} \\
\midrule
\rowcolor{tablebluealt}
CLIP-B       & $12{\times}768$ & $1.49\!\rightarrow\!1.78$ $(\mathbf{+0.29})$ & $348.1\!\rightarrow\!399.9$ $(+51.8)$ \\
\rowcolor{tableblue}
DeiT3-B      & $12{\times}768$ & $1.56\!\rightarrow\!1.89$ $(\mathbf{+0.33})$ & $349.7\!\rightarrow\!401.6$ $(+51.8)$ \\
\rowcolor{tablebluealt}
ViT-B AugReg & $12{\times}768$ & $1.50\!\rightarrow\!1.80$ $(\mathbf{+0.30})$ & $349.7\!\rightarrow\!401.5$ $(+51.8)$ \\
\rowcolor{tableblue}
Swin-B       & $4{\times}1024$ & $5.30\!\rightarrow\!5.61$ $(\mathbf{+0.31})$ & $368.8\!\rightarrow\!404.5$ $(+35.6)$ \\
\rowcolor{tablebluealt}
ConvNeXt-B   & $4{\times}1024$ & $3.10\!\rightarrow\!3.45$ $(\mathbf{+0.35})$ & $360.5\!\rightarrow\!396.3$ $(+35.8)$ \\
\rowcolor{tableblue}
ResNet-50    & $4{\times}2048$ & $1.56\!\rightarrow\!1.93$ $(\mathbf{+0.36})$ & $119.3\!\rightarrow\!201.5$ $(+82.3)$ \\
\specialrule{0.1em}{0pt}{0pt}
\end{tabularx}
\end{table}

Latency is the median of seven groups of 50 iterations after 20 warm-up
iterations. Peak allocation is \texttt{torch.cuda.max\_memory\_allocated} and
includes the resident checkpoint plus all path-only parameters, class-route
statistics, error scales, and stage maps. The absolute latency increment is
narrowly concentrated at $0.29$--$0.36$ ms/image, with $35.6$--$82.3$ MiB
additional peak allocation across the six representatives. The corresponding
percentage depends strongly on backbone speed, so Table~\ref{tab:deployment-cost}
reports added time directly rather than a denominator-sensitive ratio. Swin-B's
four-stage trajectory adds $0.31$ ms/image; ResNet-50 pays the largest memory
increment because its common stage width is 2,048. Script
\texttt{219\_benchmark\_current\_path.py} and
\texttt{deployment\_cost\_current\_path.json} contain the exact checkpoints,
software versions, raw timing groups, and measurement settings.

\begin{algorithm}[H]
\caption{Test-time path evidence with ID-fitted components}
\label{alg:path-evidence}
\begin{algorithmic}[1]
\REQUIRE States $z_{1:L}$; class paths $\mu^{c}_{1:L}$; predictors $f_l$;
error scales $\widehat\sigma^2_l$; final score $S_0$
\STATE $\widehat c\gets\arg\min_c\lVert z_L-\mu_L^c\rVert_2^2$
\FOR{$l=1,\ldots,L$}
  \STATE $r_l\gets z_l-\mu_l^{\widehat c}$
\ENDFOR
\STATE $D\gets\sum_{l,d}(r_{l+1,d}-f_l(r_l)_d)^2/\widehat\sigma^2_{l,d}$
\STATE \textbf{OOD:} $S\gets z(S_0)+0.3\,[z(D)-a\,z(S_0)-b]$
\STATE \textbf{Recognition:} $q_{\rm path}\gets\sum_lw_lq_l(y\mid z_{l+1}-z_l)$
\STATE $q_{\rm fuse}\gets(1-\alpha)q_L+\alpha q_{\rm path}$ using the fitted gate
\RETURN $S$ for OOD, or $q_{\rm fuse}$ for recognition
\end{algorithmic}
\end{algorithm}

Algorithm~\ref{alg:path-evidence} omits fitting for compactness: every displayed
component is learned, standardised, residualised, or selected on ID/source data
under Appendix~\ref{app:optimization}; no test OOD, corruption, or target-domain
sample enters that process.

\section{Complementarity Controls}
\label{app:complementarity}

We test whether the gain comes from trajectory information rather than from
adding an arbitrary second score. We keep Mahalanobis++, the ID-only
calibration, the decorrelated fusion rule, and the fixed weight $0.3$ unchanged,
and vary only the added score. The expanded audit covers 12 checkpoints on
CIFAR-10 and ImageNet-200, using three seeds and all prescribed OOD splits.

\begingroup
\setlength{\stripsep}{6pt plus 2pt minus 2pt}

\begin{strip}
\centering
\captionsetup{
  type=table,
  width=0.98\textwidth,
  justification=raggedright,
  singlelinecheck=false,
  hypcap=false
}

\caption{\textbf{Expanded partner control from Mahalanobis++.}
Mean FPR95 reduction from adding each score to the same base; positive values
are better. ``Improved'' counts checkpoint--benchmark cells with lower FPR95.}
\label{tab:partner-control}

\vspace{-1mm}
\scriptsize
\setlength{\tabcolsep}{4pt}
\renewcommand{\arraystretch}{1.08}

\begin{tabularx}{0.98\textwidth}{
p{31mm}
*{6}{>{\centering\arraybackslash}X}}
\toprule
\rowcolor{headerblue}
&
\multicolumn{2}{c}{\textbf{CIFAR-10}} &
\multicolumn{2}{c}{\textbf{ImageNet-200}} &
\multicolumn{2}{c}{\textbf{Overall}} \\
\cmidrule(lr){2-3}
\cmidrule(lr){4-5}
\cmidrule(lr){6-7}

\rowcolor{headerblue}
\textbf{Added score} &
\boldmath$\Delta$\textbf{FPR95} & \textbf{Improved} &
\boldmath$\Delta$\textbf{FPR95} & \textbf{Improved} &
\boldmath$\Delta$\textbf{FPR95} & \textbf{Improved} \\
\midrule

\rowcolor{insightfill}
Path surprise &
\textcolor{insightaccent}{\textbf{+4.75}} & \textbf{12/12} &
\textcolor{insightaccent}{\textbf{+2.05}} & 10/12 &
\textcolor{insightaccent}{\textbf{+3.40}} & \textbf{22/24} \\

\rowcolor{tableblue}
Relative Mahalanobis &
\textcolor{insightaccent}{+3.71} & 12/12 &
\textcolor{insightaccent}{+2.54} & 10/12 &
\textcolor{insightaccent}{+3.12} & 22/24 \\

\rowcolor{tablebluealt}
X-Mahalanobis &
\textcolor{insightaccent}{+2.10} & 11/12 &
\textcolor{insightaccent}{+0.82} & 10/12 &
\textcolor{insightaccent}{+1.46} & 21/24 \\

\rowcolor{tableblue}
Random noise &
$-0.51$ & 0/12 &
$-1.07$ & 0/12 &
$\mathbf{-0.79}$ & \textbf{0/24} \\
\bottomrule
\end{tabularx}

\vspace{3mm}

\caption{\textbf{Expanded sequential complementarity.}
Relative Mahalanobis and path surprise are added in both orders to the same
Mahalanobis++ base. ``First'' and ``second'' are the reductions contributed at
each step; ``total'' is the reduction after both additions.}
\label{tab:sequential-control}

\vspace{-1mm}
\begin{tabularx}{0.98\textwidth}{
p{34mm}
p{22mm}
*{5}{>{\centering\arraybackslash}X}}
\toprule
\rowcolor{headerblue}
\textbf{Addition order} &
\textbf{Benchmark} &
\textbf{First} &
\textbf{First improved} &
\textbf{Second} &
\textbf{Second improved} &
\textbf{Total} \\
\midrule

\rowcolor{tableblue}
Relative Maha $\rightarrow$ path &
CIFAR-10 &
\textcolor{insightaccent}{+3.70} & 12/12 &
\textcolor{insightaccent}{\textbf{+3.32}} & 11/12 &
\textcolor{insightaccent}{\textbf{+7.03}} \\

\rowcolor{tablebluealt}
&
ImageNet-200 &
\textcolor{insightaccent}{+2.54} & 10/12 &
\textcolor{insightaccent}{\textbf{+2.11}} & 10/12 &
\textcolor{insightaccent}{\textbf{+4.65}} \\

\rowcolor{sectionblue}
&
Overall &
\textcolor{insightaccent}{\textbf{+3.12}} & 22/24 &
\textcolor{insightaccent}{\textbf{+2.71}} & \textbf{21/24} &
\textcolor{insightaccent}{\textbf{+5.84}} \\
\midrule

\rowcolor{tableblue}
Path $\rightarrow$ Relative Maha &
CIFAR-10 &
\textcolor{insightaccent}{+4.75} & 12/12 &
\textcolor{insightaccent}{\textbf{+2.27}} & 12/12 &
\textcolor{insightaccent}{\textbf{+7.03}} \\

\rowcolor{tablebluealt}
&
ImageNet-200 &
\textcolor{insightaccent}{+2.04} & 10/12 &
\textcolor{insightaccent}{\textbf{+2.62}} & 10/12 &
\textcolor{insightaccent}{\textbf{+4.66}} \\

\rowcolor{sectionblue}
&
Overall &
\textcolor{insightaccent}{\textbf{+3.40}} & 22/24 &
\textcolor{insightaccent}{\textbf{+2.45}} & \textbf{22/24} &
\textcolor{insightaccent}{\textbf{+5.84}} \\
\bottomrule
\end{tabularx}

\end{strip}
\endgroup

Path surprise produces the largest direct reduction, lowering mean FPR95 by
$3.40$ points and improving 22/24 cells. Relative Mahalanobis reaches the same
coverage with a slightly smaller $3.12$-point reduction. X-Mahalanobis also
helps broadly, but its $1.46$-point reduction is less than half the path gain.
Random noise improves no cell and worsens both benchmarks, ruling out the
combination rule itself as the source of improvement.

The sequential control provides the stronger complementarity test. After
Relative Mahalanobis reduces FPR95 by $3.12$ points, path surprise contributes
another $2.71$ points and improves 21/24 cells. In the reverse order, path
surprise contributes $3.40$ points before Relative Mahalanobis adds another
$2.45$ points in 22/24 cells. Both orders reach the same $5.84$-point total
reduction, showing that final-representation distance and trajectory surprise
provide complementary rather than interchangeable evidence.

Having established score complementarity, Appendix~\ref{app:details} reports
the complete per-dataset OOD results and matched trajectory controls.
\section{Experimental Protocol and Result Provenance}
\label{app:details}

\subsection{Evaluation and uncertainty}

Appendix~\ref{app:optimization} gives the complete data-separation and optimizer
protocol. Path-paired OOD scores and fitted MSP/kNN references average the
prescribed OOD splits and three ID-support seeds; native-head MSP is
deterministic. AUROC
and FPR95 are percentages; larger anomaly scores indicate OOD, and positive
$\Delta$FPR95 means
$\mathrm{FPR95}_{\rm final}-\mathrm{FPR95}_{\rm path}>0$. Recognition gains
are top-1 percentage-point changes from the final state probe, while
$\Delta_{\rm ctl}$ additionally subtracts a separately seeded final state ensemble.
CIFAR-100-C intervals resample the 15 corruption families 20,000 times after
averaging severities and probe seeds within a family, rather than treating
individual images as independent.

\subsection{Recognition controls}
\label{app:recognition-controls}

The final-representation ensemble repeats the final-state probe with an independent
optimisation seed and fuses it through the same validation protocol. State
probes use $z_l$; native-update probes use $z_{l+1}-z_l$ only where coordinates
are compatible. Comparing these branches separates generic ensembling, access
to intermediate states, and the residual/update parameterisation. For
corruption experiments, clean model selection is frozen before loading any
corrupted images. For PACS and Office-Home, source-domain splits are recreated
inside each held-out-domain fold so that the target domain cannot influence the
gate.

\subsection{What becomes readable across depth?}
\label{app:qualitative-layer-probes}

Aggregate accuracy does not show how class evidence changes for one image.
Figure~\ref{fig:layer-probe-examples} therefore follows six real DTD test
images through DeiT3-B. DTD is useful for this diagnostic because its labels
name visible attributes such as \emph{gauzy}, \emph{paisley}, and
\emph{crosshatched}. Each column is an independently trained linear probe on a
frozen block output; it reports the predicted texture and the
validation-calibrated probability assigned to the true class.

The cases expose two behaviours hidden by an aggregate score. First, evidence
can sharpen across depth: for the crosshatched image, the early
\emph{spiralled} prediction becomes the correct texture after block 6. Second,
useful evidence can be compressed or reorganised before the final
representation: block 9 decodes the visible attribute correctly even when
block 12 and the post-normalisation final representation confuse related
textures. The fixed late-output vote recovers
these cases by retaining agreement that no single final-state decision exposes.
This is a diagnostic of linearly readable class evidence, not a claim that a
block reconstructs the depicted concept.

\begin{figure*}[t]
  \centering
  \includegraphics[width=0.97\textwidth]{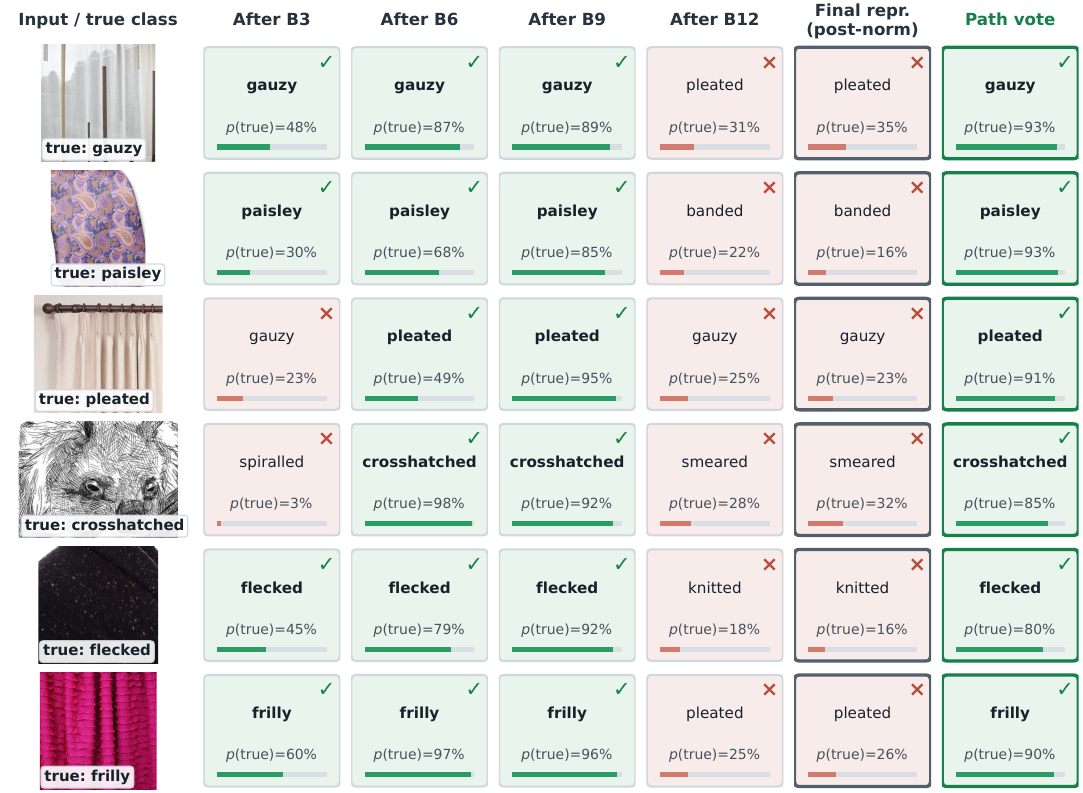}
  \caption{\textbf{Class evidence can peak before the final representation.}
  Cards probe the actual outputs of blocks 3, 6, 9, and 12; green/rose denotes
  correct/incorrect and bars give calibrated true-class probability. The final
  representation applies DeiT3's terminal LayerNorm after block 12, so those
  columns are distinct but adjacent computations. The path vote averages the
  four block outputs ranked highest on validation data (blocks 8--11). Shown cases are the
  highest-margin, distinct-class final-representation errors corrected by that fixed vote;
  they illustrate the mechanism, while the tables quantify its frequency.}
  \label{fig:layer-probe-examples}
\end{figure*}

\subsection{Which depths carry classification utility?}
\label{app:classification-layer-shapley}

The image cases show what becomes readable in one architecture; they do not
compare where architectures store classification evidence. We therefore form
an exact Shapley game over native readouts~\citep{shapley1953value}. A player is
an independently tuned linear probe on one block or stage, including the final
representation. A coalition's value is the top-1 accuracy of its equally
weighted, calibrated-logit vote. One stratified half of the CIFAR-100
fusion-validation set fits temperatures and the other half evaluates every
coalition. Enumerating all subsets gives exact marginal contributions without
accessing test predictions or labels. This measures probe-readable utility,
not a causal effect of deleting a backbone block.

Figure~\ref{fig:classification-layer-shapley} reveals different schedules.
CLIP concentrates utility in its last blocks, ViT-B grows gradually, and
DeiT3-B forms a late plateau in which several readouts are almost equally
useful. The hierarchical models are more concentrated: SwinV2-B and ConvNeXt-B
place nearly all utility in their last two stages. ResNet-50 is similar but
retains a smoother rise through the preceding stage. In every family, the
penultimate readout remains competitive with the final representation. Thus
path fusion is not exploiting one universally privileged depth; it combines
multiple late readouts whose class evidence is strong but not identical.

\begin{figure*}[t]
  \centering
  \includegraphics[width=0.985\textwidth]{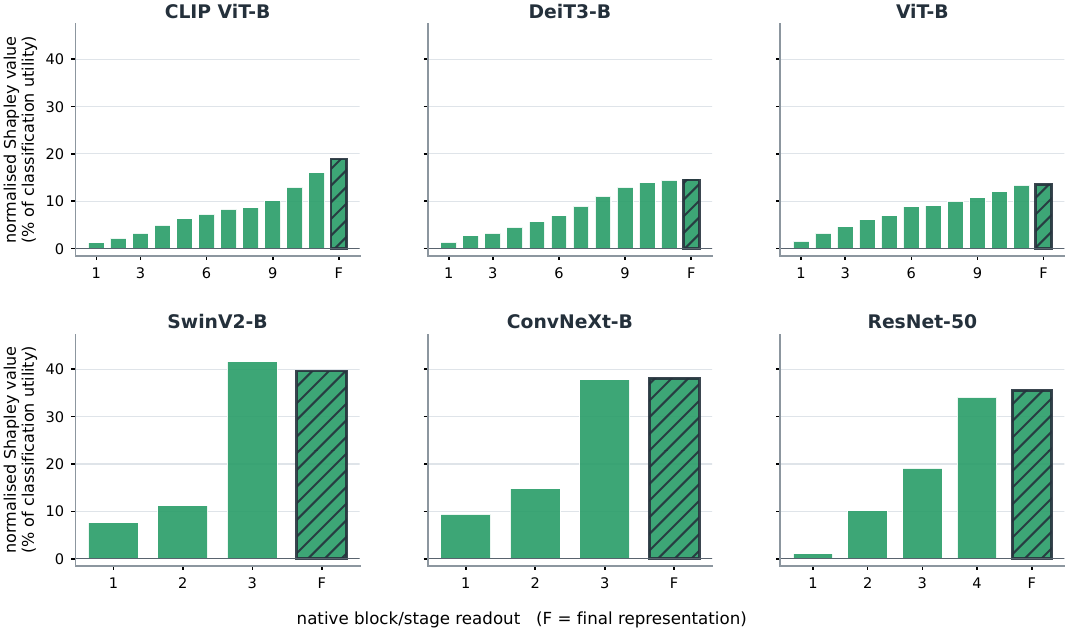}
  \caption{\textbf{Where is classification evidence readable?}
  Exact Shapley values for CIFAR-100 validation accuracy under the calibrated
  probe-voting game, normalised to 100\% within each architecture. Bars are
  native block/stage readouts; the outlined hatched bar (F) is the final
  representation. The decomposition is a held-out decoding audit and is not
  used to select or evaluate the reported test method.}
  \label{fig:classification-layer-shapley}
\end{figure*}

\subsection{Does useful depth move with the dataset?}
\label{app:classification-dataset-depth}

The CIFAR-100 attribution isolates architecture, but cannot establish task
dependence. Figure~\ref{fig:classification-dataset-depth} therefore retains the
complete validation-accuracy profile for every native readout in all 72
dataset--architecture cells. Unlike a best-layer summary, the curves expose
whether class evidence rises smoothly, saturates, forms a broad plateau, or is
reorganised near the final representation. No test labels or predictions enter
the audit.

The central observation is the absence of an architecture-independent curve.
CIFAR-10, CIFAR-100, and
Caltech-101 mostly improve toward a late plateau, whereas EuroSAT saturates by
mid-depth for nearly every family and DTD has broad intermediate optima. Food-101
and SUN397 already separate the schedules: supervised Transformers plateau or
decline late, while CLIP continues consolidating evidence. The contrast becomes
largest on fine-grained recognition. DeiT3-B, ViT-B, and Swin peak around the
last quarter on Cars, Flowers-102, and Aircraft, but CLIP gains sharply at its
final representation; the CNNs remain dataset-specific. Thus there is no
dataset- or architecture-independent ``best layer,'' monotonic refinement law,
or common movement magnitude.

This variability is a feature of the framework rather than a violation of its
assumptions. Each checkpoint is analysed in its native depth sequence, and its
class-relative route and transition typicality are learned within that model.
The method therefore does not require architectures to move alike or even to
use comparable coordinates; it can expose and exploit whichever schedule a
particular architecture--dataset pair implements.

\insightbox{Different movement is the signal}{The framework assumes neither a
shared best layer nor monotonic semantic refinement. It measures each model in
its native sequence, making architecture-specific plateaus, late jumps, and
reorganisations directly comparable outcomes rather than confounds.}

\begin{figure*}[t]
  \centering
  \includegraphics[width=0.985\textwidth]{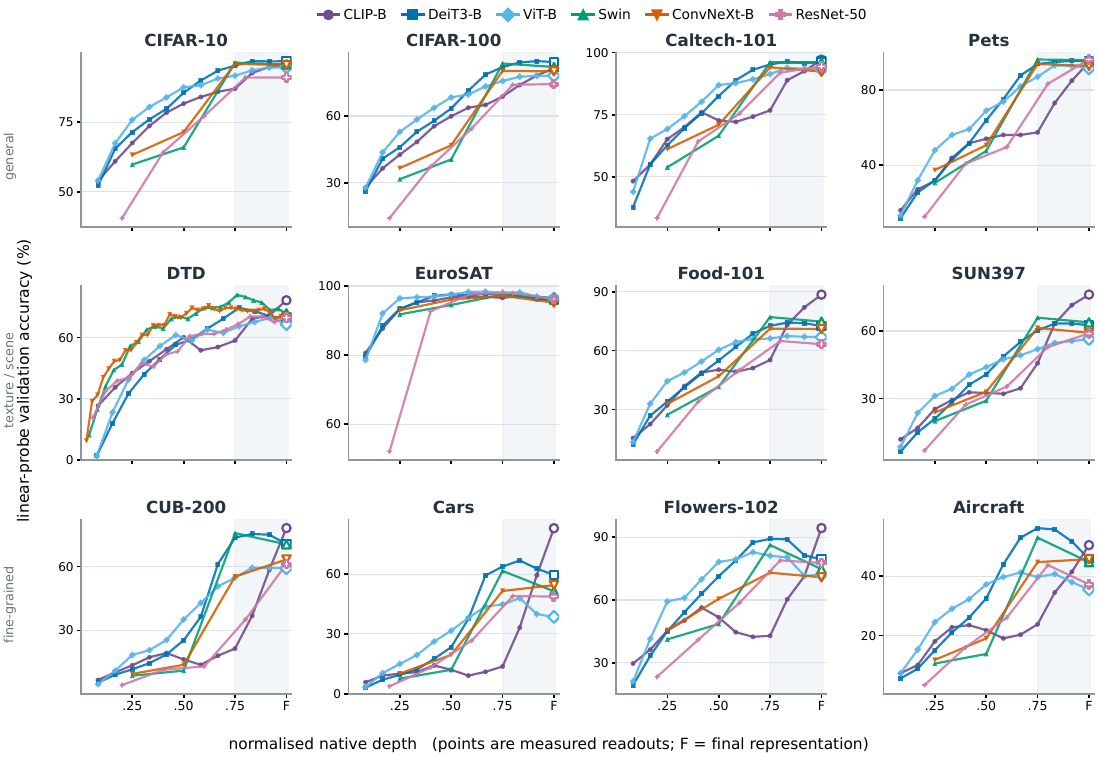}
  \caption{\textbf{There is no universal layer-readout trajectory.}
  Validation top-1 accuracy of an independently tuned frozen linear probe at
  every available native readout for the 12 datasets and six architectures in
  Table~\ref{tab:clean-classification}. Depth is normalised within each
  checkpoint; filled markers are intermediate readouts and the open marker F
  is the final representation. Panels use separately labelled $y$-scales so
  both low- and high-accuracy tasks retain their shape. DTD has block-resolved
  supervised caches; the other hierarchical-model cells use stage readouts.
  Lines only connect measured points, and no test data are accessed.}
  \label{fig:classification-dataset-depth}
\end{figure*}

\subsection{Result provenance}
\label{app:result-provenance}

The complete OOD tables are the numerical source parsed by the balanced summary
plot and retain the ImageNet-only rows excluded from its common grid. Clean
recognition values are saved per dataset, backbone, representation, and seed.
CIFAR-100-C aggregates all 10,000 images within each corruption--severity
condition before the corruption-family bootstrap. The paper reports
checkpoint-level evidence; it does not treat images or corruption severities as
independent runs. Experiments used PyTorch 2.x on Linux, primarily with one
24-GB RTX 4090. Feature extraction
dominates compute; probes, calibrators, and score combinations operate on cached
frozen features.

\section{Path Continuity Controls}
\label{app:continuity-controls}

The ID-only controls use a stratified eight-checkpoint subset on CIFAR-10 and
ImageNet-200. One 64-D projection is shared across depths within a run, and
true-class means exclude held-out prediction rows. Linear predictors are
evaluated out of sample. Identity shuffling permutes the source state across
images while preserving the target layer and its marginal distribution. The
history control compares $r_l$ with $(r_{l-1},r_l)$ on matched rows. Gap
experiments use the same protocol for $r_l\rightarrow r_{l+k}$,
$k\in\{1,2,3,4\}$. These complete controls support the continuity interpretation
of Fig.~\ref{fig:path-coherence}; they are not OOD-accuracy measurements.

\begin{table}[H]
\centering
\caption{\textbf{Complete identity-pairing control behind
Figure~\ref{fig:path-coherence}.} Mean adjacent held-out $R^2$: P uses true
image pairs, S fits with shuffled source identities, and
$\Delta=\mathrm{P}-\mathrm{S}$. Layer marginals are unchanged.}
\label{tab:continuity-full}
\tiny
\setlength{\tabcolsep}{1.2pt}
\renewcommand{\arraystretch}{1.02}
\begin{tabularx}{\linewidth}{>{\raggedright\arraybackslash}p{14mm}*{6}{>{\centering\arraybackslash}X}}
\toprule
\rowcolor{headerblue}
\textbf{Model} &
\multicolumn{3}{c}{\cellcolor{vitheader}\textbf{CIFAR-10}} &
\multicolumn{3}{c}{\cellcolor{hierheader}\textbf{ImageNet-200}} \\
\rowcolor{headerblue}
& \textbf{P} & \textbf{S} & \boldmath$\Delta$ &
\textbf{P} & \textbf{S} & \boldmath$\Delta$ \\
\midrule
\rowcolor{tablebluealt} DeiT3 & .772 & $-.007$ & \textcolor{insightaccent}{\textbf{+.779}} & .723 & $-.010$ & \textcolor{insightaccent}{\textbf{+.733}} \\
\rowcolor{tableblue} ViT AugReg & .770 & $-.013$ & \textcolor{insightaccent}{\textbf{+.783}} & .717 & $-.006$ & \textcolor{insightaccent}{\textbf{+.722}} \\
\rowcolor{tablebluealt} BeiT & .739 & $-.012$ & \textcolor{insightaccent}{\textbf{+.752}} & .713 & $-.017$ & \textcolor{insightaccent}{\textbf{+.730}} \\
\rowcolor{tableblue} Swin & .428 & $-.011$ & \textcolor{insightaccent}{\textbf{+.439}} & .356 & $-.012$ & \textcolor{insightaccent}{\textbf{+.368}} \\
\rowcolor{tablebluealt} ConvNeXt & .417 & $-.003$ & \textcolor{insightaccent}{\textbf{+.419}} & .302 & $-.014$ & \textcolor{insightaccent}{\textbf{+.316}} \\
\rowcolor{tableblue} ResNet & .548 & $-.007$ & \textcolor{insightaccent}{\textbf{+.555}} & .440 & $-.012$ & \textcolor{insightaccent}{\textbf{+.453}} \\
\rowcolor{tablebluealt} CLIP & .726 & $-.012$ & \textcolor{insightaccent}{\textbf{+.738}} & .666 & $-.006$ & \textcolor{insightaccent}{\textbf{+.672}} \\
\rowcolor{tableblue} DINOv2 & .665 & $-.005$ & \textcolor{insightaccent}{\textbf{+.670}} & .671 & $-.002$ & \textcolor{insightaccent}{\textbf{+.673}} \\
\bottomrule
\end{tabularx}
\end{table}

\section{Classification under Natural Domain Shift}
\label{app:domain-shift}

PACS and Office-Home keep the label space fixed while changing visual domain.
Every fold trains and validates on source domains only; the six checkpoints
exactly match the clean-classification architecture columns. The path branch
concatenates source-selected intermediate states (excluding the final state), and
the matched control ensembles two independently seeded final state probes.

\begin{table}[H]
\centering
\caption{\textbf{Matched natural-domain shifts.} The same six checkpoints as clean classification. Mean source-only top-1 gain (positive domains/4); $\Delta_{\rm ctl}$ subtracts the final-state ensemble gain.}
\label{tab:domain-matched}
\scriptsize
\begin{tabularx}{\linewidth}{l*{4}{>{\centering\arraybackslash}X}}
\toprule
\rowcolor{headerblue}
 & \multicolumn{2}{c}{\textbf{PACS}} & \multicolumn{2}{c}{\textbf{Office-Home}} \\
\textbf{Checkpoint} & \textbf{Gain (pos.)} & \boldmath$\Delta_{\rm ctl}$ & \textbf{Gain (pos.)} & \boldmath$\Delta_{\rm ctl}$ \\
\midrule
\rowcolor{tablebluealt}
CLIP-B & $+0.16$ (2/4) & $-0.03$ & $-0.05$ (2/4) & $-0.34$ \\
\rowcolor{tableblue}
DeiT3-B & $+5.47$ (4/4) & $+5.28$ & $+0.97$ (4/4) & $+0.70$ \\
\rowcolor{tablebluealt}
ViT-B & $+3.90$ (3/4) & $+3.69$ & $+1.40$ (4/4) & $+1.28$ \\
\rowcolor{tableblue}
SwinV2-B & $+2.99$ (3/4) & $+2.84$ & $+0.94$ (4/4) & $+0.79$ \\
\rowcolor{tablebluealt}
ConvNeXt-B & $+3.84$ (4/4) & $+3.83$ & $+1.04$ (4/4) & $+0.88$ \\
\rowcolor{tableblue}
ResNet-50 & $+2.67$ (4/4) & $+2.44$ & $+0.69$ (4/4) & $+0.56$ \\
\bottomrule
\end{tabularx}
\end{table}

\insightbox{Architecture and domain both matter}{DeiT3-B has the largest mean
PACS gain, driven especially by Photo, whereas Office-Home improvements are
smaller and more uniform. CLIP starts from the strongest final state but is the
only architecture with a negative Office-Home mean. Thus headroom alone does
not explain the ordering: the path is most useful when the frozen final state has
not already absorbed the source-to-target invariances.}

\begin{table*}[t]
\centering
\caption{\textbf{Complete natural-domain shift results.} Source-only final state $\rightarrow$ residual-path top-1 accuracy for every held-out domain (three seeds). The last columns give the mean path gain and the gain beyond an independently seeded final state ensemble.}
\label{tab:domain-shift-full}
\scriptsize
\setlength{\tabcolsep}{3.2pt}
\renewcommand{\arraystretch}{0.90}
\begin{tabularx}{\textwidth}{p{25mm}*{4}{>{\centering\arraybackslash}X}*{2}{>{\centering\arraybackslash}p{16mm}}}
\specialrule{0.1em}{0pt}{0pt}
\rowcolor{headerblue}
\textbf{PACS} & \textbf{Photo} & \textbf{Art} & \textbf{Cartoon} & \textbf{Sketch} & \textbf{Mean $\Delta$} & \boldmath$\Delta_{\rm ctl}$ \\
\midrule
\rowcolor{tablebluealt}
CLIP-B & \shortstack{$99.1\!\rightarrow\!\mathbf{99.8}$\\[-1pt]{\tiny\textcolor{insightaccent}{$\Delta=+0.62$}}} & \shortstack{$96.4\!\rightarrow\!\mathbf{96.1}$\\[-1pt]{\tiny\textcolor{red!70!black}{$\Delta=-0.24$}}} & \shortstack{$98.2\!\rightarrow\!\mathbf{97.7}$\\[-1pt]{\tiny\textcolor{red!70!black}{$\Delta=-0.44$}}} & \shortstack{$87.2\!\rightarrow\!\mathbf{87.9}$\\[-1pt]{\tiny\textcolor{insightaccent}{$\Delta=+0.70$}}} & \textcolor{insightaccent}{$+0.16$} & \textcolor{red!70!black}{$-0.03$} \\
\rowcolor{tableblue}
DeiT3-B & \shortstack{$75.0\!\rightarrow\!\mathbf{89.4}$\\[-1pt]{\tiny\textcolor{insightaccent}{$\Delta=+14.47$}}} & \shortstack{$87.3\!\rightarrow\!\mathbf{90.4}$\\[-1pt]{\tiny\textcolor{insightaccent}{$\Delta=+3.16$}}} & \shortstack{$64.9\!\rightarrow\!\mathbf{66.7}$\\[-1pt]{\tiny\textcolor{insightaccent}{$\Delta=+1.85$}}} & \shortstack{$53.2\!\rightarrow\!\mathbf{55.6}$\\[-1pt]{\tiny\textcolor{insightaccent}{$\Delta=+2.40$}}} & \textcolor{insightaccent}{$+5.47$} & \textcolor{insightaccent}{$+5.28$} \\
\rowcolor{tablebluealt}
ViT-B & \shortstack{$88.4\!\rightarrow\!\mathbf{96.0}$\\[-1pt]{\tiny\textcolor{insightaccent}{$\Delta=+7.58$}}} & \shortstack{$65.6\!\rightarrow\!\mathbf{69.2}$\\[-1pt]{\tiny\textcolor{insightaccent}{$\Delta=+3.68$}}} & \shortstack{$57.1\!\rightarrow\!\mathbf{62.3}$\\[-1pt]{\tiny\textcolor{insightaccent}{$\Delta=+5.25$}}} & \shortstack{$36.5\!\rightarrow\!\mathbf{35.6}$\\[-1pt]{\tiny\textcolor{red!70!black}{$\Delta=-0.89$}}} & \textcolor{insightaccent}{$+3.90$} & \textcolor{insightaccent}{$+3.69$} \\
\rowcolor{tableblue}
SwinV2-B & \shortstack{$75.8\!\rightarrow\!\mathbf{84.1}$\\[-1pt]{\tiny\textcolor{insightaccent}{$\Delta=+8.26$}}} & \shortstack{$81.7\!\rightarrow\!\mathbf{85.0}$\\[-1pt]{\tiny\textcolor{insightaccent}{$\Delta=+3.34$}}} & \shortstack{$61.0\!\rightarrow\!\mathbf{61.9}$\\[-1pt]{\tiny\textcolor{insightaccent}{$\Delta=+0.85$}}} & \shortstack{$41.4\!\rightarrow\!\mathbf{40.9}$\\[-1pt]{\tiny\textcolor{red!70!black}{$\Delta=-0.51$}}} & \textcolor{insightaccent}{$+2.99$} & \textcolor{insightaccent}{$+2.84$} \\
\rowcolor{tablebluealt}
ConvNeXt-B & \shortstack{$80.2\!\rightarrow\!\mathbf{87.8}$\\[-1pt]{\tiny\textcolor{insightaccent}{$\Delta=+7.62$}}} & \shortstack{$81.0\!\rightarrow\!\mathbf{84.1}$\\[-1pt]{\tiny\textcolor{insightaccent}{$\Delta=+3.11$}}} & \shortstack{$60.2\!\rightarrow\!\mathbf{63.5}$\\[-1pt]{\tiny\textcolor{insightaccent}{$\Delta=+3.36$}}} & \shortstack{$56.7\!\rightarrow\!\mathbf{58.0}$\\[-1pt]{\tiny\textcolor{insightaccent}{$\Delta=+1.26$}}} & \textcolor{insightaccent}{$+3.84$} & \textcolor{insightaccent}{$+3.83$} \\
\rowcolor{tableblue}
ResNet-50 & \shortstack{$94.3\!\rightarrow\!\mathbf{94.4}$\\[-1pt]{\tiny\textcolor{insightaccent}{$\Delta=+0.06$}}} & \shortstack{$75.1\!\rightarrow\!\mathbf{77.3}$\\[-1pt]{\tiny\textcolor{insightaccent}{$\Delta=+2.16$}}} & \shortstack{$57.8\!\rightarrow\!\mathbf{60.6}$\\[-1pt]{\tiny\textcolor{insightaccent}{$\Delta=+2.77$}}} & \shortstack{$52.4\!\rightarrow\!\mathbf{58.1}$\\[-1pt]{\tiny\textcolor{insightaccent}{$\Delta=+5.68$}}} & \textcolor{insightaccent}{$+2.67$} & \textcolor{insightaccent}{$+2.44$} \\
\midrule
\addlinespace[2pt]
\rowcolor{headerblue}
\textbf{Office-Home} & \textbf{Art} & \textbf{Clipart} & \textbf{Product} & \textbf{Real} & \textbf{Mean $\Delta$} & \boldmath$\Delta_{\rm ctl}$ \\
\midrule
\rowcolor{tablebluealt}
CLIP-B & \shortstack{$80.5\!\rightarrow\!\mathbf{80.7}$\\[-1pt]{\tiny\textcolor{insightaccent}{$\Delta=+0.23$}}} & \shortstack{$69.1\!\rightarrow\!\mathbf{68.7}$\\[-1pt]{\tiny\textcolor{red!70!black}{$\Delta=-0.35$}}} & \shortstack{$90.0\!\rightarrow\!\mathbf{89.9}$\\[-1pt]{\tiny\textcolor{red!70!black}{$\Delta=-0.11$}}} & \shortstack{$89.2\!\rightarrow\!\mathbf{89.3}$\\[-1pt]{\tiny\textcolor{insightaccent}{$\Delta=+0.02$}}} & \textcolor{red!70!black}{$-0.05$} & \textcolor{red!70!black}{$-0.34$} \\
\rowcolor{tableblue}
DeiT3-B & \shortstack{$75.0\!\rightarrow\!\mathbf{76.2}$\\[-1pt]{\tiny\textcolor{insightaccent}{$\Delta=+1.11$}}} & \shortstack{$56.9\!\rightarrow\!\mathbf{57.7}$\\[-1pt]{\tiny\textcolor{insightaccent}{$\Delta=+0.78$}}} & \shortstack{$84.5\!\rightarrow\!\mathbf{85.1}$\\[-1pt]{\tiny\textcolor{insightaccent}{$\Delta=+0.63$}}} & \shortstack{$84.7\!\rightarrow\!\mathbf{86.1}$\\[-1pt]{\tiny\textcolor{insightaccent}{$\Delta=+1.35$}}} & \textcolor{insightaccent}{$+0.97$} & \textcolor{insightaccent}{$+0.70$} \\
\rowcolor{tablebluealt}
ViT-B & \shortstack{$64.7\!\rightarrow\!\mathbf{66.0}$\\[-1pt]{\tiny\textcolor{insightaccent}{$\Delta=+1.22$}}} & \shortstack{$48.6\!\rightarrow\!\mathbf{50.1}$\\[-1pt]{\tiny\textcolor{insightaccent}{$\Delta=+1.42$}}} & \shortstack{$79.6\!\rightarrow\!\mathbf{80.7}$\\[-1pt]{\tiny\textcolor{insightaccent}{$\Delta=+1.10$}}} & \shortstack{$79.5\!\rightarrow\!\mathbf{81.3}$\\[-1pt]{\tiny\textcolor{insightaccent}{$\Delta=+1.87$}}} & \textcolor{insightaccent}{$+1.40$} & \textcolor{insightaccent}{$+1.28$} \\
\rowcolor{tableblue}
SwinV2-B & \shortstack{$74.8\!\rightarrow\!\mathbf{76.0}$\\[-1pt]{\tiny\textcolor{insightaccent}{$\Delta=+1.17$}}} & \shortstack{$55.3\!\rightarrow\!\mathbf{55.9}$\\[-1pt]{\tiny\textcolor{insightaccent}{$\Delta=+0.53$}}} & \shortstack{$84.9\!\rightarrow\!\mathbf{85.8}$\\[-1pt]{\tiny\textcolor{insightaccent}{$\Delta=+0.89$}}} & \shortstack{$85.4\!\rightarrow\!\mathbf{86.6}$\\[-1pt]{\tiny\textcolor{insightaccent}{$\Delta=+1.16$}}} & \textcolor{insightaccent}{$+0.94$} & \textcolor{insightaccent}{$+0.79$} \\
\rowcolor{tablebluealt}
ConvNeXt-B & \shortstack{$73.5\!\rightarrow\!\mathbf{74.4}$\\[-1pt]{\tiny\textcolor{insightaccent}{$\Delta=+0.96$}}} & \shortstack{$53.7\!\rightarrow\!\mathbf{54.5}$\\[-1pt]{\tiny\textcolor{insightaccent}{$\Delta=+0.76$}}} & \shortstack{$83.1\!\rightarrow\!\mathbf{84.2}$\\[-1pt]{\tiny\textcolor{insightaccent}{$\Delta=+1.17$}}} & \shortstack{$84.1\!\rightarrow\!\mathbf{85.4}$\\[-1pt]{\tiny\textcolor{insightaccent}{$\Delta=+1.28$}}} & \textcolor{insightaccent}{$+1.04$} & \textcolor{insightaccent}{$+0.88$} \\
\rowcolor{tableblue}
ResNet-50 & \shortstack{$68.6\!\rightarrow\!\mathbf{69.0}$\\[-1pt]{\tiny\textcolor{insightaccent}{$\Delta=+0.41$}}} & \shortstack{$47.4\!\rightarrow\!\mathbf{47.6}$\\[-1pt]{\tiny\textcolor{insightaccent}{$\Delta=+0.22$}}} & \shortstack{$79.5\!\rightarrow\!\mathbf{80.6}$\\[-1pt]{\tiny\textcolor{insightaccent}{$\Delta=+1.13$}}} & \shortstack{$82.3\!\rightarrow\!\mathbf{83.3}$\\[-1pt]{\tiny\textcolor{insightaccent}{$\Delta=+1.01$}}} & \textcolor{insightaccent}{$+0.69$} & \textcolor{insightaccent}{$+0.56$} \\
\specialrule{0.1em}{0pt}{0pt}
\end{tabularx}
\end{table*}

\section{Ablations and Design Choices}
\label{app:ablations}

We isolate implementation choices on frozen CIFAR-10 features while retaining
the deployed OOD protocol: 1,000 fit and 300 calibration examples per class,
seed 0, an 80-epoch transition MLP, Mahalanobis++ as the final state member, and a
fixed path weight of $0.3$.  The deterministic evaluation screen contains
5,000 ID examples and at most 3,000 examples from each of the six OpenOOD
splits.  These experiments are diagnostic rather than a replacement for the
three-seed complete tables.

\subsection{Why class residuals rather than raw activations?}

\begin{table*}[t]
\centering
\caption{\textbf{Class-specific route residualization exposes OOD evidence
hidden in raw activations (FPR95 $\downarrow$).}  Within each panel, examples,
layers, detector recipe, and ID calibration are identical; only the coordinates
given to the path detector change.  Positive $\Delta$ is the reduction from raw
activations.}
\label{tab:representation-ablation}
\scriptsize
\setlength{\tabcolsep}{2.5pt}
\renewcommand{\arraystretch}{1.08}
\begin{minipage}[t]{0.545\textwidth}
\centering
\textbf{(a) Deployed MLP+Mahalanobis++ screen, CIFAR-10 ID}\par\vspace{2pt}
\resizebox{\linewidth}{!}{%
\begin{tabular}{lrrrrrrrr}
\toprule
\rowcolor{headerblue}
\textbf{Path coordinates} & \textbf{DeiT3} & \textbf{ConvN} & \textbf{RN50} & \textbf{Swin} & \textbf{Near} & \textbf{Far} & \textbf{Mean} & \boldmath$\Delta$ \\
\midrule
\rowcolor{tablebluealt} Raw states & 16.29 & 22.97 & 24.74 & 6.73 & 25.99 & 13.53 & 17.68 & --- \\
\rowcolor{tableblue} Global-centred & 16.12 & 23.02 & 24.82 & 6.80 & 26.07 & 13.50 & 17.69 & $-0.01$ \\
\rowcolor{tablebluealt} Final-state residual & 15.84 & \bestcell{22.84} & 24.32 & 6.79 & 25.75 & 13.30 & 17.45 & \textcolor{insightaccent}{$+0.23$} \\
\rowcolor{sectionblue} \textbf{Class-route residual} & \bestcell{15.39} & 23.12 & \bestcell{23.75} & \bestcell{6.58} & \bestcell{25.25} & \bestcell{13.19} & \bestcell{17.21} & \textcolor{insightaccent}{$\mathbf{+0.47}$} \\
\bottomrule
\end{tabular}%
}
\end{minipage}\hfill
\begin{minipage}[t]{0.445\textwidth}
\centering
\textbf{(b) Standalone trajectory replication, ImageNet-1K ID}\par\vspace{2pt}
\resizebox{\linewidth}{!}{%
\begin{tabular}{lrrrrrrr}
\toprule
\rowcolor{headerblue}
\textbf{Path coordinates} & \textbf{DINOv3} & \textbf{DINOv2} & \textbf{CLIP} & \textbf{PE} & \textbf{SigLIP2} & \textbf{Mean} & \boldmath$\Delta$ \\
\midrule
\rowcolor{tablebluealt} Raw states & 43.19 & 37.38 & 50.63 & 53.52 & 45.97 & 46.14 & --- \\
\rowcolor{tableblue} Global-centred & 44.75 & 45.35 & 54.02 & 47.63 & 42.89 & 46.93 & $-0.79$ \\
\rowcolor{sectionblue} \textbf{Class-route residual} & \bestcell{30.46} & \bestcell{35.72} & \bestcell{47.74} & \bestcell{41.62} & \bestcell{34.99} & \bestcell{38.11} & \textcolor{insightaccent}{$\mathbf{+8.03}$} \\
\bottomrule
\end{tabular}%
}
\end{minipage}
\end{table*}

Table~\ref{tab:representation-ablation} changes only the coordinates supplied
to the path member.  Raw states retain between-class transport; global
centring removes one layer mean; final-state-only residualization subtracts the
assigned class prototype only at the last state; the proposed representation
subtracts the assigned class path at every depth.  All evaluation assignments
use the nearest ID class at the final state---never an evaluation label.

Under the deployed MLP+Mahalanobis++ protocol (panel a), the class route has the
best architecture-mean AUROC and lowers mean FPR95 from $17.68$ to $17.21$.
Final-state-only subtraction reaches $17.45$, so approximately half of this
reduction requires class conditioning along the intermediate route rather than
only at its destination.  The effect is architecture-dependent: class-route
residuals win mean FPR95 on DeiT3, ResNet, and Swin; ConvNeXt is $0.15$ points
worse than raw on the six-split mean despite improving near-OOD FPR95 by
$1.09$ and mean AUROC by $0.25$.

Panel b removes the strong final state partner and repeats the representation
test with one matched autoregressive-plus-static trajectory detector on five
ImageNet foundation encoders.  Class-route residuals win all 5/5 backbones,
reducing average FPR95 from $46.14$ to $38.11$.  Global centring instead gives
$46.93$; all arms are ID-standardised before combination, ruling out generic
mean subtraction or score scale as the cause.
Residualization does not create information: conditional on the selected
anchor it is deterministic.  It exposes input-specific innovation by removing
the dominant class-coherent route, which makes that information easier for a
finite-capacity ID predictor and density model to use.

\insightbox{What the representation controls rule out}{The gain is not caused
by evaluation-label access, generic centring, final state class conditioning
alone, a single foundation encoder, or the absence of a strong final state
detector.  Its magnitude varies by architecture, but the independent
ImageNet-1K replication shows that the residual advantage is broad.}

\subsection{How much history does the OOD predictor need?}

\begin{table*}[t]
\centering
\caption{\textbf{OOD design ablations on CIFAR-10 ID (FPR95 $\downarrow$).}
The final state member, fit/calibration split, whitening, and fixed path weight are
unchanged.  Positive $\Delta$ is the FPR95 reduction relative to the deployed
current-only/Gaussian choice.  No OOD sample is used for fitting or selection.}
\label{tab:ood-design-ablation}
\scriptsize
\setlength{\tabcolsep}{2.6pt}
\renewcommand{\arraystretch}{1.06}
\begin{minipage}[t]{0.493\textwidth}
\centering
\textbf{(a) Residual context supplied to the transition MLP}\par\vspace{2pt}
\begin{tabular}{llrrrr}
\toprule
\rowcolor{headerblue}
\textbf{Checkpoint} & \textbf{Context} & \textbf{Near} & \textbf{Far} & \textbf{Mean} & \boldmath$\Delta$ \\
\midrule
\rowcolor{tablebluealt} DeiT3-B & current & 25.30 & 10.43 & 15.39 & --- \\
\rowcolor{tablebluealt} & two-state & 24.70 & 10.33 & 15.12 & \textcolor{insightaccent}{$+0.27$} \\
\rowcolor{tablebluealt} & full prefix & 24.85 & 9.59 & \bestcell{14.68} & \textcolor{insightaccent}{$\mathbf{+0.71}$} \\
\rowcolor{tableblue} ConvNeXt-B & current & 26.58 & 21.39 & 23.12 & --- \\
\rowcolor{tableblue} & two-state & 26.55 & 21.91 & 23.46 & $-0.33$ \\
\rowcolor{tableblue} & full prefix & 26.25 & 21.32 & \bestcell{22.96} & \textcolor{insightaccent}{$\mathbf{+0.16}$} \\
\rowcolor{tablebluealt} ResNet-50 & current & 29.58 & 20.83 & 23.75 & --- \\
\rowcolor{tablebluealt} & two-state & 29.25 & 20.37 & \bestcell{23.33} & \textcolor{insightaccent}{$\mathbf{+0.42}$} \\
\rowcolor{tablebluealt} & full prefix & 29.37 & 20.35 & 23.36 & \textcolor{insightaccent}{$+0.39$} \\
\rowcolor{tableblue} Swin-B & current & 19.52 & 0.11 & 6.58 & --- \\
\rowcolor{tableblue} & two-state & 19.38 & 0.06 & \bestcell{6.50} & \textcolor{insightaccent}{$\mathbf{+0.08}$} \\
\rowcolor{tableblue} & full prefix & 19.35 & 0.07 & \bestcell{6.50} & \textcolor{insightaccent}{$\mathbf{+0.08}$} \\
\midrule
\rowcolor{sectionblue} \textbf{Architecture mean} & current & 25.25 & 13.19 & 17.21 & --- \\
\rowcolor{sectionblue} & two-state & 24.97 & 13.17 & 17.10 & \textcolor{insightaccent}{$+0.11$} \\
\rowcolor{sectionblue} & full prefix & 24.95 & 12.83 & \bestcell{16.87} & \textcolor{insightaccent}{$\mathbf{+0.34}$} \\
\bottomrule
\end{tabular}
\end{minipage}\hfill
\begin{minipage}[t]{0.493\textwidth}
\centering
\textbf{(b) Common-width map for variable-width stages}\par\vspace{2pt}
\begin{tabular}{llrrrr}
\toprule
\rowcolor{headerblue}
\textbf{Checkpoint} & \textbf{Stage map} & \textbf{Near} & \textbf{Far} & \textbf{Mean} & \boldmath$\Delta$ \\
\midrule
\rowcolor{tablebluealt} ConvNeXt-B & Gaussian & 26.58 & 21.39 & 23.12 & --- \\
\rowcolor{tablebluealt} & zero-pad & 27.22 & 24.97 & 25.72 & $-2.60$ \\
\rowcolor{tablebluealt} & interpolate & 25.78 & 21.48 & \bestcell{22.92} & \textcolor{insightaccent}{$\mathbf{+0.21}$} \\
\rowcolor{tableblue} ResNet-50 & Gaussian & 29.58 & 20.83 & 23.75 & --- \\
\rowcolor{tableblue} & zero-pad & 28.93 & 21.92 & 24.26 & $-0.51$ \\
\rowcolor{tableblue} & interpolate & 28.78 & 19.58 & \bestcell{22.65} & \textcolor{insightaccent}{$\mathbf{+1.10}$} \\
\rowcolor{tablebluealt} Swin-B & Gaussian & 19.52 & 0.11 & 6.58 & --- \\
\rowcolor{tablebluealt} & zero-pad & 21.85 & 0.83 & 7.83 & $-1.26$ \\
\rowcolor{tablebluealt} & interpolate & 19.45 & 0.13 & \bestcell{6.57} & \textcolor{insightaccent}{$\mathbf{+0.01}$} \\
\midrule
\rowcolor{sectionblue} \textbf{Architecture mean} & Gaussian & 25.23 & 14.11 & 17.82 & --- \\
\rowcolor{sectionblue} & zero-pad & 26.00 & 15.91 & 19.27 & $-1.46$ \\
\rowcolor{sectionblue} & interpolate & 24.67 & 13.73 & \bestcell{17.38} & \textcolor{insightaccent}{$\mathbf{+0.44}$} \\
\bottomrule
\end{tabular}
\end{minipage}
\end{table*}

Table~\ref{tab:ood-design-ablation}a compares the deployed Markov input $r_l$
with $(r_{l-1},r_l)$ and with $r_l$ plus the mean of every preceding residual.
Full-prefix context lowers mean FPR95 for all four checkpoints, but the average
reduction is only $0.34$ points; the two-state variant is mixed and is worse on
ConvNeXt-B.  The largest prefix benefit occurs for the 12-block DeiT3 path
($0.71$), whereas the four-stage Swin path changes by only $0.08$.  This is
consistent with the current state already carrying most earlier computation,
while longer paths leave more non-redundant context to aggregate.  We retain
the current-only member in the main benchmark because it is the simpler fixed
design; the stronger prefix result establishes that additional history is a
small, consistently positive extension rather than a missing prerequisite.

\subsection{Does the common-width map have to be random?}

Table~\ref{tab:ood-design-ablation}b replaces the seed-0 Gaussian stage map for
ConvNeXt, ResNet, and Swin.  Channel interpolation is deterministic and lowers
mean FPR95 for every checkpoint ($0.44$ points on average, $1.10$ on
ResNet-50), so the result cannot be attributed to a fortunate random matrix.
Zero-padding is worse for all three models ($1.46$ points on average).  A
plausible explanation is that its stage-dependent inactive coordinates expose
width changes to the predictor, whereas Gaussian and interpolated maps keep
every common coordinate active.  Because channels from different stages need
not be semantically aligned, the main protocol keeps the architecture-agnostic
Gaussian map; interpolation is a promising deterministic alternative.

The original pooled stage directions are recovered exactly from the cached
Gaussian trajectories using the right inverse
$P_l^\top(P_lP_l^\top)^{-1}$, then remapped and renormalized.  Thus this
comparison changes the common-width map without re-extracting images or
changing the backbone, examples, final state score, or class anchors.

\subsection{Predictor capacity and direction}
\label{app:direction-capacity}

The reverse-prediction advantage is not an artifact of using a weak linear
predictor. MLP, gated recurrent, attention, and temporal-convolution models all
receive the same projected ID paths and are trained separately in each
direction. Increasing sequence-model capacity changes error magnitude but not
the ordering: reverse prediction remains easier in every
architecture--benchmark case.

\begin{figure*}[t]
\centering
\includegraphics[width=0.82\textwidth]{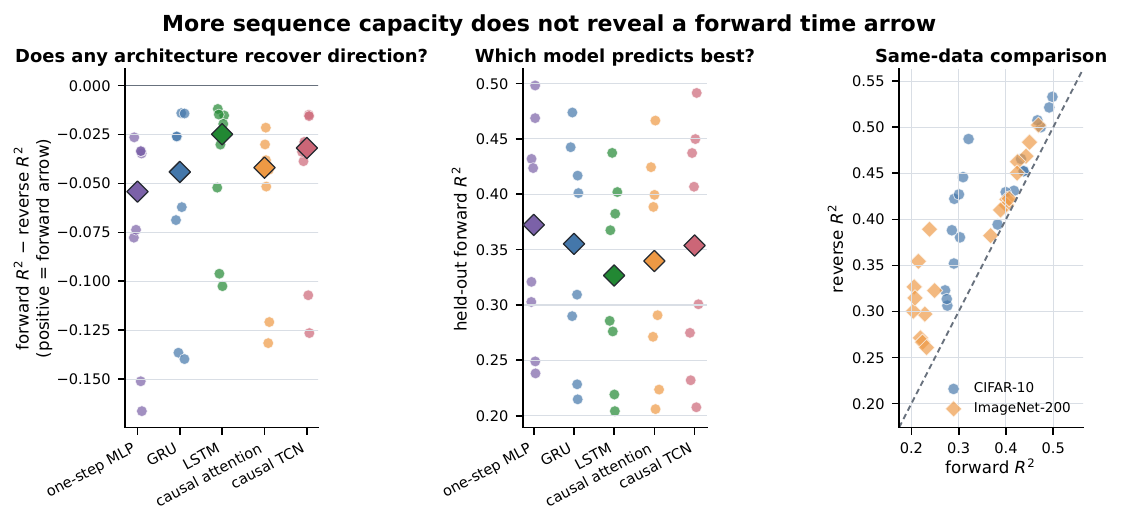}
\caption{\textbf{More sequence capacity does not reveal a forward time arrow.}
MLP, GRU, LSTM, causal-attention, and temporal-convolution predictors are
trained separately in both directions on the same projected ID paths. Reverse
prediction remains easier throughout.}
\label{fig:direction-sweep}
\end{figure*}


The prefix-mean result in Table~\ref{tab:ood-design-ablation}a shows that
earlier states contain some additional information, but does not establish
that a recurrent or attention-based predictor is necessary. We therefore
replace the shared depth-conditioned MLP with approximately
parameter-matched GRU, LSTM, causal-attention, and causal-TCN predictors.
At transition $l$, these models receive the complete preceding trajectory
$r_{1:l}$; all residuals, data partitions, OOD-score partner, calibration
statistics, and fusion coefficients remain unchanged. Model fitting and
selection remain ID-only.

\begin{figure}[t]
\centering
\includegraphics[width=\columnwidth]
  {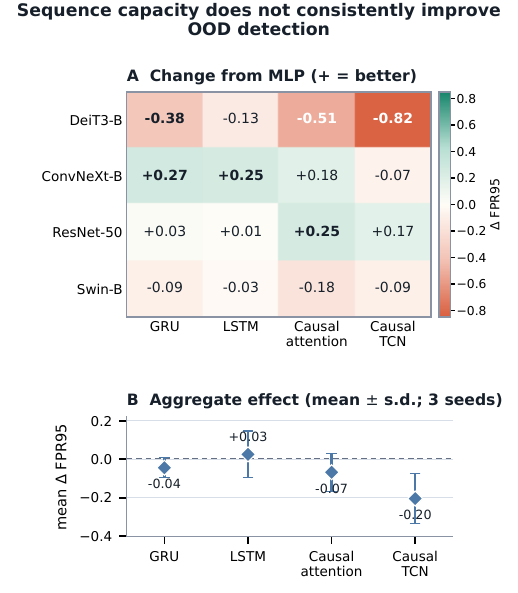}
\caption{\textbf{More sequence capacity does not consistently improve OOD detection.}
We compare the shared depth-conditioned MLP with approximately
parameter-matched GRU, LSTM, causal-attention, and causal-TCN predictors that
use the preceding trajectory. Results use CIFAR-10 as ID, six near/far OOD
datasets, four backbones, and three seeds. (A) Mean FPR95 change relative to
the shared MLP for each architecture; positive values indicate improvement.
(B) Aggregate change across backbones, reported as mean $\pm$ standard
deviation across seeds. Full-prefix models provide no consistent advantage
over the shared MLP.}
\label{fig:ood-predictor-family}
\end{figure}

Figure~\ref{fig:ood-predictor-family} separates the availability of history
from the value of additional sequence-model capacity. No predictor family
improves consistently across architectures. ConvNeXt-B benefits from the GRU,
LSTM, and causal-attention variants, while ResNet-50 benefits most from causal
attention and the TCN. In contrast, every sequence model degrades DeiT3-B, and
the changes for Swin-B are small but consistently negative. These opposing
effects largely cancel in the aggregate: the LSTM is effectively tied with the
shared MLP, whereas the GRU, causal-attention, and causal-TCN variants are
slightly worse on average.

The result also clarifies why prediction accuracy and OOD utility should not
be conflated. A more expressive model may predict ID transitions differently
without producing errors that separate ID from OOD more reliably. The current
residual state already summarizes much of the preceding computation, and the
small benefit of the explicit prefix mean does not translate into a general
advantage for learned full-prefix models. We therefore retain the shared MLP:
it is simpler, competitive across architectures, and avoids introducing
architecture-dependent sequence-model choices.

\subsection{How much should the path count?}
\label{app:weight}

The path weight in Equation~7 was fixed at $0.3$ before evaluation and
held constant across every checkpoint, benchmark, and OOD split
reported in this paper. It is not selected per case, per architecture,
or per benchmark, and no OOD sample enters the choice.

The value is conservative by design. Our claim is that computation
paths \emph{complement} final-state detectors rather than replace
them, so the weight was set low enough that the final-state score
remains the dominant term and the path acts as a correction to it. A
weight chosen to maximise average performance would test a different
and weaker claim, since it would make the reported gains partly a
property of the tuning procedure rather than of the trajectory signal.

\begin{figure}[h]
\centering
\includegraphics[width=\columnwidth]{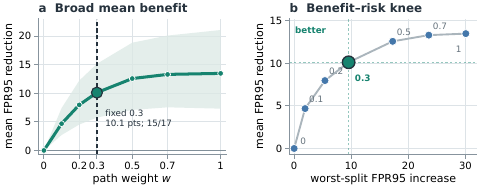}
\caption{\textbf{The fixed path weight is not a tuned parameter.}
The weight $w=0.3$ was fixed a priori; this sweep is a post-hoc audit
rather than a selection procedure. \emph{(a)} Mean FPR95 reduction
against $w$ over 17 checkpoint--benchmark cases; the band is the
interquartile range. Every $w \in [0.1,1]$ yields a positive mean
reduction. \emph{(b)} The same sweep against the largest FPR95
increase on any single OOD split. Mean benefit saturates while
worst-split risk keeps growing, so weights beyond $0.3$ buy little
average improvement at substantially higher tail cost.}
\label{fig:weight-sweep}
\end{figure}

Figure~\ref{fig:weight-sweep} audits the choice after the fact. Mean
benefit is positive across the entire range $[0.1,1]$, so the reported
conclusions do not depend on the exact value. Benefit also continues
to rise past $0.3$ and saturates near $0.5$, meaning the deployed
weight is not the one that maximises mean FPR95 reduction.

Two properties motivate keeping it anyway. First, the largest FPR95
increase on any single OOD split grows steadily with $w$ while mean
benefit flattens, so higher weights trade tail reliability for little
additional average gain. Second, case-to-case dispersion widens in the
same direction: the interquartile range in panel (a) is substantially
broader at $w=1$ than at $w=0.3$, so a larger weight also makes the
outcome less predictable for an unseen checkpoint or benchmark.

The main results therefore understate what a tuned weight would
achieve, which is the intended trade. A single fixed value that
transfers across 42 checkpoints and five benchmarks without per-case
selection is more useful in deployment than an optimal value that
would require OOD data to find.

\subsection{Robustness analyses beyond representation choice}

Appendix~\ref{app:continuity-controls} provides complementary analyses of
sample identity, temporal context, and prediction horizon. Paired history
improves held-out ID prediction in all 16 architecture--dataset combinations,
whereas shuffling that history removes the gain. Prediction quality also
degrades progressively as the target is moved one to four layers ahead. These
results indicate that the modest OOD improvement obtained from a representation
prefix arises from meaningful sample-specific temporal context rather than
simply increasing the input dimensionality.

Appendix~\ref{app:complementarity} further evaluates different final-state
partners (Maha++, RelMaha, X-Maha, kNN, and trajectory density) under the same
ID-only calibration protocol. Collectively, these analyses examine coordinate
alignment, temporal context, predictor family, sample identity, transition
horizon, and score combination while varying one design component at a time.

\takeawaybox{Summary of robustness analyses}{
Representation choice: Table~\ref{tab:representation-ablation}. Context length:
Table~\ref{tab:ood-design-ablation}a. Sample identity:
Table~\ref{tab:ood-design-ablation}b and
Table~\ref{tab:continuity-full}. Prediction horizon:
Fig.~\ref{fig:direction-sweep}. Score complementarity:
Table~\ref{tab:partner-control}.}

\section{Movement versus Relational Reorganization}
\label{app:geometry-audit}
A shared rotation can move every state without changing pairwise
relations, while a smaller step can still reorganize neighbourhoods. We
therefore report movement and CKA deformation separately in
Fig.~\ref{fig:geometry-axes}.

\begin{figure*}[t]
\centering
\includegraphics[width=0.98\textwidth]{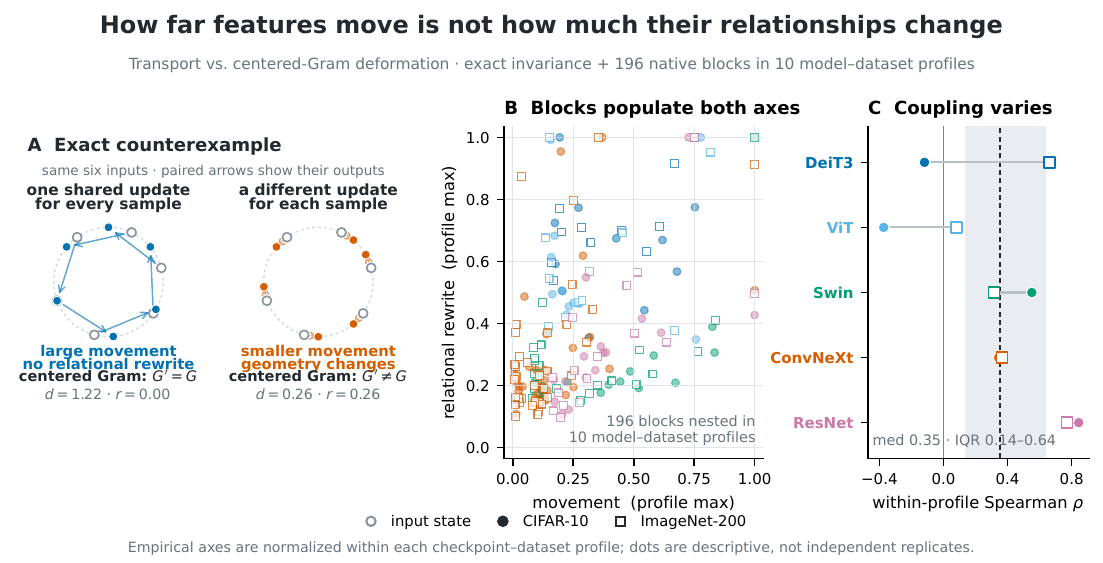}
\caption{\textbf{Coordinate movement does not determine relational rewrite.}
A shared rotation preserves the centered sample Gram matrix despite moving
every state, whereas smaller sample-specific updates can rewrite it. Here $d$
is RMS paired-feature displacement and $r=\sqrt{1-\mathrm{CKA}}$. Native
blocks likewise show only an imperfect within-profile association between the
two axes.}
\label{fig:geometry-axes}
\end{figure*}

Across the ten checkpoint--dataset profiles, the median within-profile
Spearman association between the two axes is only $0.355$ (IQR
$0.140$--$0.636$), and two profiles are negative. The exact rotation control
shows why this is not a noisy version of one quantity: coherent transport can
be large while pairwise geometry is unchanged. Conversely, stage entries can
rewrite neighbours without producing the largest vector step. This separation
prevents us from interpreting a generic ``velocity peak'' as a universal
semantic event.

\section{Architecture Breadth and Intervention Stability}
\label{app:architecture-breadth}

The five representative checkpoints establish native every-block behavior;
the breadth audit then repeats the analysis for small/base/large DeiT3, Swin,
ConvNeXt, and ResNet-50/101/152. A fingerprint concatenates depth-standardised
rewrite, final state alignment, prototype accuracy, Fisher separation, and
innovation-share profiles. Within-family similarity exceeds between-family
similarity on CIFAR-10 ($0.750$ vs. $0.654$, permutation $p=3.25\times10^{-4}$)
and ImageNet-200 ($0.785$ vs. $0.697$, $p=4.55\times10^{-4}$). More decisively,
a CIFAR-10 fingerprint retrieves the exact ImageNet-200 checkpoint for 11/12
models; same-checkpoint correlation is $0.843$ versus $0.638$ off diagonal.
\begin{figure*}[t]
\centering
{\small\bfseries Computation schedules recur across model scale and datasets\par}
\vspace{1mm}
\includegraphics[width=0.49\textwidth,trim=10pt 22pt 235pt 25pt,clip]{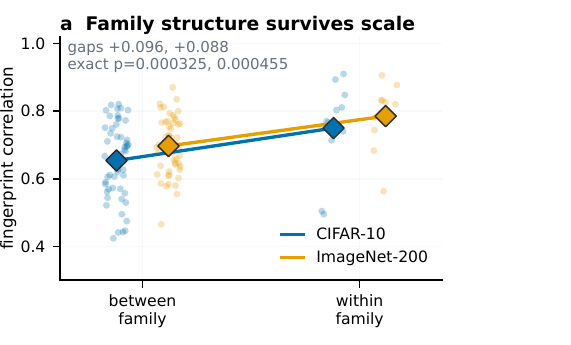}
\vspace{1mm}
~
\includegraphics[width=0.49\textwidth,trim=285pt 22pt 5pt 25pt,clip]{figures/fig_aaai_architecture_recurrence.pdf}
\vspace{1mm}
\begingroup
\scriptsize
\setlength{\tabcolsep}{3pt}
\renewcommand{\arraystretch}{1.08}
\begin{tabularx}{0.94\textwidth}{p{29mm}X}
\toprule
\rowcolor{headerblue}
\textbf{Audit} & \textbf{Evidence} \\
\midrule
\rowcolor{tablebluealt}
Family recurrence & CIFAR-10: $0.750$ vs. $0.654$ ($p=3.25\!\times\!10^{-4}$); ImageNet-200: $0.785$ vs. $0.697$ ($p=4.55\!\times\!10^{-4}$). \\
\rowcolor{tableblue}
Checkpoint recovery & 11/12 exact cross-dataset matches; correlation $0.843$ for the same checkpoint vs. $0.638$ off diagonal. \\
\bottomrule
\end{tabularx}
\endgroup
\caption{\textbf{Architecture-associated schedules survive scale and dataset.}
\emph{(a)} Within-family fingerprints are more similar than between-family
ones on both evaluation sets. \emph{(b)} Cross-dataset matching identifies the
exact checkpoint in 11/12 cases. These are descriptive properties of public
checkpoints, not a controlled causal architecture comparison.}
\label{fig:architecture-fingerprints}
\end{figure*}

The controlled perturbation screen provides a different test: it asks where a
fixed 5\% channel perturbation is contracted or amplified. The depth response
is architecture-specific---ViT-B usually contracts, Swin amplifies toward the
end of stage 3, ConvNeXt amplifies strongly in late stage 3 before a stage-4
reset, and ResNet is most sensitive early. Yet the response profile transfers
from CIFAR-10 to ImageNet-200 with correlations $0.94$--$1.00$ across the five
checkpoints (direction-specific range $0.884$--$0.999$). This intervention
evidence rules out a universal late-layer protection story and independently
supports stable computation schedules.

\begin{figure*}[t!]
\centering
\includegraphics[width=0.97\textwidth]{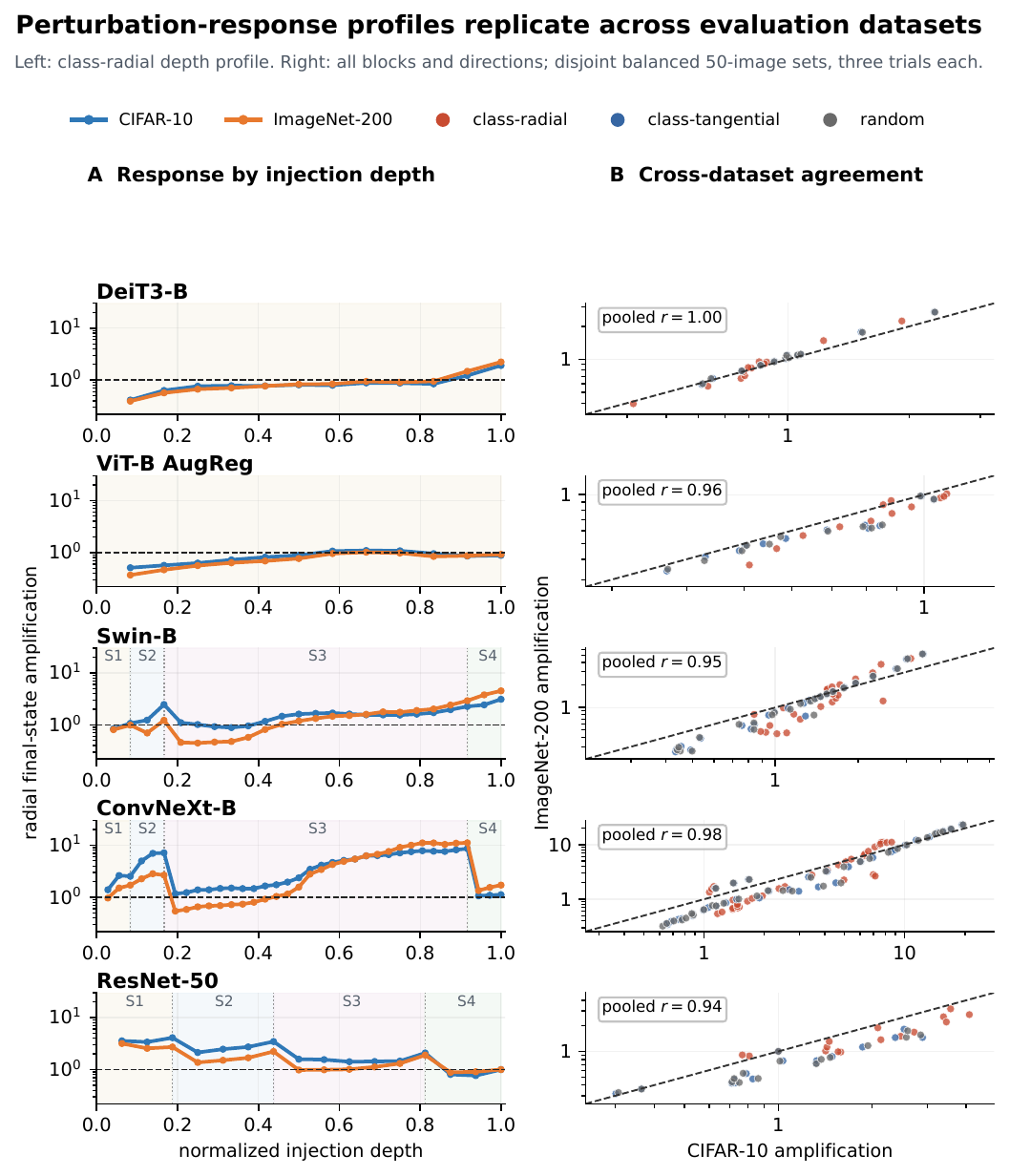}
\caption{\textbf{Perturbation-response schedules replicate across datasets.}
The same radial, tangential, and random channel interventions are injected at
every native block on disjoint CIFAR-10 and ImageNet-200 samples. Absolute
amplification is read within a checkpoint; the replicated depth profile is the
cross-dataset result.}
\label{fig:causal-replication}
\end{figure*}

\subsection{Texture-label replication}

DTD changes both image statistics and the semantic target: its 47 classes are
texture attributes rather than object identities. Nevertheless, native-block
movement, relational rewrite, and within-class innovation schedules transfer
strongly from the object datasets. Across checkpoint--reference pairs, median
DTD profile correlations are $0.993$ for movement, $0.933$ for rewrite, and
$0.912$ for within-class residual share. Prototype-accuracy gain is much less
stable ($0.350$), showing that mechanics recur more strongly than the depth at
which a new label space becomes separable.

The per-class analysis reaches the same conclusion at finer resolution. The
median cross-checkpoint rank correlation of formation order across 47 textures
is $0.381$ (class-bootstrap 95\% interval $[0.226,0.535]$) and remains $0.399$
after controlling for both checkpoints' final class difficulty. Variance
decomposition assigns $13.3\%$ to checkpoint, $32.4\%$ to texture class, and
$54.3\%$ to their interaction. Architecture therefore supplies a recurring
schedule, but class formation is not a universal depth clock; it is jointly
conditioned on checkpoint and label semantics.

\begin{figure*}[t]
\centering
\includegraphics[width=\textwidth]{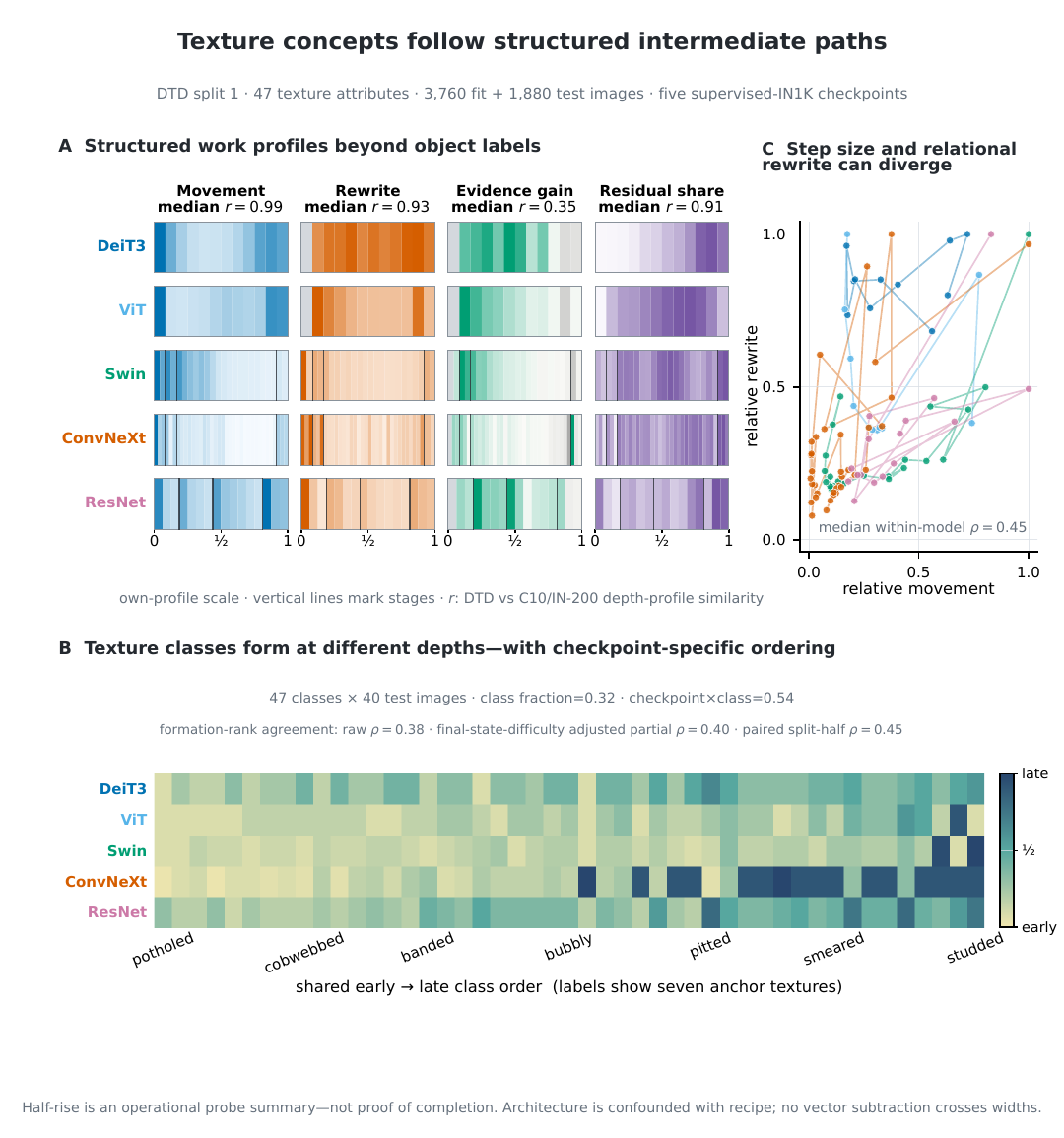}
\caption{\textbf{Native-block dynamics replicate on DTD, while semantic timing
changes with the label space.} The figure covers all 47 texture attributes and
the five representative checkpoints. Mechanical profiles remain strongly
correlated with object-dataset profiles; class-evidence timing is substantially
more task-dependent.}
\label{fig:dtd-replication}
\end{figure*}
\FloatBarrier
\section{Classification under Synthetic Corruption}
\label{app:classification-shift}

The matched screen behind Figure~\ref{fig:shift-generalization} uses exactly
the six checkpoints from clean and natural-shift classification.
Table~\ref{tab:cifar100c-matched-full} gives the complete main-paper summary:
final state and route accuracy, raw and control-adjusted gains, and positive-cell
counts over all 180 checkpoint--corruption--severity conditions. Every
checkpoint retains a positive control-adjusted mean; ResNet-50 is strongest,
while ViT-B shows that the effect is not CNN-specific. The few failures are
concentrated in CLIP/DeiT3 additive noise and the strongest contrast setting.

\begin{table}[t]
\centering
\caption{\textbf{Matched synthetic-shift summary behind Figure~\ref{fig:shift-generalization}b.} Means cover six corruptions, five severities, and three seeds. $\Delta_{\rm ctl}=\Delta_{\rm route}-\Delta_{\rm ctrl}$; positive R/C counts are route/control-adjusted conditions out of 30.}
\label{tab:cifar100c-matched-full}
\footnotesize
\setlength{\tabcolsep}{3.0pt}
\renewcommand{\arraystretch}{1.08}
\resizebox{\columnwidth}{!}{%
\begin{tabular}{lcrrrc}
\toprule
\rowcolor{headerblue}
\textbf{Checkpoint} & \textbf{End$\rightarrow$Route} & \boldmath$\Delta_{\rm route}$ & \boldmath$\Delta_{\rm ctrl}$ & \boldmath$\Delta_{\rm ctl}$ & \textbf{Positive R/C} \\
\midrule
\rowcolor{tablebluealt} CLIP-B & 60.79$\rightarrow$61.90 & \textcolor{insightaccent}{$+1.11$} & \textcolor{insightaccent}{$+0.12$} & \textcolor{insightaccent}{$\mathbf{+0.99}$} & 24/24 \\
\rowcolor{tableblue} DeiT3-B & 69.75$\rightarrow$71.06 & \textcolor{insightaccent}{$+1.32$} & \textcolor{insightaccent}{$+0.20$} & \textcolor{insightaccent}{$\mathbf{+1.12}$} & 27/26 \\
\rowcolor{tablebluealt} ViT-B & 58.52$\rightarrow$60.83 & \textcolor{insightaccent}{$+2.30$} & \textcolor{insightaccent}{$+0.07$} & \textcolor{insightaccent}{$\mathbf{+2.24}$} & 29/29 \\
\rowcolor{tableblue} SwinV2-B & 65.37$\rightarrow$66.91 & \textcolor{insightaccent}{$+1.54$} & \textcolor{insightaccent}{$+0.14$} & \textcolor{insightaccent}{$\mathbf{+1.40}$} & 30/30 \\
\rowcolor{tablebluealt} ConvNeXt-B & 62.92$\rightarrow$64.66 & \textcolor{insightaccent}{$+1.73$} & \textcolor{insightaccent}{$+0.37$} & \textcolor{insightaccent}{$\mathbf{+1.36}$} & 29/29 \\
\rowcolor{tableblue} ResNet-50 & 54.00$\rightarrow$57.66 & \textcolor{insightaccent}{$+3.67$} & \textcolor{insightaccent}{$+0.56$} & \textcolor{insightaccent}{$\mathbf{+3.11}$} & 30/30 \\
\midrule
\rowcolor{sectionblue} \textbf{Architecture mean} & \textbf{61.89$\rightarrow$63.84} & \textcolor{insightaccent}{$\mathbf{+1.94}$} & \textcolor{insightaccent}{$\mathbf{+0.24}$} & \textcolor{insightaccent}{$\mathbf{+1.70}$} & \textbf{169/168} \\
\bottomrule
\end{tabular}%
}
\end{table}

\FloatBarrier
\section{Complete OOD Results}
\label{app:ood-full}

\subsection{Alignment with the recent Mahalanobis++ evaluation}

Mahalanobis++ appeared at ICML 2025, so we treat it as a recent reference
rather than a long-established
reporting convention~\citep{mueller2025mahapp}. For a direct and auditable
comparison, we follow its OpenOOD evaluation and presentation choices wherever
the studies overlap: public frozen checkpoints, pre-logit features, $\ell_2$
normalization for Mahalanobis++, the prescribed individual OOD datasets, and
joint reporting of AUROC and FPR95. We also follow the model-row/method-column
structure used by its full appendix Tables 13--16 in our
Tables~\ref{tab:ood-full-mnist}--
\ref{tab:ood-full-imagenet1k}: checkpoints are rows, post-hoc detectors are
columns, and the complete grid is shown rather than only an architecture
average. The MSP and kNN columns are explicit, so every path result can be read
against the standard references displayed by Mahalanobis++.

This is protocol alignment, not a claim that the two papers evaluate identical
systems. Mahalanobis++ studies final-representation post-hoc detectors; our
tables hold the checkpoint, ID support, split, and seed fixed and then add the
same path-surprise score. MSP uses a checkpoint's native 1,000-class classifier
when the ImageNet label space and head are compatible. When the target labels
differ (MNIST/CIFAR) or an encoder exposes no compatible target classifier, we
fit the same ID-only linear probe used throughout this paper and average three
probe seeds. No OOD image is used to fit the probe, select a detector, or
choose a combination weight. Our
additional Relative Mahalanobis, X-Mahalanobis, kNN, and trajectory-density
columns test whether the path contribution persists beyond the particular
Mahalanobis++ partner.

\subsection{Layer-wise class and OOD neighbourhoods}

Figure~\ref{fig:layerwise-umap} separates the raw states from the coordinates
used by the method. In raw space, class organisation becomes clearer through
depth while OOD overlap increases at the final state. Subtracting the assigned
class path collapses the ID class centres, but near-OOD remains entangled with
the ID residual cloud and far-OOD retains off-route islands and tails. UMAP is
descriptive rather than a detector: the quantitative method uses the complete
sequence of residual transitions, not a two-dimensional projection.

\begin{figure*}[t]
\centering
\includegraphics[width=0.98\textwidth]{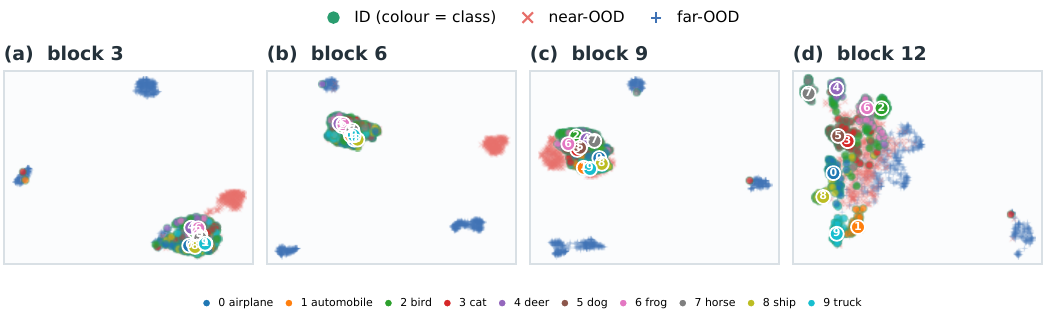}
\vspace{1.5mm}
\includegraphics[width=0.98\textwidth]{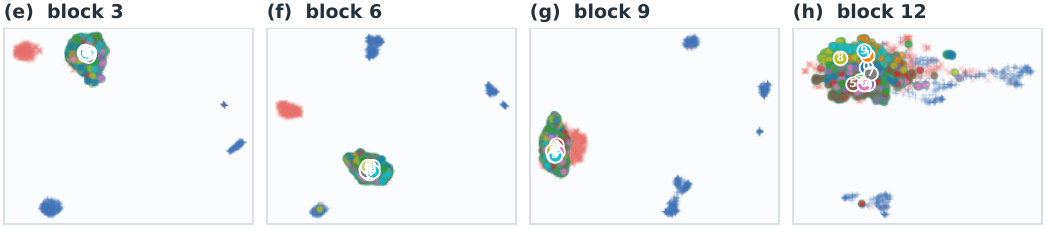}
\caption{\textbf{Raw and class-residual OOD neighbourhoods across depth.}
DINOv3-B on CIFAR-10 ID, with CIFAR-100/Tiny ImageNet as near-OOD and
MNIST/SVHN/DTD/Places365 as far-OOD. \emph{(a--d)} Raw states.
\emph{(e--h)} Residuals after assigning the nearest final state class and
subtracting its 1,000-example ID prototype path. Each group contains 600
balanced samples; colour and numbered medians identify true ID classes. A
separate UMAP is fitted at each block, so topology is compared within panels,
not by aligning axes across depth.}
\label{fig:layerwise-umap}
\end{figure*}

\begin{figure*}[t]
\centering
\begin{minipage}[t]{0.54\textwidth}
\centering
\includegraphics[width=\linewidth]{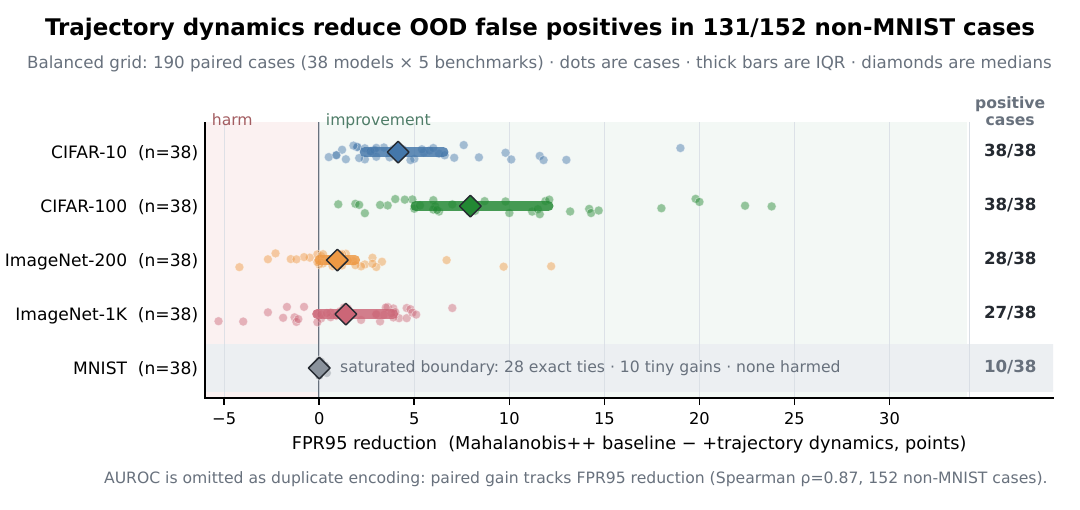}
\end{minipage}\hfill
\begin{minipage}[t]{0.43\textwidth}
\centering
\includegraphics[width=\linewidth]{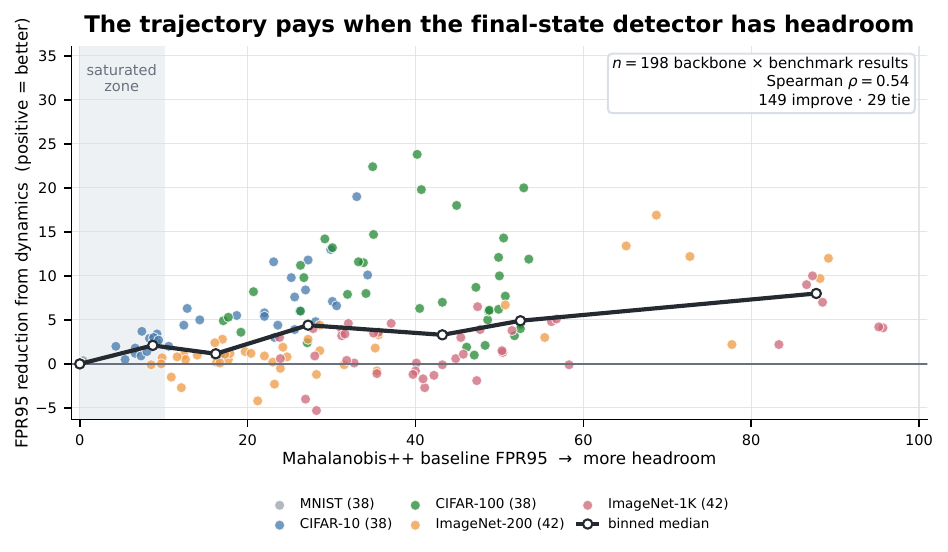}
\end{minipage}
\caption{\textbf{Exhaustive OOD distribution and detector headroom.} Left:
the balanced 190-case grid, including saturated MNIST. Right: all 198 stored
final-state/path comparisons after adding the eight non-common ImageNet rows,
plotted against final-state FPR95. The binned median exposes a moderate rather
than deterministic headroom trend; positive reduction means improvement.}
\label{fig:ood-full-distribution}
\end{figure*}

\paragraph{Cross-table synthesis.}
Across the 160 non-MNIST checkpoint--benchmark rows in
Tables~\ref{tab:ood-full-cifar10}--\ref{tab:ood-full-imagenet1k}, the path lowers
FPR95 in 150/160 Relative Mahalanobis pairs, 139/160 Mahalanobis++ pairs, and
135/160 kNN pairs. It remains positive in 114/160 X-Mahalanobis and 115/160
trajectory-Mahalanobis pairs, but with smaller mean reductions; those partners
already mix layers and therefore leave less novel route evidence. This ordering
is the checkpoint-level form of the complementarity controls in
Appendix~\ref{app:complementarity}.

The breadth is not supplied by one family. Pooling the five partners, positive
FPR95 changes occur in 26/30 self-supervised, 65/70 vision--language, 228/280
plain-ViT, 47/80 hierarchical-Transformer, and 287/340 CNN cells. These counts
demonstrate coverage, not a causal family ranking: checkpoint recipes and the
hard ImageNet near shifts remain confounded with architecture. Within-family
variation is consequently part of the result, rather than noise to be hidden
by one architecture average.

\begin{table*}[b]
\centering\scriptsize\setlength{\tabcolsep}{2.4pt}\renewcommand{\arraystretch}{1.00}
\caption{\textbf{Complete OpenOOD results: MNIST.} All 38 balanced checkpoints; final-representation FPR95 is saturated near zero.}
\label{tab:ood-full-mnist}
\vspace{0pt}
\resizebox{0.96\textwidth}{!}{%
\begin{tabular}{ll c cc cc cc cc cc}
\toprule
\rowcolor{headerblue}
\textbf{Model} & \textbf{Fam} & \textbf{MSP} & \textbf{kNN} & \textbf{+path} & \textbf{Maha++} & \textbf{+path} & \textbf{RelMaha} & \textbf{+path} & \textbf{X-Maha} & \textbf{+path} & \textbf{Traj-Maha} & \textbf{+path}\\
\midrule
\rowcolor{vlmheader}\multicolumn{13}{l}{\textbf{Self-supervised foundation encoders}}\\
\rowcolor{tableblue}DINOv2 ViT-B/14 & SSL & 96.2/21.3 & 99.8/0.7 & \textcolor{insightaccent}{\textbf{100.0/0.0}} & 99.9/0.1 & \textcolor{insightaccent}{\textbf{100.0/0.0}} & 99.9/0.1 & \textcolor{insightaccent}{\textbf{100.0/0.0}} & 100.0/0.1 & 100.0/0.0 & 100.0/0.0 & 100.0/0.0\\
\rowcolor{vlmheader}\multicolumn{13}{l}{\textbf{Vision--language encoders}}\\
\rowcolor{tableblue}CLIP ViT-L/14 & VLM & 96.1/21.5 & 100.0/0.0 & 100.0/0.0 & 100.0/0.0 & 100.0/0.0 & 100.0/0.0 & 100.0/0.0 & 100.0/0.0 & 100.0/0.0 & 100.0/0.0 & 100.0/0.0\\
\rowcolor{tablebluealt}CLIP-timm ViT-B/16 & VLM & 91.8/43.0 & 100.0/0.0 & \textcolor{insightaccent}{\textbf{100.0/0.0}} & 100.0/0.0 & 100.0/0.0 & 100.0/0.0 & 100.0/0.0 & 100.0/0.0 & 100.0/0.0 & 100.0/0.0 & 100.0/0.0\\
\rowcolor{vitheader}\multicolumn{13}{l}{\textbf{Supervised plain ViTs}}\\
\rowcolor{tableblue}BeiT-B/16 & ViT & 96.9/16.9 & 99.9/0.2 & \textcolor{insightaccent}{\textbf{100.0/0.0}} & 100.0/0.0 & 100.0/0.0 & 100.0/0.0 & 100.0/0.0 & 100.0/0.0 & 100.0/0.0 & 100.0/0.0 & 100.0/0.0\\
\rowcolor{tablebluealt}BeiT-L & ViT & 96.1/19.5 & 100.0/0.1 & \textcolor{insightaccent}{\textbf{100.0/0.0}} & 100.0/0.0 & 100.0/0.0 & 100.0/0.0 & 100.0/0.0 & 100.0/0.0 & 100.0/0.0 & 100.0/0.0 & 100.0/0.0\\
\rowcolor{tableblue}BeiTv2-B & ViT & 90.7/44.3 & 99.9/0.1 & \textcolor{insightaccent}{\textbf{100.0/0.0}} & 100.0/0.0 & 100.0/0.0 & 100.0/0.0 & 100.0/0.0 & 100.0/0.0 & 100.0/0.0 & 100.0/0.0 & 100.0/0.0\\
\rowcolor{tablebluealt}DeiT3-B/16 & ViT & 91.3/41.2 & 99.9/0.3 & \textcolor{insightaccent}{\textbf{100.0/0.0}} & 100.0/0.0 & 100.0/0.0 & 100.0/0.0 & 100.0/0.0 & 100.0/0.0 & 100.0/0.0 & 100.0/0.0 & 100.0/0.0\\
\rowcolor{tableblue}DeiT3-B/16-In21k & ViT & 94.2/29.0 & 99.6/2.2 & \textcolor{insightaccent}{\textbf{100.0/0.1}} & 99.9/0.3 & \textcolor{insightaccent}{\textbf{100.0/0.0}} & 99.9/0.1 & \textcolor{insightaccent}{\textbf{100.0/0.0}} & 100.0/0.0 & 100.0/0.0 & 100.0/0.0 & 100.0/0.0\\
\rowcolor{tablebluealt}DeiT3-L/16 & ViT & 92.8/35.7 & 100.0/0.2 & \textcolor{insightaccent}{\textbf{100.0/0.0}} & 100.0/0.0 & 100.0/0.0 & 100.0/0.0 & 100.0/0.0 & 100.0/0.0 & 100.0/0.0 & 100.0/0.0 & 100.0/0.0\\
\rowcolor{tableblue}DeiT3-S & ViT & 91.0/44.6 & 99.8/1.3 & \textcolor{insightaccent}{\textbf{100.0/0.0}} & 99.9/0.4 & \textcolor{insightaccent}{\textbf{100.0/0.0}} & 99.8/0.6 & \textcolor{insightaccent}{\textbf{100.0/0.0}} & 99.9/0.3 & \textcolor{insightaccent}{\textbf{100.0/0.0}} & 100.0/0.0 & 100.0/0.0\\
\rowcolor{tablebluealt}EVA02-B & ViT & 80.5/66.8 & 100.0/0.0 & 100.0/0.0 & 100.0/0.0 & 100.0/0.0 & 100.0/0.0 & 100.0/0.0 & 100.0/0.0 & 100.0/0.0 & 100.0/0.0 & 100.0/0.0\\
\rowcolor{tableblue}ViT-B-MAE & ViT & 20.9/99.9 & 100.0/0.0 & 100.0/0.0 & 100.0/0.0 & 100.0/0.0 & 100.0/0.0 & 100.0/0.0 & 100.0/0.0 & 100.0/0.0 & 100.0/0.0 & 100.0/0.0\\
\rowcolor{tablebluealt}ViT-B/16-augreg & ViT & 90.7/43.1 & 100.0/0.0 & 100.0/0.0 & 100.0/0.0 & 100.0/0.0 & 100.0/0.0 & 100.0/0.0 & 100.0/0.0 & 100.0/0.0 & 100.0/0.0 & 100.0/0.0\\
\rowcolor{tableblue}ViT-B/16-in1k & ViT & 86.9/56.5 & 100.0/0.0 & 100.0/0.0 & 100.0/0.0 & 100.0/0.0 & 100.0/0.0 & 100.0/0.0 & 100.0/0.0 & 100.0/0.0 & 100.0/0.0 & 100.0/0.0\\
\rowcolor{tablebluealt}ViT-B/32 & ViT & 90.8/42.1 & 100.0/0.0 & 100.0/0.0 & 100.0/0.0 & 100.0/0.0 & 100.0/0.0 & 100.0/0.0 & 100.0/0.0 & 100.0/0.0 & 100.0/0.0 & 100.0/0.0\\
\rowcolor{tableblue}ViT-L/16 & ViT & 93.0/35.5 & 100.0/0.0 & 100.0/0.0 & 100.0/0.0 & 100.0/0.0 & 100.0/0.0 & 100.0/0.0 & 100.0/0.0 & 100.0/0.0 & 100.0/0.0 & 100.0/0.0\\
\rowcolor{tablebluealt}ViT-S/16 & ViT & 83.6/66.4 & 100.0/0.0 & 100.0/0.0 & 100.0/0.0 & 100.0/0.0 & 100.0/0.0 & 100.0/0.0 & 100.0/0.0 & 100.0/0.0 & 100.0/0.0 & 100.0/0.0\\
\rowcolor{hierheader}\multicolumn{13}{l}{\textbf{Hierarchical Transformers}}\\
\rowcolor{tableblue}Swin-B & Swin & 95.8/20.3 & 100.0/0.0 & 100.0/0.0 & 100.0/0.0 & 100.0/0.0 & 100.0/0.0 & 100.0/0.0 & 100.0/0.0 & 100.0/0.0 & 100.0/0.0 & 100.0/0.0\\
\rowcolor{tablebluealt}SwinV2-B & Swin & 91.1/40.5 & 100.0/0.0 & \textcolor{insightaccent}{\textbf{100.0/0.0}} & 100.0/0.0 & 100.0/0.0 & 100.0/0.0 & 100.0/0.0 & 100.0/0.0 & 100.0/0.0 & 100.0/0.0 & 100.0/0.0\\
\rowcolor{tableblue}SwinV2-B-In21k & Swin & 93.8/32.3 & 100.0/0.0 & \textcolor{insightaccent}{\textbf{100.0/0.0}} & 100.0/0.0 & 100.0/0.0 & 100.0/0.0 & 100.0/0.0 & 100.0/0.0 & 100.0/0.0 & 100.0/0.0 & 100.0/0.0\\
\rowcolor{tablebluealt}SwinV2-T & Swin & 93.6/28.2 & 100.0/0.0 & \textcolor{insightaccent}{\textbf{100.0/0.0}} & 100.0/0.0 & 100.0/0.0 & 100.0/0.0 & 100.0/0.0 & 100.0/0.0 & 100.0/0.0 & 100.0/0.0 & 100.0/0.0\\
\rowcolor{cnnheader}\multicolumn{13}{l}{\textbf{Convolutional encoders}}\\
\rowcolor{tableblue}ConvNeXt-B & CNN & 95.1/26.8 & 100.0/0.2 & \textcolor{insightaccent}{\textbf{100.0/0.0}} & 100.0/0.0 & 100.0/0.0 & 100.0/0.0 & 100.0/0.0 & 100.0/0.0 & 100.0/0.0 & 100.0/0.0 & 100.0/0.0\\
\rowcolor{tablebluealt}ConvNeXt-B-In21k & CNN & 96.0/20.9 & 100.0/0.0 & \textcolor{insightaccent}{\textbf{100.0/0.0}} & 100.0/0.0 & 100.0/0.0 & 100.0/0.0 & 100.0/0.0 & 100.0/0.0 & 100.0/0.0 & 100.0/0.0 & 100.0/0.0\\
\rowcolor{tableblue}ConvNeXt-L & CNN & 93.8/29.8 & 99.9/0.1 & \textcolor{insightaccent}{\textbf{100.0/0.0}} & 100.0/0.0 & 100.0/0.0 & 100.0/0.0 & 100.0/0.0 & 100.0/0.0 & 100.0/0.0 & 100.0/0.0 & 100.0/0.0\\
\rowcolor{tablebluealt}ConvNeXt-T & CNN & 94.8/26.4 & 100.0/0.0 & 100.0/0.0 & 100.0/0.0 & 100.0/0.0 & 100.0/0.0 & 100.0/0.0 & 100.0/0.0 & 100.0/0.0 & 100.0/0.0 & 100.0/0.0\\
\rowcolor{tableblue}ConvNeXtV2-B & CNN & 92.8/37.4 & 99.8/1.2 & \textcolor{insightaccent}{\textbf{99.9/0.5}} & 100.0/0.0 & 100.0/0.0 & 100.0/0.0 & 100.0/0.0 & 99.9/0.1 & 100.0/0.1 & 100.0/0.0 & 100.0/0.0\\
\rowcolor{tablebluealt}ConvNeXtV2-B-In21k & CNN & 96.9/16.8 & 99.9/0.2 & \textcolor{insightaccent}{\textbf{100.0/0.1}} & 100.0/0.0 & 100.0/0.0 & 100.0/0.0 & 100.0/0.0 & 100.0/0.0 & 100.0/0.0 & 100.0/0.0 & 100.0/0.0\\
\rowcolor{tableblue}ConvNeXtV2-T & CNN & 94.3/31.2 & 100.0/0.0 & \textcolor{insightaccent}{\textbf{100.0/0.0}} & 100.0/0.0 & 100.0/0.0 & 100.0/0.0 & 100.0/0.0 & 100.0/0.0 & 100.0/0.0 & 100.0/0.0 & 100.0/0.0\\
\rowcolor{tablebluealt}DenseNet-121 & CNN & 87.0/56.6 & 100.0/0.1 & \textcolor{insightaccent}{\textbf{100.0/0.0}} & 100.0/0.0 & 100.0/0.0 & 100.0/0.0 & 100.0/0.0 & 100.0/0.0 & 100.0/0.0 & 100.0/0.0 & 100.0/0.0\\
\rowcolor{tableblue}EffNetV2-L & CNN & 97.0/17.2 & 99.9/0.2 & \textcolor{insightaccent}{\textbf{100.0/0.0}} & 100.0/0.1 & 100.0/0.0 & 100.0/0.1 & 100.0/0.0 & 100.0/0.0 & 100.0/0.0 & 100.0/0.0 & 100.0/0.0\\
\rowcolor{tablebluealt}EffNetV2-M & CNN & 93.0/37.7 & 100.0/0.1 & \textcolor{insightaccent}{\textbf{100.0/0.0}} & 100.0/0.1 & 100.0/0.0 & 100.0/0.1 & 100.0/0.0 & 100.0/0.0 & 100.0/0.0 & 100.0/0.0 & 100.0/0.0\\
\rowcolor{tableblue}EffNetV2-S & CNN & 94.9/27.5 & 100.0/0.1 & \textcolor{insightaccent}{\textbf{100.0/0.0}} & 100.0/0.1 & 100.0/0.0 & 100.0/0.1 & 100.0/0.0 & 100.0/0.0 & 100.0/0.0 & 100.0/0.0 & 100.0/0.0\\
\rowcolor{tablebluealt}RegNetY-040 & CNN & 87.8/57.5 & 100.0/0.0 & \textcolor{insightaccent}{\textbf{100.0/0.0}} & 100.0/0.0 & 100.0/0.0 & 100.0/0.0 & 100.0/0.0 & 100.0/0.0 & 100.0/0.0 & 100.0/0.0 & 100.0/0.0\\
\rowcolor{tableblue}ResNet-101 & CNN & 96.2/16.0 & 100.0/0.0 & \textcolor{insightaccent}{\textbf{100.0/0.0}} & 100.0/0.1 & 100.0/0.0 & 99.9/0.5 & \textcolor{insightaccent}{\textbf{100.0/0.0}} & 100.0/0.1 & 100.0/0.0 & 100.0/0.0 & 100.0/0.0\\
\rowcolor{tablebluealt}ResNet-152 & CNN & 90.5/50.2 & 100.0/0.0 & \textcolor{insightaccent}{\textbf{100.0/0.0}} & 100.0/0.1 & 100.0/0.0 & 99.9/0.2 & \textcolor{insightaccent}{\textbf{100.0/0.0}} & 100.0/0.1 & 100.0/0.0 & 100.0/0.0 & 100.0/0.0\\
\rowcolor{tableblue}ResNet-50 & CNN & 94.5/29.6 & 100.0/0.0 & \textcolor{insightaccent}{\textbf{100.0/0.0}} & 100.0/0.1 & 100.0/0.0 & 100.0/0.2 & 100.0/0.0 & 100.0/0.0 & 100.0/0.0 & 100.0/0.0 & 100.0/0.0\\
\rowcolor{tablebluealt}ResNetV2-50 & CNN & 95.0/23.3 & 100.0/0.0 & \textcolor{insightaccent}{\textbf{100.0/0.0}} & 100.0/0.1 & 100.0/0.0 & 99.9/0.2 & \textcolor{insightaccent}{\textbf{100.0/0.0}} & 100.0/0.0 & 100.0/0.0 & 100.0/0.0 & 100.0/0.0\\
\rowcolor{tableblue}WideResNet50-2 & CNN & 95.9/23.0 & 100.0/0.0 & 100.0/0.0 & 100.0/0.0 & 100.0/0.0 & 100.0/0.0 & 100.0/0.0 & 100.0/0.0 & 100.0/0.0 & 100.0/0.0 & 100.0/0.0\\
\bottomrule
\end{tabular}%
}
\end{table*}

\begingroup
\small
\paragraph{The partner changes which architecture benefits.}
Relative Mahalanobis is the most stable density partner: path evidence lowers
FPR95 in 66/68 CNN rows, 54/56 plain-ViT rows, and every SSL and VLM row.  The
trajectory-density partner behaves differently.  It improves all 68 CNN rows,
but only 27/56 plain-ViT and 2/6 SSL rows.  This is an informative redundancy
test rather than a contradiction: trajectory density already consumes a
multi-layer summary, so it leaves little unused path signal when a Transformer's
states are strongly aligned; CNN residual coordinates still supply distinct
local transitions.  kNN shows the complementary pattern---all 56 plain-ViT
rows improve, compared with 49/68 CNN and 10/16 hierarchical-Transformer rows.
No single final-state witness is therefore uniformly optimal for exposing path
value.

\paragraph{Coverage and effect size are separate.}
CIFAR-100 has the broadest checkpoint-level support (182/190 positive
score--checkpoint cells), followed by CIFAR-10 (167/190), ImageNet-200
(154/210), and ImageNet-1K (150/210).  Yet this ordering is not simply dataset
scale: the two ImageNet protocols have similar coverage despite their different
label counts, and their failures concentrate on the same semantically near
splits.  At model level, DeiT3-S and ResNet-50 improve in all 20 available
non-MNIST benchmark--partner cells, but their routes are not mechanically
similar elsewhere in the paper.  Conversely, ViT-B-MAE improves in 19/20 cells
with a much smaller mean reduction.  Thus unanimity measures robustness, not
how much recoverable information a path contains; both statistics are needed
to interpret the full tables.
\endgroup

\newpage
\begin{table*}[p]
\centering\scriptsize\setlength{\tabcolsep}{2.4pt}\renewcommand{\arraystretch}{1.08}
\caption{\textbf{Complete OpenOOD results: CIFAR-10.} 38 checkpoints and six OOD datasets; fit support is 1,000 images per class (2,000 for the four IN21k rows).}
\label{tab:ood-full-cifar10}
\vspace{0pt}
\resizebox{\textwidth}{!}{%
\begin{tabular}{ll c cc cc cc cc cc}
\toprule
\rowcolor{headerblue}
\textbf{Model} & \textbf{Fam} & \textbf{MSP} & \textbf{kNN} & \textbf{+path} & \textbf{Maha++} & \textbf{+path} & \textbf{RelMaha} & \textbf{+path} & \textbf{X-Maha} & \textbf{+path} & \textbf{Traj-Maha} & \textbf{+path}\\
\midrule
\rowcolor{vlmheader}\multicolumn{13}{l}{\textbf{Self-supervised foundation encoders}}\\
\rowcolor{tableblue}DINOv2 ViT-B/14 & SSL & 99.0/3.5 & 98.2/10.3 & \textcolor{insightaccent}{\textbf{99.6/1.9}} & 99.0/4.3 & \textcolor{insightaccent}{\textbf{99.5/2.3}} & 98.5/4.7 & \textcolor{insightaccent}{\textbf{99.5/1.8}} & 99.2/3.5 & \textcolor{insightaccent}{\textbf{99.5/2.2}} & 99.0/4.2 & 99.0/4.4\\
\rowcolor{vlmheader}\multicolumn{13}{l}{\textbf{Vision--language encoders}}\\
\rowcolor{tableblue}CLIP ViT-L/14 & VLM & 96.8/12.4 & 96.8/12.6 & \textcolor{insightaccent}{\textbf{97.5/9.5}} & 97.1/10.6 & \textcolor{insightaccent}{\textbf{97.5/8.6}} & 98.4/8.6 & \textcolor{insightaccent}{\textbf{99.3/3.7}} & 97.5/8.9 & 97.5/9.3 & 97.2/9.7 & 97.2/10.1\\
\rowcolor{tablebluealt}CLIP-timm ViT-B/16 & VLM & 94.5/24.3 & 93.5/25.4 & \textcolor{insightaccent}{\textbf{96.4/9.7}} & 95.3/14.3 & \textcolor{insightaccent}{\textbf{96.5/9.3}} & 97.1/15.8 & \textcolor{insightaccent}{\textbf{98.6/6.8}} & 96.3/10.0 & \textcolor{insightaccent}{\textbf{96.6/9.2}} & 96.5/9.6 & 96.6/9.7\\
\rowcolor{vitheader}\multicolumn{13}{l}{\textbf{Supervised plain ViTs}}\\
\rowcolor{tableblue}BeiT-B/16 & ViT & 98.5/5.1 & 98.5/6.2 & \textcolor{insightaccent}{\textbf{99.2/4.4}} & 98.2/8.8 & \textcolor{insightaccent}{\textbf{98.8/6.1}} & 98.5/5.0 & \textcolor{insightaccent}{\textbf{99.4/3.0}} & 98.8/5.6 & 98.8/5.5 & 98.5/6.2 & 98.5/6.3\\
\rowcolor{tablebluealt}BeiT-L & ViT & 98.9/3.7 & 96.9/19.3 & \textcolor{insightaccent}{\textbf{99.1/5.5}} & 98.2/9.2 & \textcolor{insightaccent}{\textbf{98.9/5.8}} & 99.0/2.8 & \textcolor{insightaccent}{\textbf{99.6/1.2}} & 99.2/4.4 & 99.2/4.4 & 99.0/5.2 & 99.0/5.0\\
\rowcolor{tableblue}BeiTv2-B & ViT & 98.3/6.3 & 98.5/6.9 & \textcolor{insightaccent}{\textbf{99.2/4.6}} & 98.8/6.6 & \textcolor{insightaccent}{\textbf{99.1/4.8}} & 98.5/4.8 & \textcolor{insightaccent}{\textbf{99.5/2.1}} & 99.2/4.2 & 99.2/4.1 & 98.8/5.5 & 98.8/5.6\\
\rowcolor{tablebluealt}DeiT3-B/16 & ViT & 96.7/15.2 & 92.8/41.1 & \textcolor{insightaccent}{\textbf{97.5/11.7}} & 94.8/25.2 & \textcolor{insightaccent}{\textbf{97.2/15.4}} & 97.0/12.8 & \textcolor{insightaccent}{\textbf{98.6/6.7}} & 96.2/19.7 & \textcolor{insightaccent}{\textbf{97.7/11.7}} & 97.4/13.3 & \textcolor{insightaccent}{\textbf{98.0/9.6}}\\
\rowcolor{tableblue}DeiT3-B/16-In21k & ViT & 98.1/7.3 & 98.3/6.2 & \textcolor{insightaccent}{\textbf{99.4/3.3}} & 98.3/7.4 & \textcolor{insightaccent}{\textbf{99.2/3.7}} & 98.2/7.3 & \textcolor{insightaccent}{\textbf{99.4/2.5}} & 98.8/5.6 & \textcolor{insightaccent}{\textbf{99.2/3.6}} & 99.0/4.3 & 99.0/4.2\\
\rowcolor{tablebluealt}DeiT3-L/16 & ViT & 97.8/8.9 & 90.0/53.0 & \textcolor{insightaccent}{\textbf{97.2/13.0}} & 95.3/23.1 & \textcolor{insightaccent}{\textbf{97.5/11.5}} & 97.4/10.5 & \textcolor{insightaccent}{\textbf{98.8/5.8}} & 97.0/14.2 & \textcolor{insightaccent}{\textbf{98.0/8.7}} & 97.9/8.7 & \textcolor{insightaccent}{\textbf{98.3/7.2}}\\
\rowcolor{tableblue}DeiT3-S & ViT & 95.7/21.9 & 93.4/30.6 & \textcolor{insightaccent}{\textbf{97.2/12.5}} & 93.6/25.6 & \textcolor{insightaccent}{\textbf{96.7/18.0}} & 94.6/28.3 & \textcolor{insightaccent}{\textbf{97.9/11.6}} & 93.9/24.7 & \textcolor{insightaccent}{\textbf{96.8/17.1}} & 97.5/10.4 & \textcolor{insightaccent}{\textbf{97.7/8.8}}\\
\rowcolor{tablebluealt}EVA02-B & ViT & 97.8/8.4 & 93.0/24.0 & \textcolor{insightaccent}{\textbf{97.0/13.9}} & 98.1/8.3 & \textcolor{insightaccent}{\textbf{98.6/5.4}} & 98.4/7.5 & \textcolor{insightaccent}{\textbf{99.2/4.0}} & 98.6/5.2 & \textcolor{insightaccent}{\textbf{98.7/4.7}} & 98.7/4.5 & 98.7/4.5\\
\rowcolor{tableblue}ViT-B-MAE & ViT & 56.5/94.1 & 90.0/36.0 & \textcolor{insightaccent}{\textbf{91.9/30.6}} & 93.7/17.1 & \textcolor{insightaccent}{\textbf{94.0/16.2}} & 95.3/15.5 & \textcolor{insightaccent}{\textbf{95.4/15.0}} & 93.9/16.6 & \textcolor{insightaccent}{\textbf{94.2/16.1}} & 94.5/14.8 & 94.4/15.2\\
\rowcolor{tablebluealt}ViT-B/16-augreg & ViT & 98.1/7.6 & 98.2/8.6 & \textcolor{insightaccent}{\textbf{98.8/5.9}} & 98.5/6.6 & \textcolor{insightaccent}{\textbf{98.7/5.4}} & 98.3/6.9 & \textcolor{insightaccent}{\textbf{99.4/2.4}} & 98.6/5.9 & 98.5/6.4 & 98.2/7.5 & 98.0/8.2\\
\rowcolor{tableblue}ViT-B/16-in1k & ViT & 93.6/33.9 & 93.7/38.4 & \textcolor{insightaccent}{\textbf{97.3/10.0}} & 93.9/33.0 & \textcolor{insightaccent}{\textbf{96.5/14.0}} & 95.3/24.2 & \textcolor{insightaccent}{\textbf{98.2/8.3}} & 96.1/15.7 & \textcolor{insightaccent}{\textbf{96.7/11.8}} & 96.8/10.3 & \textcolor{insightaccent}{\textbf{97.0/9.9}}\\
\rowcolor{tablebluealt}ViT-B/32 & ViT & 97.7/9.3 & 97.4/10.9 & \textcolor{insightaccent}{\textbf{98.2/7.0}} & 97.1/12.4 & \textcolor{insightaccent}{\textbf{98.0/8.0}} & 97.3/11.6 & \textcolor{insightaccent}{\textbf{99.2/3.6}} & 97.4/9.6 & \textcolor{insightaccent}{\textbf{97.5/8.9}} & 97.7/8.2 & 97.6/8.6\\
\rowcolor{tableblue}ViT-L/16 & ViT & 98.7/4.5 & 98.8/5.9 & \textcolor{insightaccent}{\textbf{99.1/4.8}} & 98.8/5.4 & \textcolor{insightaccent}{\textbf{99.0/4.9}} & 99.3/3.3 & \textcolor{insightaccent}{\textbf{99.7/0.7}} & 98.2/7.4 & 98.1/7.7 & 98.6/6.7 & 98.4/7.2\\
\rowcolor{tablebluealt}ViT-S/16 & ViT & 96.9/13.6 & 96.7/15.9 & \textcolor{insightaccent}{\textbf{98.4/7.1}} & 97.3/12.8 & \textcolor{insightaccent}{\textbf{98.3/6.5}} & 96.9/13.9 & \textcolor{insightaccent}{\textbf{99.1/4.8}} & 97.9/8.6 & \textcolor{insightaccent}{\textbf{98.1/7.1}} & 97.7/8.3 & 97.6/8.8\\
\rowcolor{hierheader}\multicolumn{13}{l}{\textbf{Hierarchical Transformers}}\\
\rowcolor{tableblue}Swin-B & Swin & 97.8/8.6 & 98.3/7.8 & \textcolor{insightaccent}{\textbf{98.6/7.7}} & 98.3/9.2 & \textcolor{insightaccent}{\textbf{98.7/6.8}} & 98.0/7.5 & \textcolor{insightaccent}{\textbf{99.0/4.9}} & 97.3/11.1 & \textcolor{insightaccent}{\textbf{97.4/10.9}} & 97.4/10.9 & \textcolor{insightaccent}{\textbf{97.8/9.3}}\\
\rowcolor{tablebluealt}SwinV2-B & Swin & 96.3/16.2 & 95.0/29.6 & \textcolor{insightaccent}{\textbf{97.8/9.9}} & 94.1/27.2 & \textcolor{insightaccent}{\textbf{97.0/15.4}} & 97.0/15.9 & \textcolor{insightaccent}{\textbf{98.6/7.2}} & 98.1/7.2 & 98.1/7.4 & 97.7/9.4 & \textcolor{insightaccent}{\textbf{98.3/7.2}}\\
\rowcolor{tableblue}SwinV2-B-In21k & Swin & 97.7/8.5 & 98.3/7.7 & \textcolor{insightaccent}{\textbf{98.8/6.4}} & 98.2/8.8 & \textcolor{insightaccent}{\textbf{98.7/6.7}} & 98.4/6.7 & \textcolor{insightaccent}{\textbf{99.4/2.8}} & 98.8/5.7 & 98.7/6.3 & 99.0/4.8 & 98.9/5.2\\
\rowcolor{tablebluealt}SwinV2-T & Swin & 94.6/26.8 & 94.5/29.0 & \textcolor{insightaccent}{\textbf{97.4/11.6}} & 93.4/29.9 & \textcolor{insightaccent}{\textbf{96.5/16.9}} & 94.7/29.2 & \textcolor{insightaccent}{\textbf{98.0/9.7}} & 96.8/9.7 & 96.9/9.7 & 96.7/10.8 & \textcolor{insightaccent}{\textbf{97.2/9.0}}\\
\rowcolor{cnnheader}\multicolumn{13}{l}{\textbf{Convolutional encoders}}\\
\rowcolor{tableblue}ConvNeXt-B & CNN & 95.9/18.8 & 94.3/28.0 & \textcolor{insightaccent}{\textbf{96.0/21.6}} & 92.4/28.1 & \textcolor{insightaccent}{\textbf{94.8/23.3}} & 97.1/15.2 & \textcolor{insightaccent}{\textbf{97.7/11.3}} & 90.8/39.1 & \textcolor{insightaccent}{\textbf{93.6/31.4}} & 91.1/37.1 & \textcolor{insightaccent}{\textbf{94.3/32.4}}\\
\rowcolor{tablebluealt}ConvNeXt-B-In21k & CNN & 98.0/8.6 & 98.7/6.0 & 98.5/7.0 & 98.6/7.3 & \textcolor{insightaccent}{\textbf{98.7/6.4}} & 98.4/6.3 & \textcolor{insightaccent}{\textbf{98.8/5.2}} & 98.3/8.9 & \textcolor{insightaccent}{\textbf{98.4/8.3}} & 92.0/32.5 & \textcolor{insightaccent}{\textbf{95.9/23.0}}\\
\rowcolor{tableblue}ConvNeXt-L & CNN & 96.0/19.8 & 95.2/27.2 & \textcolor{insightaccent}{\textbf{96.7/15.1}} & 94.0/25.6 & \textcolor{insightaccent}{\textbf{95.9/21.7}} & 97.0/17.6 & \textcolor{insightaccent}{\textbf{97.6/13.6}} & 92.2/34.9 & \textcolor{insightaccent}{\textbf{94.5/29.1}} & 89.8/37.1 & \textcolor{insightaccent}{\textbf{94.3/31.0}}\\
\rowcolor{tablebluealt}ConvNeXt-T & CNN & 95.0/24.2 & 92.5/33.2 & \textcolor{insightaccent}{\textbf{95.3/23.2}} & 91.5/30.1 & \textcolor{insightaccent}{\textbf{94.3/23.0}} & 94.9/28.6 & \textcolor{insightaccent}{\textbf{97.2/15.1}} & 89.5/39.3 & \textcolor{insightaccent}{\textbf{93.2/29.3}} & 93.2/25.1 & \textcolor{insightaccent}{\textbf{96.0/12.9}}\\
\rowcolor{tableblue}ConvNeXtV2-B & CNN & 96.9/14.3 & 95.6/22.9 & \textcolor{insightaccent}{\textbf{96.7/18.3}} & 94.2/23.2 & \textcolor{insightaccent}{\textbf{95.8/20.2}} & 97.3/13.2 & \textcolor{insightaccent}{\textbf{97.9/10.1}} & 92.0/35.0 & \textcolor{insightaccent}{\textbf{94.1/29.7}} & 87.3/39.5 & \textcolor{insightaccent}{\textbf{92.3/35.8}}\\
\rowcolor{tablebluealt}ConvNeXtV2-B-In21k & CNN & 98.4/6.4 & 98.5/7.4 & \textcolor{insightaccent}{\textbf{98.8/6.0}} & 98.5/8.7 & \textcolor{insightaccent}{\textbf{98.8/6.3}} & 98.5/5.8 & \textcolor{insightaccent}{\textbf{99.0/4.5}} & 98.3/9.2 & \textcolor{insightaccent}{\textbf{98.7/7.2}} & 76.7/48.0 & \textcolor{insightaccent}{\textbf{92.7/40.2}}\\
\rowcolor{tableblue}ConvNeXtV2-T & CNN & 95.2/22.8 & 93.1/31.9 & \textcolor{insightaccent}{\textbf{94.9/23.7}} & 91.2/30.6 & \textcolor{insightaccent}{\textbf{93.5/24.0}} & 95.4/23.8 & \textcolor{insightaccent}{\textbf{96.8/17.1}} & 91.1/34.1 & \textcolor{insightaccent}{\textbf{93.2/27.4}} & 82.4/40.1 & \textcolor{insightaccent}{\textbf{91.3/36.1}}\\
\rowcolor{tablebluealt}DenseNet-121 & CNN & 89.9/43.6 & 93.7/22.6 & \textcolor{insightaccent}{\textbf{95.3/18.5}} & 92.8/23.6 & \textcolor{insightaccent}{\textbf{94.5/19.2}} & 95.6/21.5 & \textcolor{insightaccent}{\textbf{96.9/14.3}} & 93.2/21.8 & \textcolor{insightaccent}{\textbf{94.5/19.1}} & 93.8/15.1 & \textcolor{insightaccent}{\textbf{94.2/14.9}}\\
\rowcolor{tableblue}EffNetV2-L & CNN & 97.3/11.2 & 98.6/5.9 & 98.7/6.0 & 98.2/8.0 & \textcolor{insightaccent}{\textbf{98.4/6.6}} & 98.1/8.6 & \textcolor{insightaccent}{\textbf{99.1/3.8}} & 98.7/5.5 & 98.5/6.0 & 93.7/18.1 & \textcolor{insightaccent}{\textbf{94.8/17.0}}\\
\rowcolor{tablebluealt}EffNetV2-M & CNN & 97.3/13.0 & 98.5/5.4 & \textcolor{insightaccent}{\textbf{98.9/4.7}} & 98.0/8.8 & \textcolor{insightaccent}{\textbf{98.6/5.8}} & 97.7/10.9 & \textcolor{insightaccent}{\textbf{99.0/4.3}} & 98.7/5.3 & 98.7/5.5 & 93.9/16.7 & \textcolor{insightaccent}{\textbf{95.0/15.4}}\\
\rowcolor{tableblue}EffNetV2-S & CNN & 96.9/15.4 & 98.4/7.0 & \textcolor{insightaccent}{\textbf{98.9/4.9}} & 97.9/9.4 & \textcolor{insightaccent}{\textbf{98.5/6.7}} & 98.0/9.8 & \textcolor{insightaccent}{\textbf{99.0/4.6}} & 98.0/8.9 & \textcolor{insightaccent}{\textbf{98.3/7.1}} & 93.7/18.7 & \textcolor{insightaccent}{\textbf{94.9/16.6}}\\
\rowcolor{tablebluealt}RegNetY-040 & CNN & 91.0/42.2 & 96.2/14.7 & \textcolor{insightaccent}{\textbf{97.2/10.2}} & 95.0/18.7 & \textcolor{insightaccent}{\textbf{96.3/13.2}} & 93.0/35.4 & \textcolor{insightaccent}{\textbf{97.1/13.8}} & 95.1/18.3 & \textcolor{insightaccent}{\textbf{96.2/13.6}} & 94.7/13.3 & \textcolor{insightaccent}{\textbf{95.2/12.9}}\\
\rowcolor{tableblue}ResNet-101 & CNN & 95.4/23.1 & 97.0/12.0 & \textcolor{insightaccent}{\textbf{97.7/9.3}} & 94.9/22.0 & \textcolor{insightaccent}{\textbf{96.5/16.2}} & 96.4/19.5 & \textcolor{insightaccent}{\textbf{97.8/11.5}} & 95.0/21.4 & \textcolor{insightaccent}{\textbf{96.5/15.6}} & 93.5/17.3 & \textcolor{insightaccent}{\textbf{94.9/15.9}}\\
\rowcolor{tablebluealt}ResNet-152 & CNN & 96.3/17.8 & 96.9/14.9 & \textcolor{insightaccent}{\textbf{98.0/9.0}} & 94.4/22.0 & \textcolor{insightaccent}{\textbf{96.6/16.6}} & 96.6/17.7 & \textcolor{insightaccent}{\textbf{98.1/9.6}} & 94.4/22.3 & \textcolor{insightaccent}{\textbf{96.7/16.7}} & 93.7/16.8 & \textcolor{insightaccent}{\textbf{95.4/14.9}}\\
\rowcolor{tableblue}ResNet-50 & CNN & 90.8/44.0 & 94.3/25.2 & \textcolor{insightaccent}{\textbf{96.0/15.9}} & 88.4/34.3 & \textcolor{insightaccent}{\textbf{93.1/24.2}} & 93.1/34.2 & \textcolor{insightaccent}{\textbf{96.2/19.7}} & 89.7/31.1 & \textcolor{insightaccent}{\textbf{93.5/23.0}} & 93.8/16.9 & \textcolor{insightaccent}{\textbf{94.5/16.4}}\\
\rowcolor{tablebluealt}ResNetV2-50 & CNN & 93.4/31.2 & 96.0/17.8 & \textcolor{insightaccent}{\textbf{97.3/10.2}} & 92.7/26.2 & \textcolor{insightaccent}{\textbf{95.5/20.2}} & 94.8/28.4 & \textcolor{insightaccent}{\textbf{97.1/15.2}} & 96.2/13.6 & \textcolor{insightaccent}{\textbf{96.5/10.9}} & 93.8/15.7 & \textcolor{insightaccent}{\textbf{94.8/14.7}}\\
\rowcolor{tableblue}WideResNet50-2 & CNN & 94.2/27.9 & 95.9/19.0 & \textcolor{insightaccent}{\textbf{97.0/11.7}} & 93.3/26.9 & \textcolor{insightaccent}{\textbf{95.7/18.5}} & 94.4/30.5 & \textcolor{insightaccent}{\textbf{97.0/15.4}} & 94.2/24.2 & \textcolor{insightaccent}{\textbf{96.0/16.5}} & 93.8/15.7 & \textcolor{insightaccent}{\textbf{94.7/15.4}}\\
\bottomrule
\end{tabular}%
}
\end{table*}

\newpage
\begin{table*}[p]
\centering\scriptsize\setlength{\tabcolsep}{2.4pt}\renewcommand{\arraystretch}{1.08}
\caption{\textbf{Complete OpenOOD results: CIFAR-100.} 38 checkpoints, 200 fit images per class, and six OOD datasets.}
\label{tab:ood-full-cifar100}
\vspace{0pt}
\resizebox{\textwidth}{!}{%
\begin{tabular}{ll c cc cc cc cc cc}
\toprule
\rowcolor{headerblue}
\textbf{Model} & \textbf{Fam} & \textbf{MSP} & \textbf{kNN} & \textbf{+path} & \textbf{Maha++} & \textbf{+path} & \textbf{RelMaha} & \textbf{+path} & \textbf{X-Maha} & \textbf{+path} & \textbf{Traj-Maha} & \textbf{+path}\\
\midrule
\rowcolor{vlmheader}\multicolumn{13}{l}{\textbf{Self-supervised foundation encoders}}\\
\rowcolor{tableblue}DINOv2 ViT-B/14 & SSL & 91.3/33.1 & 89.0/41.3 & \textcolor{insightaccent}{\textbf{98.2/9.4}} & 92.1/34.9 & \textcolor{insightaccent}{\textbf{97.9/12.5}} & 90.3/38.8 & \textcolor{insightaccent}{\textbf{97.7/14.4}} & 95.4/25.5 & \textcolor{insightaccent}{\textbf{98.4/8.3}} & 97.2/12.2 & 97.1/11.5\\
\rowcolor{vlmheader}\multicolumn{13}{l}{\textbf{Vision--language encoders}}\\
\rowcolor{tableblue}CLIP ViT-L/14 & VLM & 81.9/73.0 & 92.7/28.3 & \textcolor{insightaccent}{\textbf{94.8/17.1}} & 94.8/19.2 & \textcolor{insightaccent}{\textbf{95.5/15.6}} & 94.5/28.3 & \textcolor{insightaccent}{\textbf{98.2/9.5}} & 95.3/14.6 & 95.0/14.7 & 94.2/15.1 & 94.0/15.2\\
\rowcolor{tablebluealt}CLIP-timm ViT-B/16 & VLM & 73.6/91.2 & 87.1/43.8 & \textcolor{insightaccent}{\textbf{93.5/23.9}} & 92.3/26.7 & \textcolor{insightaccent}{\textbf{94.4/16.9}} & 92.4/36.6 & \textcolor{insightaccent}{\textbf{96.6/13.8}} & 94.2/18.4 & \textcolor{insightaccent}{\textbf{94.6/15.0}} & 93.8/15.4 & 93.8/15.2\\
\rowcolor{vitheader}\multicolumn{13}{l}{\textbf{Supervised plain ViTs}}\\
\rowcolor{tableblue}BeiT-B/16 & ViT & 92.9/32.8 & 93.3/29.9 & \textcolor{insightaccent}{\textbf{96.5/13.2}} & 94.4/26.3 & \textcolor{insightaccent}{\textbf{96.4/15.1}} & 93.7/31.1 & \textcolor{insightaccent}{\textbf{97.3/13.8}} & 95.9/15.8 & 95.9/14.7 & 95.3/14.9 & 95.3/14.9\\
\rowcolor{tablebluealt}BeiT-L & ViT & 93.8/26.5 & 90.5/44.0 & \textcolor{insightaccent}{\textbf{96.8/14.9}} & 94.1/29.2 & \textcolor{insightaccent}{\textbf{96.9/15.0}} & 95.0/24.6 & \textcolor{insightaccent}{\textbf{98.4/9.7}} & 97.5/12.8 & 97.4/12.6 & 96.6/14.1 & \textcolor{insightaccent}{\textbf{96.7/14.0}}\\
\rowcolor{tableblue}BeiTv2-B & ViT & 92.9/30.1 & 94.1/23.7 & \textcolor{insightaccent}{\textbf{96.3/14.4}} & 95.5/20.7 & \textcolor{insightaccent}{\textbf{96.9/12.5}} & 94.0/28.8 & \textcolor{insightaccent}{\textbf{97.7/11.7}} & 96.8/12.2 & 96.7/11.8 & 95.6/14.3 & 95.5/14.5\\
\rowcolor{tablebluealt}DeiT3-B/16 & ViT & 87.3/49.9 & 81.9/73.1 & \textcolor{insightaccent}{\textbf{92.2/33.1}} & 85.2/50.5 & \textcolor{insightaccent}{\textbf{91.8/36.2}} & 88.9/49.3 & \textcolor{insightaccent}{\textbf{95.4/22.1}} & 86.2/47.8 & \textcolor{insightaccent}{\textbf{92.2/36.3}} & 92.2/37.2 & \textcolor{insightaccent}{\textbf{94.4/26.7}}\\
\rowcolor{tableblue}DeiT3-B/16-In21k & ViT & 90.1/38.7 & 91.3/44.0 & \textcolor{insightaccent}{\textbf{96.9/14.2}} & 91.2/40.7 & \textcolor{insightaccent}{\textbf{96.4/20.9}} & 91.1/41.0 & \textcolor{insightaccent}{\textbf{97.2/15.2}} & 92.8/36.6 & \textcolor{insightaccent}{\textbf{96.5/20.4}} & 95.6/21.7 & \textcolor{insightaccent}{\textbf{96.7/14.4}}\\
\rowcolor{tablebluealt}DeiT3-L/16 & ViT & 89.5/43.5 & 78.1/81.1 & \textcolor{insightaccent}{\textbf{91.5/34.6}} & 85.3/53.5 & \textcolor{insightaccent}{\textbf{92.2/41.6}} & 89.8/45.1 & \textcolor{insightaccent}{\textbf{95.4/22.9}} & 87.8/47.7 & \textcolor{insightaccent}{\textbf{92.9/36.2}} & 92.6/39.9 & \textcolor{insightaccent}{\textbf{94.8/26.0}}\\
\rowcolor{tableblue}DeiT3-S & ViT & 84.4/52.0 & 83.5/59.2 & \textcolor{insightaccent}{\textbf{92.2/32.2}} & 83.7/49.9 & \textcolor{insightaccent}{\textbf{91.4/37.8}} & 85.1/57.2 & \textcolor{insightaccent}{\textbf{93.8/26.1}} & 83.5/49.9 & \textcolor{insightaccent}{\textbf{91.3/39.2}} & 94.6/21.9 & \textcolor{insightaccent}{\textbf{95.1/18.5}}\\
\rowcolor{tablebluealt}EVA02-B & ViT & 86.3/60.5 & 83.9/46.5 & \textcolor{insightaccent}{\textbf{91.0/37.6}} & 89.1/40.5 & \textcolor{insightaccent}{\textbf{93.0/34.2}} & 88.3/48.2 & \textcolor{insightaccent}{\textbf{94.3/33.2}} & 91.8/36.0 & \textcolor{insightaccent}{\textbf{93.8/32.4}} & 93.2/32.7 & \textcolor{insightaccent}{\textbf{94.4/27.6}}\\
\rowcolor{tableblue}ViT-B-MAE & ViT & 44.8/96.6 & 83.3/43.0 & \textcolor{insightaccent}{\textbf{86.8/40.1}} & 88.9/27.1 & \textcolor{insightaccent}{\textbf{89.1/24.7}} & 90.8/29.3 & \textcolor{insightaccent}{\textbf{91.4/25.3}} & 88.9/25.7 & \textcolor{insightaccent}{\textbf{89.0/24.7}} & 88.9/24.8 & 88.9/24.5\\
\rowcolor{tablebluealt}ViT-B/16-augreg & ViT & 89.9/45.5 & 94.0/28.3 & \textcolor{insightaccent}{\textbf{96.8/12.1}} & 96.2/17.1 & \textcolor{insightaccent}{\textbf{96.9/12.2}} & 93.4/35.2 & \textcolor{insightaccent}{\textbf{98.1/9.5}} & 96.6/13.5 & 96.2/13.1 & 95.2/15.0 & 94.7/15.2\\
\rowcolor{tableblue}ViT-B/16-in1k & ViT & 81.4/65.7 & 84.3/58.1 & \textcolor{insightaccent}{\textbf{92.5/32.4}} & 82.8/48.6 & \textcolor{insightaccent}{\textbf{90.2/43.6}} & 86.0/56.2 & \textcolor{insightaccent}{\textbf{94.5/22.2}} & 89.6/39.9 & \textcolor{insightaccent}{\textbf{91.9/30.0}} & 92.1/27.1 & \textcolor{insightaccent}{\textbf{92.7/23.4}}\\
\rowcolor{tablebluealt}ViT-B/32 & ViT & 89.4/47.1 & 90.7/45.1 & \textcolor{insightaccent}{\textbf{95.1/15.7}} & 91.3/33.8 & \textcolor{insightaccent}{\textbf{94.3/22.3}} & 92.4/36.5 & \textcolor{insightaccent}{\textbf{97.2/13.7}} & 92.8/24.7 & \textcolor{insightaccent}{\textbf{93.3/19.5}} & 94.1/17.1 & 93.8/16.3\\
\rowcolor{tableblue}ViT-L/16 & ViT & 92.6/35.2 & 95.5/21.0 & \textcolor{insightaccent}{\textbf{97.2/11.2}} & 96.3/17.7 & \textcolor{insightaccent}{\textbf{97.0/12.4}} & 95.4/27.2 & \textcolor{insightaccent}{\textbf{98.7/7.1}} & 94.9/17.1 & 94.8/15.2 & 95.8/14.2 & 95.4/14.7\\
\rowcolor{tablebluealt}ViT-S/16 & ViT & 87.0/53.9 & 89.0/49.6 & \textcolor{insightaccent}{\textbf{95.7/13.6}} & 91.0/40.2 & \textcolor{insightaccent}{\textbf{95.7/16.4}} & 89.4/43.5 & \textcolor{insightaccent}{\textbf{97.1/14.7}} & 94.7/21.7 & \textcolor{insightaccent}{\textbf{95.5/14.6}} & 94.4/15.5 & 94.1/15.6\\
\rowcolor{hierheader}\multicolumn{13}{l}{\textbf{Hierarchical Transformers}}\\
\rowcolor{tableblue}Swin-B & Swin & 89.5/40.4 & 93.4/26.1 & \textcolor{insightaccent}{\textbf{95.0/16.7}} & 93.5/30.1 & \textcolor{insightaccent}{\textbf{95.6/16.9}} & 91.2/39.6 & \textcolor{insightaccent}{\textbf{95.9/22.7}} & 91.8/29.8 & \textcolor{insightaccent}{\textbf{92.5/19.4}} & 91.3/25.9 & \textcolor{insightaccent}{\textbf{92.7/19.1}}\\
\rowcolor{tablebluealt}SwinV2-B & Swin & 85.8/51.0 & 86.4/51.8 & \textcolor{insightaccent}{\textbf{94.0/25.3}} & 82.6/47.2 & \textcolor{insightaccent}{\textbf{91.3/38.5}} & 89.0/47.4 & \textcolor{insightaccent}{\textbf{95.8/20.4}} & 95.2/16.0 & 95.2/15.2 & 90.2/38.1 & \textcolor{insightaccent}{\textbf{94.3/26.9}}\\
\rowcolor{tableblue}SwinV2-B-In21k & Swin & 89.4/42.0 & 93.8/25.2 & \textcolor{insightaccent}{\textbf{95.7/16.0}} & 93.5/33.2 & \textcolor{insightaccent}{\textbf{95.6/21.6}} & 91.7/41.6 & \textcolor{insightaccent}{\textbf{96.5/20.8}} & 96.0/13.3 & 95.7/14.0 & 96.5/13.2 & 96.4/12.6\\
\rowcolor{tablebluealt}SwinV2-T & Swin & 85.0/52.6 & 85.0/53.7 & \textcolor{insightaccent}{\textbf{92.1/30.8}} & 80.6/48.7 & \textcolor{insightaccent}{\textbf{89.0/42.7}} & 85.9/53.2 & \textcolor{insightaccent}{\textbf{94.0/23.3}} & 91.5/17.6 & \textcolor{insightaccent}{\textbf{91.8/17.0}} & 87.4/36.7 & \textcolor{insightaccent}{\textbf{90.1/29.1}}\\
\rowcolor{cnnheader}\multicolumn{13}{l}{\textbf{Convolutional encoders}}\\
\rowcolor{tableblue}ConvNeXt-B & CNN & 84.5/52.7 & 85.3/51.5 & \textcolor{insightaccent}{\textbf{88.8/42.1}} & 80.5/47.0 & \textcolor{insightaccent}{\textbf{85.5/46.0}} & 88.7/48.4 & \textcolor{insightaccent}{\textbf{91.6/37.5}} & 78.7/60.3 & \textcolor{insightaccent}{\textbf{84.1/55.4}} & 79.4/57.2 & \textcolor{insightaccent}{\textbf{84.8/53.0}}\\
\rowcolor{tablebluealt}ConvNeXt-B-In21k & CNN & 88.7/41.4 & 94.5/24.6 & \textcolor{insightaccent}{\textbf{95.4/18.3}} & 93.6/31.9 & \textcolor{insightaccent}{\textbf{95.3/24.0}} & 91.6/40.6 & \textcolor{insightaccent}{\textbf{94.5/28.0}} & 93.8/29.2 & \textcolor{insightaccent}{\textbf{94.9/23.3}} & 80.8/49.0 & \textcolor{insightaccent}{\textbf{88.9/42.7}}\\
\rowcolor{tableblue}ConvNeXt-L & CNN & 85.8/52.0 & 86.6/51.3 & \textcolor{insightaccent}{\textbf{91.3/35.5}} & 84.6/43.2 & \textcolor{insightaccent}{\textbf{89.4/36.2}} & 90.6/41.0 & \textcolor{insightaccent}{\textbf{93.3/30.5}} & 82.5/56.5 & \textcolor{insightaccent}{\textbf{87.7/45.2}} & 79.1/53.2 & \textcolor{insightaccent}{\textbf{86.5/48.6}}\\
\rowcolor{tablebluealt}ConvNeXt-T & CNN & 81.7/54.8 & 80.9/53.3 & \textcolor{insightaccent}{\textbf{87.7/40.8}} & 76.9/48.8 & \textcolor{insightaccent}{\textbf{84.3/42.7}} & 84.6/54.9 & \textcolor{insightaccent}{\textbf{91.1/33.8}} & 77.0/58.6 & \textcolor{insightaccent}{\textbf{84.4/46.5}} & 83.9/38.6 & \textcolor{insightaccent}{\textbf{89.0/32.0}}\\
\rowcolor{tableblue}ConvNeXtV2-B & CNN & 86.4/47.1 & 86.1/50.1 & \textcolor{insightaccent}{\textbf{89.0/46.0}} & 82.5/46.1 & \textcolor{insightaccent}{\textbf{86.3/44.2}} & 88.4/49.8 & \textcolor{insightaccent}{\textbf{91.6/39.0}} & 79.3/57.8 & \textcolor{insightaccent}{\textbf{84.0/53.9}} & 76.2/56.0 & \textcolor{insightaccent}{\textbf{82.2/51.0}}\\
\rowcolor{tablebluealt}ConvNeXtV2-B-In21k & CNN & 90.1/40.0 & 93.1/31.7 & \textcolor{insightaccent}{\textbf{95.1/22.6}} & 93.1/34.1 & \textcolor{insightaccent}{\textbf{95.1/26.1}} & 90.3/48.1 & \textcolor{insightaccent}{\textbf{94.2/30.9}} & 93.2/35.2 & \textcolor{insightaccent}{\textbf{95.1/25.7}} & 68.7/60.7 & \textcolor{insightaccent}{\textbf{81.8/53.0}}\\
\rowcolor{tableblue}ConvNeXtV2-T & CNN & 84.0/54.4 & 82.9/53.9 & \textcolor{insightaccent}{\textbf{86.9/46.9}} & 78.0/48.3 & \textcolor{insightaccent}{\textbf{83.2/46.2}} & 86.2/55.0 & \textcolor{insightaccent}{\textbf{90.5/40.5}} & 79.9/53.6 & \textcolor{insightaccent}{\textbf{84.3/49.0}} & 71.8/55.9 & \textcolor{insightaccent}{\textbf{79.2/51.9}}\\
\rowcolor{tablebluealt}DenseNet-121 & CNN & 71.7/82.2 & 83.3/48.4 & \textcolor{insightaccent}{\textbf{88.8/31.9}} & 79.5/50.0 & \textcolor{insightaccent}{\textbf{85.8/40.0}} & 87.4/44.5 & \textcolor{insightaccent}{\textbf{92.7/25.9}} & 81.5/47.2 & \textcolor{insightaccent}{\textbf{86.4/36.7}} & 90.7/20.8 & 90.7/20.5\\
\rowcolor{tableblue}EffNetV2-L & CNN & 86.3/49.7 & 94.6/22.1 & \textcolor{insightaccent}{\textbf{96.2/14.0}} & 93.0/26.3 & \textcolor{insightaccent}{\textbf{95.4/20.3}} & 91.7/37.8 & \textcolor{insightaccent}{\textbf{97.2/15.2}} & 94.7/24.5 & \textcolor{insightaccent}{\textbf{95.5/16.5}} & 89.7/24.3 & \textcolor{insightaccent}{\textbf{90.9/21.4}}\\
\rowcolor{tablebluealt}EffNetV2-M & CNN & 82.7/52.1 & 90.7/33.5 & \textcolor{insightaccent}{\textbf{96.1/14.4}} & 90.7/35.0 & \textcolor{insightaccent}{\textbf{95.5/20.3}} & 87.1/47.9 & \textcolor{insightaccent}{\textbf{95.8/21.8}} & 94.2/29.7 & \textcolor{insightaccent}{\textbf{95.9/16.3}} & 90.2/22.0 & \textcolor{insightaccent}{\textbf{91.4/19.7}}\\
\rowcolor{tableblue}EffNetV2-S & CNN & 84.1/53.4 & 89.7/39.7 & \textcolor{insightaccent}{\textbf{95.5/18.9}} & 88.0/44.9 & \textcolor{insightaccent}{\textbf{93.4/26.9}} & 89.2/43.9 & \textcolor{insightaccent}{\textbf{95.9/20.0}} & 90.5/40.7 & \textcolor{insightaccent}{\textbf{94.5/26.1}} & 89.3/25.9 & \textcolor{insightaccent}{\textbf{90.9/22.4}}\\
\rowcolor{tablebluealt}RegNetY-040 & CNN & 72.9/78.7 & 88.6/41.1 & \textcolor{insightaccent}{\textbf{92.0/22.9}} & 85.0/52.9 & \textcolor{insightaccent}{\textbf{89.7/32.9}} & 84.4/52.2 & \textcolor{insightaccent}{\textbf{92.8/25.1}} & 85.7/49.5 & \textcolor{insightaccent}{\textbf{89.6/32.0}} & 89.8/19.5 & \textcolor{insightaccent}{\textbf{90.2/19.2}}\\
\rowcolor{tableblue}ResNet-101 & CNN & 85.2/52.6 & 90.0/38.9 & \textcolor{insightaccent}{\textbf{92.4/23.8}} & 83.5/50.7 & \textcolor{insightaccent}{\textbf{88.5/43.0}} & 89.5/42.2 & \textcolor{insightaccent}{\textbf{93.4/26.0}} & 84.1/50.1 & \textcolor{insightaccent}{\textbf{88.8/42.4}} & 89.2/25.0 & \textcolor{insightaccent}{\textbf{89.9/23.6}}\\
\rowcolor{tablebluealt}ResNet-152 & CNN & 86.8/52.1 & 89.5/48.9 & \textcolor{insightaccent}{\textbf{93.3/26.5}} & 80.3/49.9 & \textcolor{insightaccent}{\textbf{87.6/43.7}} & 89.4/45.3 & \textcolor{insightaccent}{\textbf{94.1/24.2}} & 80.7/49.6 & \textcolor{insightaccent}{\textbf{87.8/43.4}} & 89.3/25.4 & \textcolor{insightaccent}{\textbf{90.4/22.7}}\\
\rowcolor{tableblue}ResNet-50 & CNN & 76.9/73.1 & 84.9/52.0 & \textcolor{insightaccent}{\textbf{89.3/38.0}} & 74.3/51.8 & \textcolor{insightaccent}{\textbf{83.0/48.6}} & 87.1/49.3 & \textcolor{insightaccent}{\textbf{92.0/31.1}} & 76.8/50.8 & \textcolor{insightaccent}{\textbf{83.9/47.7}} & 89.0/26.5 & \textcolor{insightaccent}{\textbf{89.4/25.1}}\\
\rowcolor{tablebluealt}ResNetV2-50 & CNN & 81.5/63.5 & 87.2/50.1 & \textcolor{insightaccent}{\textbf{91.5/31.4}} & 76.9/52.5 & \textcolor{insightaccent}{\textbf{85.1/48.0}} & 87.3/47.6 & \textcolor{insightaccent}{\textbf{92.0/28.4}} & 87.9/40.7 & \textcolor{insightaccent}{\textbf{90.0/36.1}} & 89.2/23.7 & \textcolor{insightaccent}{\textbf{89.7/21.6}}\\
\rowcolor{tableblue}WideResNet50-2 & CNN & 81.6/58.1 & 85.1/52.8 & \textcolor{insightaccent}{\textbf{89.8/37.6}} & 77.2/52.5 & \textcolor{insightaccent}{\textbf{84.9/48.5}} & 87.5/47.1 & \textcolor{insightaccent}{\textbf{92.1/28.0}} & 80.1/51.1 & \textcolor{insightaccent}{\textbf{86.4/47.0}} & 89.2/26.3 & \textcolor{insightaccent}{\textbf{89.6/24.2}}\\
\bottomrule
\end{tabular}%
}
\end{table*}

\newpage
\begin{table*}[p]
\centering\scriptsize\setlength{\tabcolsep}{2.4pt}\renewcommand{\arraystretch}{1.08}
\caption{\textbf{Complete OpenOOD results: ImageNet-200.} 42 checkpoints, 50 fit images per class, and five OOD datasets.}
\label{tab:ood-full-imagenet200}
\vspace{0pt}
\resizebox{\textwidth}{!}{%
\begin{tabular}{ll c cc cc cc cc cc}
\toprule
\rowcolor{headerblue}
\textbf{Model} & \textbf{Fam} & \textbf{MSP} & \textbf{kNN} & \textbf{+path} & \textbf{Maha++} & \textbf{+path} & \textbf{RelMaha} & \textbf{+path} & \textbf{X-Maha} & \textbf{+path} & \textbf{Traj-Maha} & \textbf{+path}\\
\midrule
\rowcolor{vlmheader}\multicolumn{13}{l}{\textbf{Self-supervised foundation encoders}}\\
\rowcolor{tableblue}DINOv2 ViT-B/14 & SSL & 95.6/15.4 & 95.7/16.0 & \textcolor{insightaccent}{\textbf{96.0/12.9}} & 96.0/14.0 & \textcolor{insightaccent}{\textbf{96.3/13.0}} & 94.8/20.6 & \textcolor{insightaccent}{\textbf{95.5/17.8}} & 96.3/13.2 & 96.3/12.4 & 95.4/22.2 & 95.2/22.8\\
\rowcolor{tablebluealt}DINOv3 ViT-B/16 & SSL & 94.5/20.5 & 82.9/55.7 & \textcolor{insightaccent}{\textbf{88.0/41.8}} & 92.7/28.6 & \textcolor{insightaccent}{\textbf{94.2/24.2}} & 93.8/24.4 & \textcolor{insightaccent}{\textbf{95.0/20.9}} & 93.5/26.1 & \textcolor{insightaccent}{\textbf{94.6/22.9}} & 93.2/40.2 & 93.0/40.6\\
\rowcolor{vlmheader}\multicolumn{13}{l}{\textbf{Vision--language encoders}}\\
\rowcolor{tableblue}CLIP ViT-B/16 & VLM & 88.0/47.0 & 61.9/96.3 & \textcolor{insightaccent}{\textbf{71.6/89.4}} & 81.0/89.2 & \textcolor{insightaccent}{\textbf{85.5/77.2}} & 90.7/44.5 & \textcolor{insightaccent}{\textbf{93.0/34.0}} & 82.6/85.4 & \textcolor{insightaccent}{\textbf{86.2/73.8}} & 85.5/66.8 & \textcolor{insightaccent}{\textbf{87.0/60.5}}\\
\rowcolor{tablebluealt}CLIP ViT-L/14 & VLM & 92.8/29.0 & 83.2/76.5 & \textcolor{insightaccent}{\textbf{88.0/58.6}} & 86.1/72.7 & \textcolor{insightaccent}{\textbf{89.3/60.5}} & 93.2/31.5 & \textcolor{insightaccent}{\textbf{95.0/23.0}} & 90.4/51.8 & \textcolor{insightaccent}{\textbf{91.1/48.2}} & 87.9/57.6 & \textcolor{insightaccent}{\textbf{88.9/54.8}}\\
\rowcolor{tableblue}CLIP-timm ViT-B/16 & VLM & 88.9/44.5 & 78.0/87.7 & \textcolor{insightaccent}{\textbf{83.7/70.2}} & 78.9/88.2 & \textcolor{insightaccent}{\textbf{83.3/78.5}} & 89.8/50.1 & \textcolor{insightaccent}{\textbf{91.7/39.5}} & 82.7/77.0 & \textcolor{insightaccent}{\textbf{85.1/68.4}} & 82.3/69.7 & \textcolor{insightaccent}{\textbf{84.2/65.6}}\\
\rowcolor{tablebluealt}PE-Core ViT-B/16 & VLM & 92.1/31.3 & 70.0/88.9 & \textcolor{insightaccent}{\textbf{76.2/73.8}} & 88.2/65.1 & \textcolor{insightaccent}{\textbf{90.9/51.7}} & 93.0/27.4 & \textcolor{insightaccent}{\textbf{94.6/23.6}} & 88.9/56.9 & \textcolor{insightaccent}{\textbf{90.5/50.5}} & 85.0/63.0 & \textcolor{insightaccent}{\textbf{87.6/58.3}}\\
\rowcolor{tableblue}SigLIP2 ViT-B/16 & VLM & 94.8/18.1 & 74.0/88.3 & \textcolor{insightaccent}{\textbf{80.4/73.4}} & 88.7/68.7 & \textcolor{insightaccent}{\textbf{91.7/51.8}} & 95.5/15.9 & \textcolor{insightaccent}{\textbf{96.4/14.1}} & 89.1/56.0 & \textcolor{insightaccent}{\textbf{91.5/46.2}} & 89.8/56.4 & \textcolor{insightaccent}{\textbf{91.8/47.2}}\\
\rowcolor{vitheader}\multicolumn{13}{l}{\textbf{Supervised plain ViTs}}\\
\rowcolor{tableblue}BeiT-B/16 & ViT & 90.3/35.7 & 95.2/18.4 & \textcolor{insightaccent}{\textbf{96.2/14.3}} & 96.0/17.0 & \textcolor{insightaccent}{\textbf{96.5/14.2}} & 96.5/12.3 & \textcolor{insightaccent}{\textbf{96.9/11.6}} & 96.1/16.3 & \textcolor{insightaccent}{\textbf{96.2/15.4}} & 96.0/17.0 & 96.0/16.6\\
\rowcolor{tablebluealt}BeiT-L & ViT & 91.5/32.0 & 95.3/24.2 & \textcolor{insightaccent}{\textbf{96.3/16.0}} & 96.4/16.1 & \textcolor{insightaccent}{\textbf{96.9/13.7}} & 97.2/9.8 & \textcolor{insightaccent}{\textbf{97.4/9.4}} & 96.9/13.1 & 96.9/12.9 & 96.7/14.7 & 96.7/14.1\\
\rowcolor{tableblue}BeiTv2-B & ViT & 90.4/36.3 & 94.4/30.3 & \textcolor{insightaccent}{\textbf{96.1/18.2}} & 96.8/12.4 & \textcolor{insightaccent}{\textbf{97.2/11.4}} & 96.8/11.7 & \textcolor{insightaccent}{\textbf{97.2/10.6}} & 97.2/10.9 & 97.1/11.3 & 96.4/14.6 & 96.4/14.5\\
\rowcolor{tablebluealt}DeiT3-B/16 & ViT & 85.1/55.3 & 90.9/50.0 & \textcolor{insightaccent}{\textbf{92.2/43.3}} & 93.6/31.5 & 93.9/31.6 & 94.7/20.3 & 94.9/20.3 & 92.7/40.6 & \textcolor{insightaccent}{\textbf{93.1/38.6}} & 93.6/32.6 & 93.7/32.6\\
\rowcolor{tableblue}DeiT3-B/16-In21k & ViT & 83.3/55.3 & 95.8/14.7 & \textcolor{insightaccent}{\textbf{96.4/13.3}} & 96.3/12.6 & \textcolor{insightaccent}{\textbf{96.7/12.1}} & 96.2/12.7 & \textcolor{insightaccent}{\textbf{96.7/12.0}} & 96.5/12.6 & \textcolor{insightaccent}{\textbf{96.8/12.3}} & 96.5/13.1 & \textcolor{insightaccent}{\textbf{96.7/12.6}}\\
\rowcolor{tablebluealt}DeiT3-L/16 & ViT & 84.5/55.6 & 91.8/42.6 & \textcolor{insightaccent}{\textbf{92.8/37.3}} & 94.3/23.9 & 94.5/24.4 & 95.3/16.2 & 95.5/16.5 & 93.8/29.3 & 94.0/29.3 & 94.5/23.1 & 94.5/23.7\\
\rowcolor{tableblue}DeiT3-S & ViT & 86.4/52.5 & 91.1/54.6 & \textcolor{insightaccent}{\textbf{92.7/40.1}} & 93.3/35.6 & \textcolor{insightaccent}{\textbf{93.9/32.3}} & 94.0/27.1 & \textcolor{insightaccent}{\textbf{94.6/25.4}} & 92.7/43.4 & \textcolor{insightaccent}{\textbf{93.4/36.8}} & 93.6/34.7 & \textcolor{insightaccent}{\textbf{93.8/33.6}}\\
\rowcolor{tablebluealt}EVA02-B & ViT & 93.9/25.7 & 87.3/60.2 & \textcolor{insightaccent}{\textbf{90.4/45.5}} & 92.9/35.2 & \textcolor{insightaccent}{\textbf{93.3/33.4}} & 94.7/20.5 & \textcolor{insightaccent}{\textbf{95.0/19.9}} & 91.9/40.1 & \textcolor{insightaccent}{\textbf{92.1/39.9}} & 91.0/43.8 & \textcolor{insightaccent}{\textbf{91.5/42.4}}\\
\rowcolor{tableblue}ViT-B-MAE & ViT & 41.8/95.5 & 64.8/79.8 & \textcolor{insightaccent}{\textbf{67.6/77.7}} & 66.0/77.7 & \textcolor{insightaccent}{\textbf{68.9/75.5}} & 82.5/69.5 & \textcolor{insightaccent}{\textbf{83.7/65.2}} & 68.3/75.9 & \textcolor{insightaccent}{\textbf{70.7/74.0}} & 63.7/77.1 & \textcolor{insightaccent}{\textbf{67.4/75.7}}\\
\rowcolor{tablebluealt}ViT-B/16-augreg & ViT & 89.5/39.9 & 94.0/29.9 & \textcolor{insightaccent}{\textbf{95.6/21.2}} & 96.7/11.6 & \textcolor{insightaccent}{\textbf{97.1/10.8}} & 96.2/13.5 & \textcolor{insightaccent}{\textbf{96.9/11.4}} & 97.5/10.0 & 97.3/11.5 & 96.6/15.6 & 96.1/18.9\\
\rowcolor{tableblue}ViT-B/16-in1k & ViT & 84.9/58.8 & 92.0/38.6 & \textcolor{insightaccent}{\textbf{93.1/33.5}} & 93.6/28.6 & \textcolor{insightaccent}{\textbf{94.0/27.1}} & 93.9/29.5 & \textcolor{insightaccent}{\textbf{94.6/25.3}} & 93.1/34.5 & \textcolor{insightaccent}{\textbf{93.3/33.8}} & 93.6/32.0 & \textcolor{insightaccent}{\textbf{93.7/31.3}}\\
\rowcolor{tablebluealt}ViT-B/32 & ViT & 87.2/49.8 & 91.8/40.9 & \textcolor{insightaccent}{\textbf{93.5/31.9}} & 95.8/17.7 & \textcolor{insightaccent}{\textbf{96.1/17.2}} & 94.5/23.6 & \textcolor{insightaccent}{\textbf{95.5/18.8}} & 95.7/20.2 & 95.1/24.5 & 94.2/29.1 & 93.8/31.8\\
\rowcolor{tableblue}ViT-L/16 & ViT & 91.1/34.6 & 95.0/22.4 & \textcolor{insightaccent}{\textbf{96.1/17.7}} & 97.2/9.8 & \textcolor{insightaccent}{\textbf{97.5/9.1}} & 97.0/10.8 & \textcolor{insightaccent}{\textbf{97.6/9.1}} & 97.0/13.9 & 96.7/16.8 & 96.8/14.3 & 96.5/16.4\\
\rowcolor{tablebluealt}ViT-S/16 & ViT & 86.9/50.4 & 93.6/31.9 & \textcolor{insightaccent}{\textbf{95.1/23.1}} & 96.1/16.1 & \textcolor{insightaccent}{\textbf{96.5/15.1}} & 95.1/20.6 & \textcolor{insightaccent}{\textbf{96.0/17.1}} & 96.6/14.7 & 96.4/15.7 & 94.4/28.9 & 94.1/31.2\\
\rowcolor{hierheader}\multicolumn{13}{l}{\textbf{Hierarchical Transformers}}\\
\rowcolor{tableblue}Swin-B & Swin & 89.1/39.3 & 95.4/18.1 & 94.1/25.3 & 96.6/12.1 & 95.8/14.8 & 95.6/14.3 & 95.4/14.9 & 91.3/44.2 & \textcolor{insightaccent}{\textbf{91.4/42.8}} & 91.4/30.4 & 91.7/30.4\\
\rowcolor{tablebluealt}SwinV2-B & Swin & 84.5/57.7 & 95.1/18.5 & 95.1/18.5 & 95.3/16.3 & \textcolor{insightaccent}{\textbf{95.5/16.1}} & 95.2/16.9 & \textcolor{insightaccent}{\textbf{95.4/15.7}} & 95.2/20.6 & \textcolor{insightaccent}{\textbf{95.4/18.8}} & 95.5/17.2 & 95.5/17.4\\
\rowcolor{tableblue}SwinV2-B-In21k & Swin & 89.3/39.0 & 95.7/15.1 & 94.8/19.2 & 96.7/10.9 & 96.1/12.4 & 95.9/12.5 & 95.7/12.8 & 94.7/22.6 & 94.7/22.1 & 96.2/15.3 & 95.8/16.8\\
\rowcolor{tablebluealt}SwinV2-T & Swin & 85.5/53.6 & 93.4/30.2 & 92.5/37.5 & 94.8/21.2 & 94.2/25.4 & 94.0/27.1 & 93.9/28.0 & 79.6/69.0 & \textcolor{insightaccent}{\textbf{83.5/66.7}} & 86.9/62.3 & 87.5/62.6\\
\rowcolor{cnnheader}\multicolumn{13}{l}{\textbf{Convolutional encoders}}\\
\rowcolor{tableblue}ConvNeXt-B & CNN & 83.6/59.3 & 94.8/20.5 & 94.5/22.6 & 94.7/22.0 & \textcolor{insightaccent}{\textbf{94.8/21.1}} & 94.3/22.6 & \textcolor{insightaccent}{\textbf{94.7/19.0}} & 94.1/27.6 & \textcolor{insightaccent}{\textbf{94.3/25.7}} & 92.6/36.5 & \textcolor{insightaccent}{\textbf{93.7/32.1}}\\
\rowcolor{tablebluealt}ConvNeXt-B-In21k & CNN & 90.3/37.3 & 96.4/12.6 & \textcolor{insightaccent}{\textbf{96.5/12.1}} & 97.1/9.7 & 97.2/9.7 & 96.1/11.6 & \textcolor{insightaccent}{\textbf{96.6/10.8}} & 96.3/13.9 & \textcolor{insightaccent}{\textbf{96.5/13.2}} & 90.9/36.0 & \textcolor{insightaccent}{\textbf{93.7/29.5}}\\
\rowcolor{tableblue}ConvNeXt-L & CNN & 83.1/60.9 & 94.9/18.9 & 94.8/19.8 & 94.8/19.7 & \textcolor{insightaccent}{\textbf{95.1/18.3}} & 94.6/19.1 & \textcolor{insightaccent}{\textbf{95.0/17.5}} & 94.3/24.2 & \textcolor{insightaccent}{\textbf{94.5/22.9}} & 92.6/38.5 & \textcolor{insightaccent}{\textbf{93.9/32.2}}\\
\rowcolor{tablebluealt}ConvNeXt-T & CNN & 83.8/57.1 & 93.2/30.8 & 93.2/31.4 & 94.5/23.2 & 94.4/25.5 & 93.8/27.2 & \textcolor{insightaccent}{\textbf{94.2/24.7}} & 92.8/39.2 & \textcolor{insightaccent}{\textbf{93.1/35.9}} & 79.3/62.2 & \textcolor{insightaccent}{\textbf{86.9/53.4}}\\
\rowcolor{tableblue}ConvNeXtV2-B & CNN & 85.4/53.7 & 95.3/17.0 & 95.0/18.6 & 95.1/16.7 & \textcolor{insightaccent}{\textbf{95.2/16.6}} & 94.8/18.3 & \textcolor{insightaccent}{\textbf{95.1/16.6}} & 94.3/26.1 & \textcolor{insightaccent}{\textbf{94.5/25.6}} & 87.8/52.3 & \textcolor{insightaccent}{\textbf{92.1/41.5}}\\
\rowcolor{tablebluealt}ConvNeXtV2-B-In21k & CNN & 90.8/33.9 & 97.3/10.5 & 97.1/10.6 & 97.7/8.5 & 97.7/8.6 & 96.9/10.0 & \textcolor{insightaccent}{\textbf{97.2/9.6}} & 97.2/10.7 & \textcolor{insightaccent}{\textbf{97.3/10.6}} & 83.2/61.8 & \textcolor{insightaccent}{\textbf{92.4/38.8}}\\
\rowcolor{tableblue}ConvNeXtV2-T & CNN & 85.6/54.4 & 93.6/29.2 & 93.7/30.0 & 94.4/23.0 & \textcolor{insightaccent}{\textbf{94.5/22.8}} & 94.1/26.9 & \textcolor{insightaccent}{\textbf{94.5/22.2}} & 93.2/35.2 & \textcolor{insightaccent}{\textbf{93.4/33.6}} & 82.1/60.4 & \textcolor{insightaccent}{\textbf{88.5/50.3}}\\
\rowcolor{tablebluealt}DenseNet-121 & CNN & 83.7/64.7 & 88.0/51.0 & 88.3/51.0 & 87.2/55.4 & \textcolor{insightaccent}{\textbf{88.0/52.4}} & 89.2/53.9 & \textcolor{insightaccent}{\textbf{90.8/45.8}} & 87.3/55.0 & \textcolor{insightaccent}{\textbf{87.9/53.0}} & 65.2/79.4 & \textcolor{insightaccent}{\textbf{74.0/75.3}}\\
\rowcolor{tableblue}EffNetV2-L & CNN & 87.6/41.0 & 96.5/13.0 & 96.4/13.8 & 95.7/17.9 & \textcolor{insightaccent}{\textbf{96.0/16.7}} & 95.3/15.1 & \textcolor{insightaccent}{\textbf{96.0/13.3}} & 96.2/14.4 & 96.1/14.9 & 76.0/76.1 & \textcolor{insightaccent}{\textbf{83.8/70.4}}\\
\rowcolor{tablebluealt}EffNetV2-M & CNN & 87.4/41.4 & 95.8/17.5 & 95.9/17.6 & 94.9/24.2 & \textcolor{insightaccent}{\textbf{95.3/22.3}} & 95.0/16.8 & \textcolor{insightaccent}{\textbf{95.7/15.3}} & 95.9/15.9 & 95.8/16.9 & 74.8/77.4 & \textcolor{insightaccent}{\textbf{83.4/72.3}}\\
\rowcolor{tableblue}EffNetV2-S & CNN & 85.2/49.4 & 95.1/22.1 & \textcolor{insightaccent}{\textbf{95.2/21.1}} & 94.5/27.2 & \textcolor{insightaccent}{\textbf{94.9/24.4}} & 94.2/23.8 & \textcolor{insightaccent}{\textbf{95.0/20.3}} & 95.0/20.8 & 94.9/21.7 & 74.5/75.9 & \textcolor{insightaccent}{\textbf{82.8/70.2}}\\
\rowcolor{tablebluealt}RegNetY-040 & CNN & 85.5/56.0 & 91.7/34.0 & 91.6/36.2 & 91.9/35.4 & 91.9/36.2 & 93.1/35.4 & \textcolor{insightaccent}{\textbf{93.8/30.7}} & 91.7/36.5 & 91.6/37.1 & 74.8/74.3 & \textcolor{insightaccent}{\textbf{80.8/69.4}}\\
\rowcolor{tableblue}ResNet-101 & CNN & 82.7/57.5 & 94.9/21.5 & 95.1/21.7 & 95.4/20.4 & \textcolor{insightaccent}{\textbf{95.6/19.2}} & 94.6/24.4 & \textcolor{insightaccent}{\textbf{95.3/21.0}} & 95.5/20.4 & \textcolor{insightaccent}{\textbf{95.7/19.1}} & 68.4/79.0 & \textcolor{insightaccent}{\textbf{84.2/68.7}}\\
\rowcolor{tablebluealt}ResNet-152 & CNN & 83.6/54.9 & 95.5/18.3 & \textcolor{insightaccent}{\textbf{95.8/17.5}} & 95.9/17.3 & \textcolor{insightaccent}{\textbf{96.2/16.2}} & 95.0/21.6 & \textcolor{insightaccent}{\textbf{95.7/18.0}} & 95.8/17.6 & \textcolor{insightaccent}{\textbf{96.1/16.5}} & 72.2/76.9 & \textcolor{insightaccent}{\textbf{87.2/63.0}}\\
\rowcolor{tableblue}ResNet-50 & CNN & 85.5/59.4 & 92.4/35.4 & \textcolor{insightaccent}{\textbf{93.1/32.5}} & 90.8/50.7 & \textcolor{insightaccent}{\textbf{92.5/44.0}} & 91.4/58.4 & \textcolor{insightaccent}{\textbf{93.3/38.1}} & 91.1/49.7 & \textcolor{insightaccent}{\textbf{92.6/43.0}} & 70.1/77.6 & \textcolor{insightaccent}{\textbf{82.0/70.3}}\\
\rowcolor{tablebluealt}ResNetV2-50 & CNN & 83.5/57.7 & 93.3/29.6 & 93.0/33.0 & 94.2/28.2 & 94.1/29.4 & 93.6/30.8 & \textcolor{insightaccent}{\textbf{94.5/25.1}} & 90.7/45.0 & 90.5/45.1 & 68.2/78.1 & \textcolor{insightaccent}{\textbf{78.5/72.2}}\\
\rowcolor{tableblue}WideResNet50-2 & CNN & 84.0/57.5 & 94.6/24.2 & 94.7/24.7 & 94.8/24.7 & \textcolor{insightaccent}{\textbf{95.0/23.9}} & 94.3/25.2 & \textcolor{insightaccent}{\textbf{95.2/21.2}} & 95.0/23.7 & 95.1/24.2 & 73.6/76.8 & \textcolor{insightaccent}{\textbf{83.8/69.6}}\\
\bottomrule
\end{tabular}%
}
\end{table*}

\newpage
\begin{table*}[p]
\centering\scriptsize\setlength{\tabcolsep}{2.4pt}\renewcommand{\arraystretch}{1.08}
\caption{\textbf{Complete OpenOOD results: ImageNet-1K.} 42 checkpoints, 50 fit images per class, and five OOD datasets.}
\label{tab:ood-full-imagenet1k}
\vspace{0pt}
\resizebox{\textwidth}{!}{%
\begin{tabular}{ll c cc cc cc cc cc}
\toprule
\rowcolor{headerblue}
\textbf{Model} & \textbf{Fam} & \textbf{MSP} & \textbf{kNN} & \textbf{+path} & \textbf{Maha++} & \textbf{+path} & \textbf{RelMaha} & \textbf{+path} & \textbf{X-Maha} & \textbf{+path} & \textbf{Traj-Maha} & \textbf{+path}\\
\midrule
\rowcolor{vlmheader}\multicolumn{13}{l}{\textbf{Self-supervised foundation encoders}}\\
\rowcolor{tableblue}DINOv2 ViT-B/14 & SSL & 88.2/41.8 & 89.5/37.1 & \textcolor{insightaccent}{\textbf{91.6/28.7}} & 91.3/32.0 & \textcolor{insightaccent}{\textbf{92.6/27.4}} & 89.3/35.4 & \textcolor{insightaccent}{\textbf{91.3/28.6}} & 92.2/27.7 & \textcolor{insightaccent}{\textbf{92.8/26.4}} & 91.8/34.2 & 91.6/35.0\\
\rowcolor{tablebluealt}DINOv3 ViT-B/16 & SSL & 83.5/54.7 & 82.9/55.7 & \textcolor{insightaccent}{\textbf{88.0/41.8}} & 85.6/47.4 & \textcolor{insightaccent}{\textbf{88.9/40.9}} & 86.8/44.5 & \textcolor{insightaccent}{\textbf{90.1/34.8}} & 87.5/43.7 & \textcolor{insightaccent}{\textbf{89.8/39.0}} & 89.5/52.0 & \textcolor{insightaccent}{\textbf{89.8/51.1}}\\
\rowcolor{vlmheader}\multicolumn{13}{l}{\textbf{Vision--language encoders}}\\
\rowcolor{tableblue}CLIP ViT-B/16 & VLM & 67.9/81.1 & 61.9/96.3 & \textcolor{insightaccent}{\textbf{71.6/89.4}} & 69.4/95.7 & \textcolor{insightaccent}{\textbf{76.6/91.6}} & 80.5/72.6 & \textcolor{insightaccent}{\textbf{85.1/63.2}} & 71.6/94.4 & \textcolor{insightaccent}{\textbf{77.4/89.4}} & 79.4/79.4 & \textcolor{insightaccent}{\textbf{81.2/75.0}}\\
\rowcolor{tablebluealt}CLIP ViT-L/14 & VLM & 73.2/78.4 & 69.5/91.9 & \textcolor{insightaccent}{\textbf{77.0/80.1}} & 74.6/88.5 & \textcolor{insightaccent}{\textbf{80.0/81.5}} & 84.5/62.1 & \textcolor{insightaccent}{\textbf{88.3/50.4}} & 82.8/70.8 & \textcolor{insightaccent}{\textbf{84.4/67.0}} & 82.0/70.8 & \textcolor{insightaccent}{\textbf{83.3/68.5}}\\
\rowcolor{tableblue}CLIP-timm ViT-B/16 & VLM & 68.1/82.6 & 62.4/95.7 & \textcolor{insightaccent}{\textbf{70.6/88.2}} & 66.2/95.2 & \textcolor{insightaccent}{\textbf{72.6/91.0}} & 79.3/73.8 & \textcolor{insightaccent}{\textbf{82.8/64.6}} & 72.2/88.9 & \textcolor{insightaccent}{\textbf{75.6/83.4}} & 74.9/79.8 & \textcolor{insightaccent}{\textbf{76.9/77.3}}\\
\rowcolor{tablebluealt}PE-Core ViT-B/16 & VLM & 71.0/81.6 & 70.0/88.9 & \textcolor{insightaccent}{\textbf{76.2/73.8}} & 78.7/86.6 & \textcolor{insightaccent}{\textbf{82.1/77.6}} & 83.4/59.3 & \textcolor{insightaccent}{\textbf{86.8/49.6}} & 80.9/72.7 & \textcolor{insightaccent}{\textbf{82.1/69.9}} & 77.8/75.2 & \textcolor{insightaccent}{\textbf{79.4/72.9}}\\
\rowcolor{tableblue}SigLIP2 ViT-B/16 & VLM & 76.8/71.8 & 74.0/88.3 & \textcolor{insightaccent}{\textbf{80.4/73.4}} & 79.5/87.3 & \textcolor{insightaccent}{\textbf{83.1/77.3}} & 87.7/46.7 & \textcolor{insightaccent}{\textbf{89.8/39.3}} & 81.2/73.7 & \textcolor{insightaccent}{\textbf{83.5/66.7}} & 82.7/71.0 & \textcolor{insightaccent}{\textbf{84.2/66.6}}\\
\rowcolor{vitheader}\multicolumn{13}{l}{\textbf{Supervised plain ViTs}}\\
\rowcolor{tableblue}BeiT-B/16 & ViT & 88.4/42.3 & 88.6/46.3 & \textcolor{insightaccent}{\textbf{90.8/36.3}} & 91.0/37.1 & \textcolor{insightaccent}{\textbf{92.1/32.5}} & 91.3/35.3 & \textcolor{insightaccent}{\textbf{92.4/31.0}} & 91.8/29.4 & 91.8/29.7 & 91.2/34.1 & 91.0/34.7\\
\rowcolor{tablebluealt}BeiT-L & ViT & 90.1/36.1 & 90.3/46.1 & \textcolor{insightaccent}{\textbf{92.0/37.4}} & 92.3/35.6 & \textcolor{insightaccent}{\textbf{93.1/32.0}} & 92.9/29.9 & \textcolor{insightaccent}{\textbf{93.5/27.3}} & 93.4/25.6 & 93.2/26.3 & 92.9/30.7 & 92.8/30.1\\
\rowcolor{tableblue}BeiTv2-B & ViT & 88.8/41.4 & 88.6/49.0 & \textcolor{insightaccent}{\textbf{91.1/37.4}} & 92.5/31.2 & \textcolor{insightaccent}{\textbf{93.4/27.3}} & 92.0/31.9 & \textcolor{insightaccent}{\textbf{93.0/27.5}} & 93.6/23.4 & 93.3/24.5 & 92.1/29.9 & 91.9/30.9\\
\rowcolor{tablebluealt}DeiT3-B/16 & ViT & 83.0/61.0 & 82.5/72.5 & \textcolor{insightaccent}{\textbf{84.3/64.9}} & 87.7/50.4 & \textcolor{insightaccent}{\textbf{88.1/49.1}} & 88.6/47.8 & \textcolor{insightaccent}{\textbf{88.9/46.1}} & 86.2/58.0 & \textcolor{insightaccent}{\textbf{86.6/56.4}} & 87.4/50.0 & 87.4/51.2\\
\rowcolor{tableblue}DeiT3-B/16-In21k & ViT & 81.5/58.9 & 88.9/43.9 & \textcolor{insightaccent}{\textbf{90.9/35.4}} & 91.2/35.0 & \textcolor{insightaccent}{\textbf{92.2/31.5}} & 90.9/35.5 & \textcolor{insightaccent}{\textbf{92.0/31.8}} & 91.9/31.3 & \textcolor{insightaccent}{\textbf{92.5/29.0}} & 92.4/29.3 & \textcolor{insightaccent}{\textbf{92.5/28.6}}\\
\rowcolor{tablebluealt}DeiT3-L/16 & ViT & 82.7/60.7 & 84.0/62.2 & \textcolor{insightaccent}{\textbf{85.5/56.9}} & 88.8/44.8 & \textcolor{insightaccent}{\textbf{88.9/44.2}} & 89.4/42.6 & \textcolor{insightaccent}{\textbf{89.5/42.3}} & 87.8/46.4 & 87.9/46.9 & 89.1/41.6 & 88.9/43.0\\
\rowcolor{tableblue}DeiT3-S & ViT & 83.6/61.3 & 81.4/79.8 & \textcolor{insightaccent}{\textbf{84.2/66.6}} & 86.5/56.2 & \textcolor{insightaccent}{\textbf{87.6/51.4}} & 87.0/53.9 & \textcolor{insightaccent}{\textbf{88.0/49.2}} & 86.1/58.6 & \textcolor{insightaccent}{\textbf{87.3/52.5}} & 87.0/52.2 & 87.0/51.5\\
\rowcolor{tablebluealt}EVA02-B & ViT & 81.5/63.1 & 75.2/75.9 & \textcolor{insightaccent}{\textbf{80.1/63.5}} & 86.5/50.3 & \textcolor{insightaccent}{\textbf{86.8/48.8}} & 88.5/39.8 & \textcolor{insightaccent}{\textbf{89.2/38.5}} & 85.5/51.2 & 85.3/52.5 & 84.2/56.3 & 84.2/56.4\\
\rowcolor{tableblue}ViT-B-MAE & ViT & 40.6/96.3 & 64.0/81.4 & \textcolor{insightaccent}{\textbf{66.1/79.3}} & 61.8/83.3 & \textcolor{insightaccent}{\textbf{64.8/81.1}} & 76.7/74.0 & \textcolor{insightaccent}{\textbf{78.8/70.5}} & 64.2/81.7 & \textcolor{insightaccent}{\textbf{66.7/79.8}} & 62.1/82.1 & \textcolor{insightaccent}{\textbf{65.2/80.1}}\\
\rowcolor{tablebluealt}ViT-B/16-augreg & ViT & 87.5/46.5 & 87.8/50.1 & \textcolor{insightaccent}{\textbf{91.2/35.7}} & 92.7/27.8 & \textcolor{insightaccent}{\textbf{93.8/23.8}} & 92.0/30.9 & \textcolor{insightaccent}{\textbf{93.6/24.6}} & 94.3/22.1 & 94.1/23.7 & 92.6/31.3 & 92.1/34.3\\
\rowcolor{tableblue}ViT-B/16-in1k & ViT & 82.7/64.2 & 82.8/65.1 & \textcolor{insightaccent}{\textbf{85.2/55.9}} & 87.6/47.7 & \textcolor{insightaccent}{\textbf{88.6/43.8}} & 88.1/46.3 & \textcolor{insightaccent}{\textbf{89.2/41.8}} & 87.3/47.6 & 87.4/48.3 & 88.2/45.2 & 87.9/46.7\\
\rowcolor{tablebluealt}ViT-B/32 & ViT & 85.0/55.3 & 84.7/58.8 & \textcolor{insightaccent}{\textbf{88.2/45.5}} & 91.5/31.2 & \textcolor{insightaccent}{\textbf{92.4/28.0}} & 89.5/39.3 & \textcolor{insightaccent}{\textbf{91.6/30.0}} & 92.3/30.6 & 91.7/34.2 & 89.5/44.8 & 89.2/47.1\\
\rowcolor{tableblue}ViT-L/16 & ViT & 89.6/40.2 & 89.7/45.4 & \textcolor{insightaccent}{\textbf{92.1/34.8}} & 94.1/23.8 & \textcolor{insightaccent}{\textbf{94.9/20.8}} & 93.8/25.7 & \textcolor{insightaccent}{\textbf{94.9/20.7}} & 94.4/24.9 & 93.8/28.6 & 93.1/29.8 & 92.8/32.7\\
\rowcolor{tablebluealt}ViT-S/16 & ViT & 84.4/57.0 & 86.9/53.2 & \textcolor{insightaccent}{\textbf{89.8/40.2}} & 91.6/31.6 & \textcolor{insightaccent}{\textbf{92.6/28.2}} & 89.5/39.8 & \textcolor{insightaccent}{\textbf{91.5/31.4}} & 92.9/27.2 & 92.7/28.9 & 89.6/46.0 & 89.2/47.6\\
\rowcolor{hierheader}\multicolumn{13}{l}{\textbf{Hierarchical Transformers}}\\
\rowcolor{tableblue}Swin-B & Swin & 87.2/46.0 & 88.4/42.9 & 87.5/48.5 & 92.2/28.2 & 91.3/33.5 & 90.8/33.4 & 90.4/36.3 & 84.7/61.0 & 84.6/59.8 & 84.9/41.5 & 85.1/42.9\\
\rowcolor{tablebluealt}SwinV2-B & Swin & 82.4/62.4 & 87.1/52.8 & \textcolor{insightaccent}{\textbf{88.0/48.8}} & 90.1/40.1 & \textcolor{insightaccent}{\textbf{90.2/40.0}} & 90.1/39.7 & \textcolor{insightaccent}{\textbf{90.3/39.5}} & 89.6/41.6 & 89.6/42.2 & 90.2/40.4 & 90.0/42.0\\
\rowcolor{tableblue}SwinV2-B-In21k & Swin & 87.6/44.2 & 88.9/41.3 & 88.5/45.0 & 92.4/26.9 & 91.7/30.9 & 91.4/30.8 & 91.0/33.4 & 89.0/39.9 & 89.1/40.1 & 91.8/29.6 & 91.2/33.1\\
\rowcolor{tablebluealt}SwinV2-T & Swin & 82.9/62.2 & 82.9/67.4 & \textcolor{insightaccent}{\textbf{83.4/67.1}} & 88.8/41.1 & 88.4/43.8 & 88.8/42.1 & 88.5/44.2 & 73.5/77.0 & \textcolor{insightaccent}{\textbf{76.9/75.1}} & 81.9/66.6 & 82.1/67.1\\
\rowcolor{cnnheader}\multicolumn{13}{l}{\textbf{Convolutional encoders}}\\
\rowcolor{tableblue}ConvNeXt-B & CNN & 81.5/64.5 & 85.7/55.8 & \textcolor{insightaccent}{\textbf{86.2/54.8}} & 89.0/42.2 & 89.0/43.5 & 88.8/43.9 & 88.8/44.1 & 87.4/51.0 & 87.5/51.2 & 85.5/53.0 & \textcolor{insightaccent}{\textbf{86.5/52.5}}\\
\rowcolor{tablebluealt}ConvNeXt-B-In21k & CNN & 88.6/42.7 & 90.3/35.8 & \textcolor{insightaccent}{\textbf{91.3/32.1}} & 93.4/23.9 & \textcolor{insightaccent}{\textbf{93.6/23.3}} & 92.1/28.9 & \textcolor{insightaccent}{\textbf{92.8/26.3}} & 91.8/31.6 & \textcolor{insightaccent}{\textbf{92.3/29.7}} & 84.4/48.1 & \textcolor{insightaccent}{\textbf{87.6/43.5}}\\
\rowcolor{tableblue}ConvNeXt-L & CNN & 80.8/65.7 & 86.2/53.7 & \textcolor{insightaccent}{\textbf{86.9/52.0}} & 89.0/43.2 & 89.1/43.3 & 88.8/44.0 & \textcolor{insightaccent}{\textbf{89.1/43.4}} & 87.7/50.4 & \textcolor{insightaccent}{\textbf{88.0/49.9}} & 86.9/52.6 & \textcolor{insightaccent}{\textbf{87.9/50.3}}\\
\rowcolor{tablebluealt}ConvNeXt-T & CNN & 81.3/65.4 & 83.0/65.1 & \textcolor{insightaccent}{\textbf{84.1/61.2}} & 88.9/40.9 & 88.8/42.6 & 88.4/44.5 & 88.5/44.6 & 86.0/55.8 & \textcolor{insightaccent}{\textbf{86.2/54.9}} & 73.0/69.5 & \textcolor{insightaccent}{\textbf{78.4/65.1}}\\
\rowcolor{tableblue}ConvNeXtV2-B & CNN & 83.2/60.0 & 87.7/50.6 & 88.0/51.2 & 90.0/40.0 & 90.0/40.8 & 89.9/41.4 & \textcolor{insightaccent}{\textbf{90.0/41.0}} & 88.4/49.5 & 88.6/49.5 & 81.3/64.7 & \textcolor{insightaccent}{\textbf{85.5/56.9}}\\
\rowcolor{tablebluealt}ConvNeXtV2-B-In21k & CNN & 89.3/38.9 & 90.4/46.5 & \textcolor{insightaccent}{\textbf{91.5/40.4}} & 93.6/28.0 & \textcolor{insightaccent}{\textbf{93.9/27.1}} & 93.2/28.2 & \textcolor{insightaccent}{\textbf{93.6/27.1}} & 93.5/28.3 & \textcolor{insightaccent}{\textbf{93.8/27.4}} & 74.1/75.4 & \textcolor{insightaccent}{\textbf{84.6/63.2}}\\
\rowcolor{tableblue}ConvNeXtV2-T & CNN & 83.2/62.3 & 83.9/68.2 & \textcolor{insightaccent}{\textbf{85.4/59.7}} & 88.5/45.7 & \textcolor{insightaccent}{\textbf{88.8/44.6}} & 88.7/45.8 & \textcolor{insightaccent}{\textbf{89.1/44.1}} & 86.6/56.2 & \textcolor{insightaccent}{\textbf{87.2/54.5}} & 75.6/70.2 & \textcolor{insightaccent}{\textbf{81.2/62.3}}\\
\rowcolor{tablebluealt}DenseNet-121 & CNN & 80.5/72.8 & 77.0/73.6 & \textcolor{insightaccent}{\textbf{79.2/69.5}} & 83.3/58.3 & 83.9/58.4 & 81.9/61.8 & \textcolor{insightaccent}{\textbf{84.8/53.8}} & 83.3/59.3 & 83.6/60.2 & 60.1/84.3 & \textcolor{insightaccent}{\textbf{69.3/81.3}}\\
\rowcolor{tableblue}EffNetV2-L & CNN & 86.1/44.9 & 91.0/37.9 & 91.2/37.9 & 89.3/45.4 & \textcolor{insightaccent}{\textbf{90.1/42.4}} & 90.0/42.1 & \textcolor{insightaccent}{\textbf{91.2/35.9}} & 91.7/31.5 & 91.4/33.1 & 72.4/79.2 & \textcolor{insightaccent}{\textbf{78.4/76.3}}\\
\rowcolor{tablebluealt}EffNetV2-M & CNN & 85.8/47.0 & 89.4/45.0 & \textcolor{insightaccent}{\textbf{90.0/43.2}} & 87.6/51.5 & \textcolor{insightaccent}{\textbf{88.8/47.7}} & 89.3/44.5 & \textcolor{insightaccent}{\textbf{90.6/38.8}} & 91.2/34.8 & 91.0/37.0 & 72.1/80.3 & \textcolor{insightaccent}{\textbf{78.2/77.3}}\\
\rowcolor{tableblue}EffNetV2-S & CNN & 83.4/54.0 & 87.4/49.8 & \textcolor{insightaccent}{\textbf{88.3/45.9}} & 86.9/52.6 & \textcolor{insightaccent}{\textbf{88.0/47.7}} & 88.3/47.4 & \textcolor{insightaccent}{\textbf{89.6/41.5}} & 88.9/43.6 & 88.8/44.0 & 70.1/80.9 & \textcolor{insightaccent}{\textbf{76.3/77.8}}\\
\rowcolor{tablebluealt}RegNetY-040 & CNN & 82.9/63.8 & 83.7/52.6 & 84.1/54.7 & 85.6/47.3 & 85.8/49.2 & 87.2/47.0 & \textcolor{insightaccent}{\textbf{88.4/44.4}} & 85.3/48.6 & 85.4/50.6 & 70.1/79.5 & \textcolor{insightaccent}{\textbf{75.5/77.0}}\\
\rowcolor{tableblue}ResNet-101 & CNN & 80.4/64.6 & 87.2/44.4 & \textcolor{insightaccent}{\textbf{88.1/42.4}} & 90.9/32.7 & \textcolor{insightaccent}{\textbf{91.1/32.6}} & 89.5/41.2 & \textcolor{insightaccent}{\textbf{90.5/36.0}} & 91.0/32.7 & 91.1/32.9 & 63.7/82.9 & \textcolor{insightaccent}{\textbf{76.8/77.5}}\\
\rowcolor{tablebluealt}ResNet-152 & CNN & 81.1/63.0 & 88.6/40.5 & \textcolor{insightaccent}{\textbf{89.5/37.6}} & 91.4/31.8 & \textcolor{insightaccent}{\textbf{91.6/31.4}} & 90.3/38.6 & \textcolor{insightaccent}{\textbf{91.2/34.6}} & 91.4/31.9 & 91.6/31.9 & 68.0/81.3 & \textcolor{insightaccent}{\textbf{80.6/73.1}}\\
\rowcolor{tableblue}ResNet-50 & CNN & 82.9/66.4 & 84.4/55.1 & \textcolor{insightaccent}{\textbf{86.2/51.7}} & 86.7/56.8 & \textcolor{insightaccent}{\textbf{88.2/51.7}} & 84.1/72.1 & \textcolor{insightaccent}{\textbf{87.7/57.0}} & 87.0/56.2 & \textcolor{insightaccent}{\textbf{88.2/51.7}} & 66.5/81.7 & \textcolor{insightaccent}{\textbf{77.1/76.8}}\\
\rowcolor{tablebluealt}ResNetV2-50 & CNN & 80.5/67.7 & 83.4/56.3 & \textcolor{insightaccent}{\textbf{84.4/54.9}} & 89.2/39.7 & 89.2/40.9 & 88.4/45.2 & \textcolor{insightaccent}{\textbf{89.7/39.4}} & 84.0/57.7 & 83.8/59.0 & 64.2/81.8 & \textcolor{insightaccent}{\textbf{73.7/77.1}}\\
\rowcolor{tableblue}WideResNet50-2 & CNN & 81.4/66.6 & 86.6/46.0 & 87.2/48.1 & 90.3/35.4 & 90.5/36.5 & 88.8/44.4 & \textcolor{insightaccent}{\textbf{90.0/39.0}} & 90.6/35.2 & 90.3/37.9 & 68.5/81.4 & \textcolor{insightaccent}{\textbf{76.7/77.8}}\\
\bottomrule
\end{tabular}%
}
\end{table*}

\fi

\end{document}